\documentclass{article} 
\usepackage{colm2024_conference}

\usepackage{booktabs}
\usepackage{graphicx}
\usepackage{enumitem}
\usepackage{wrapfig}
\usepackage{algorithm}
\usepackage{algpseudocode}
\usepackage{natbib}
\usepackage{makecell}
\usepackage{booktabs}
\usepackage{array}
\usepackage{amsmath}
\usepackage{amssymb}
\usepackage{amsfonts}
\usepackage{multirow}
\usepackage{verbatim}
\usepackage{caption}
\usepackage{longtable}
\usepackage{supertabular}
\usepackage{CJKutf8}
\usepackage[utf8]{inputenc} 
\usepackage[T1]{fontenc}
\usepackage[french,vietnamese,mongolian,greek,english]{babel}
\usepackage{pifont}

\usepackage{enumitem}
\usepackage{tablefootnote}
\usepackage{xspace}
\usepackage{textcomp}
\usepackage{makecell}
\usepackage{lscape} 
\usepackage{siunitx}
\usepackage{listings}
\usepackage{xcolor}
\usepackage{svg}
\definecolor{dt}{gray}{0.7}
\usepackage{pifont}       
\usepackage{bbding}       
\usepackage{fontawesome}

\usepackage{scrextend}

\usepackage{tgpagella}
\usepackage{latexsym}
\usepackage[T1]{fontenc}
\usepackage[utf8]{inputenc}
\usepackage{microtype}
\definecolor{mydarkblue}{rgb}{0,0.08,0.45}
\definecolor{citecolor}{HTML}{0071BC}
\usepackage{url}            
\usepackage{nicefrac}       
\usepackage{changepage}
\usepackage{xargs}          
\usepackage{wrapfig,lipsum,booktabs}
\usepackage{longtable}
\usepackage{subcaption}
\usepackage{endnotes}

\usepackage{pgfplots}
\usetikzlibrary{pgfplots.groupplots}
\pgfplotsset{compat=1.3}
\usepackage{tikz}
\usetikzlibrary{patterns}

\usepackage[most]{tcolorbox}

\usepackage{hyperref}
\definecolor{darkblue}{rgb}{0, 0, 0.5}
\hypersetup{colorlinks=true, citecolor=darkblue, linkcolor=darkblue, urlcolor=darkblue, bookmarks=true, bookmarksopen=false, bookmarksnumbered=true}

\usepackage[capitalize,noabbrev]{cleveref}

\crefname{section}{\S}{\S\S}
\Crefname{section}{\S}{\S\S}
\crefname{subsection}{\S\S}{\S\S}
\Crefname{subsection}{\S\S}{\S\S}
\crefformat{section}{\S#2#1#3}
\crefformat{subsection}{\S\S#2#1#3}

\crefname{table}{Table}{Tables}
\crefname{figure}{Figure}{Figures}
\crefname{algorithm}{Algorithm}{}
\crefname{equation}{eq.}{}
\crefname{appendix}{Appendix}{}
\crefformat{section}{Section #2#1#3}
\usepackage{multicol}
\usepackage{tcolorbox}
\usepackage{titlesec}
\titleformat*{\section}{\large\bfseries}
\usepackage{array} 
\newcolumntype{P}[1]{>{\centering\arraybackslash}p{#1}} 
\usepackage{multirow}

\usepackage{adjustbox}
\definecolor{objblue}{RGB}{3,139,221}  
\definecolor{attrred}{RGB}{255,67,67}    
\definecolor{easygreen}{RGB}{0,156,75}  
\definecolor{middleyellow}{RGB}{242,89,34}  
\definecolor{hardred}{RGB}{216,56,58}
\usepackage{colortbl}
\usepackage{array}

\usepackage[most]{tcolorbox}
\definecolor{BoxBackground}{RGB}{240, 240, 240} 
\definecolor{BoxFrame}{RGB}{0, 0, 0} 
\definecolor{TitleBackground}{RGB}{0, 0, 0} 
\definecolor{TitleText}{RGB}{255, 255, 255} 
\tcbset{
  academicbox/.style={
    boxsep=5pt,
    left=2pt,
    right=2pt,
    bottom=0.5pt,
    boxrule=0.5pt,
    colback=BoxBackground,
    colframe=BoxFrame,
    colbacktitle=TitleBackground,
    coltitle=TitleText,
    enhanced,
    attach boxed title to top left={yshift=-0.1in,xshift=0.1in},
    boxed title style={boxrule=0pt,colframe=white},
    title={#1},
  }
}
\newtcolorbox{AcademicBox}[1][]{academicbox=#1}

\title{Qwen-Image Technical Report}

\author{
\bf Qwen Team}

\begin{document}

\maketitle
\vspace{-3mm}
\begin{abstract}
\vspace{-4mm}
We present Qwen-Image, an image generation foundation model in the Qwen series that achieves significant advances in complex text rendering and precise image editing.
To address the challenges of complex text rendering, we design a comprehensive data pipeline that includes large-scale data collection, filtering, annotation, synthesis, and balancing. Moreover, we adopt a progressive training strategy that starts with non-text-to-text rendering, evolves from simple to complex textual inputs, and gradually scales up to paragraph-level descriptions. This curriculum learning approach substantially enhances the model’s native text rendering capabilities. As a result, Qwen-Image not only performs exceptionally well in alphabetic languages such as English, but also achieves remarkable progress on more challenging logographic languages like Chinese.
To enhance image editing consistency, we introduce an improved multi-task training paradigm that incorporates not only traditional text-to-image (T2I) and text-image-to-image (TI2I) tasks but also image-to-image (I2I) reconstruction, effectively aligning the latent representations between Qwen2.5-VL and MMDiT.  Furthermore, we separately feed the original image into Qwen2.5-VL and the VAE encoder to obtain semantic and reconstructive representations, respectively. This dual-encoding mechanism enables the editing module to strike a balance between preserving semantic consistency and maintaining visual fidelity. We present a comprehensive evaluation of Qwen-Image across multiple public benchmarks, including GenEval, DPG, and OneIG-Bench for general image generation, as well as GEdit, ImgEdit, and GSO for image editing. Qwen-Image achieves state-of-the-art performance, demonstrating its strong capabilities in both image generation and editing. Furthermore, results on LongText-Bench, ChineseWord, and CVTG-2K show that it excels in text rendering—particularly in Chinese text generation—outperforming existing state-of-the-art models by a significant margin. This highlights Qwen-Image’s unique position as a leading image generation model that combines broad general capability with exceptional text rendering precision.

\end{abstract}

\begin{figure}[ht]
  \centering
  \begin{subfigure}[b]{0.48\textwidth}
    \centering
    \includegraphics[width=\textwidth]{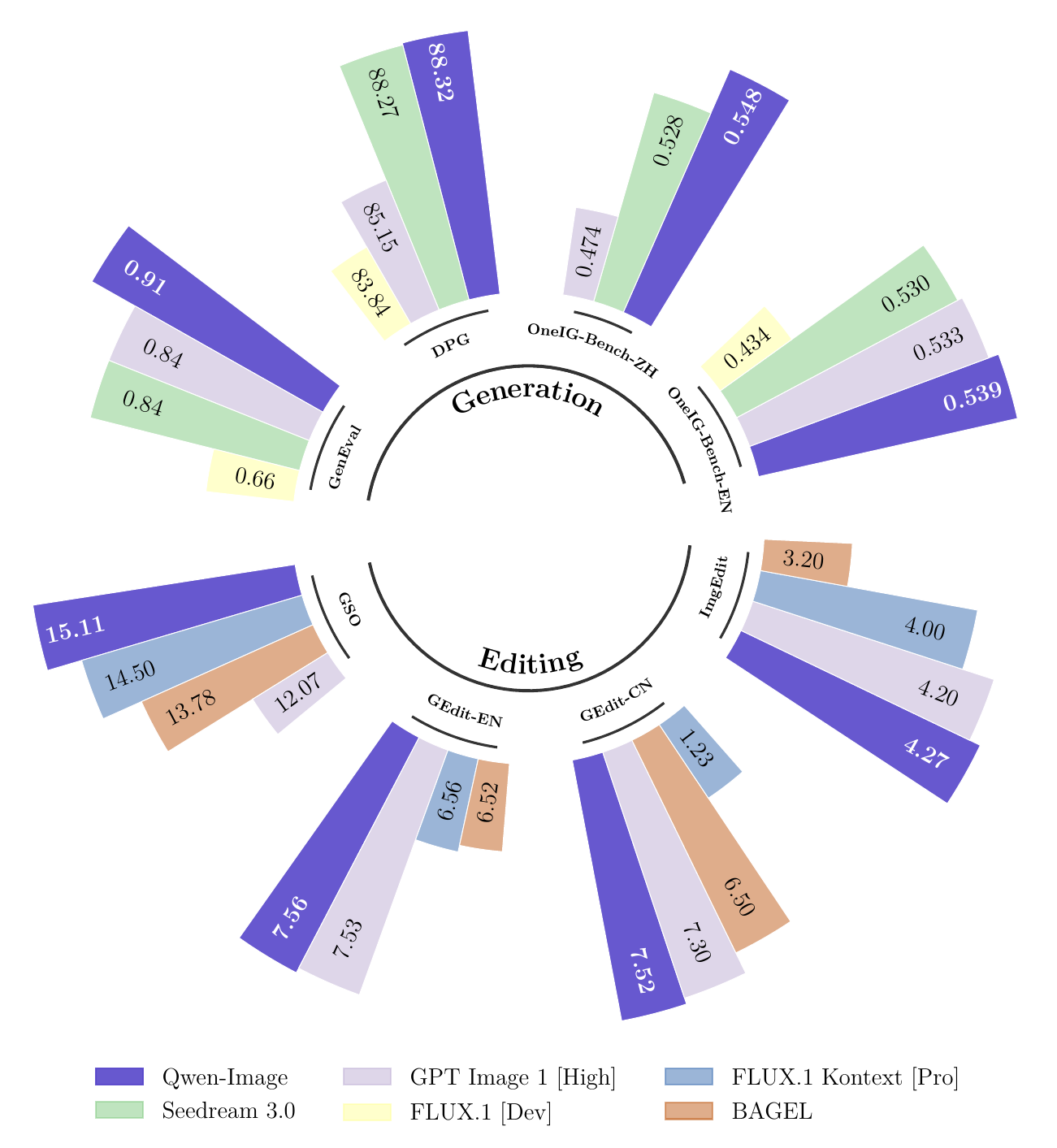}
    \caption{Image generation and editing benchmarks}
    \label{fig:radar}
  \end{subfigure}
  \hspace{1em}
  \begin{subfigure}[b]{0.48\textwidth}
    \centering
    \includegraphics[width=\textwidth]{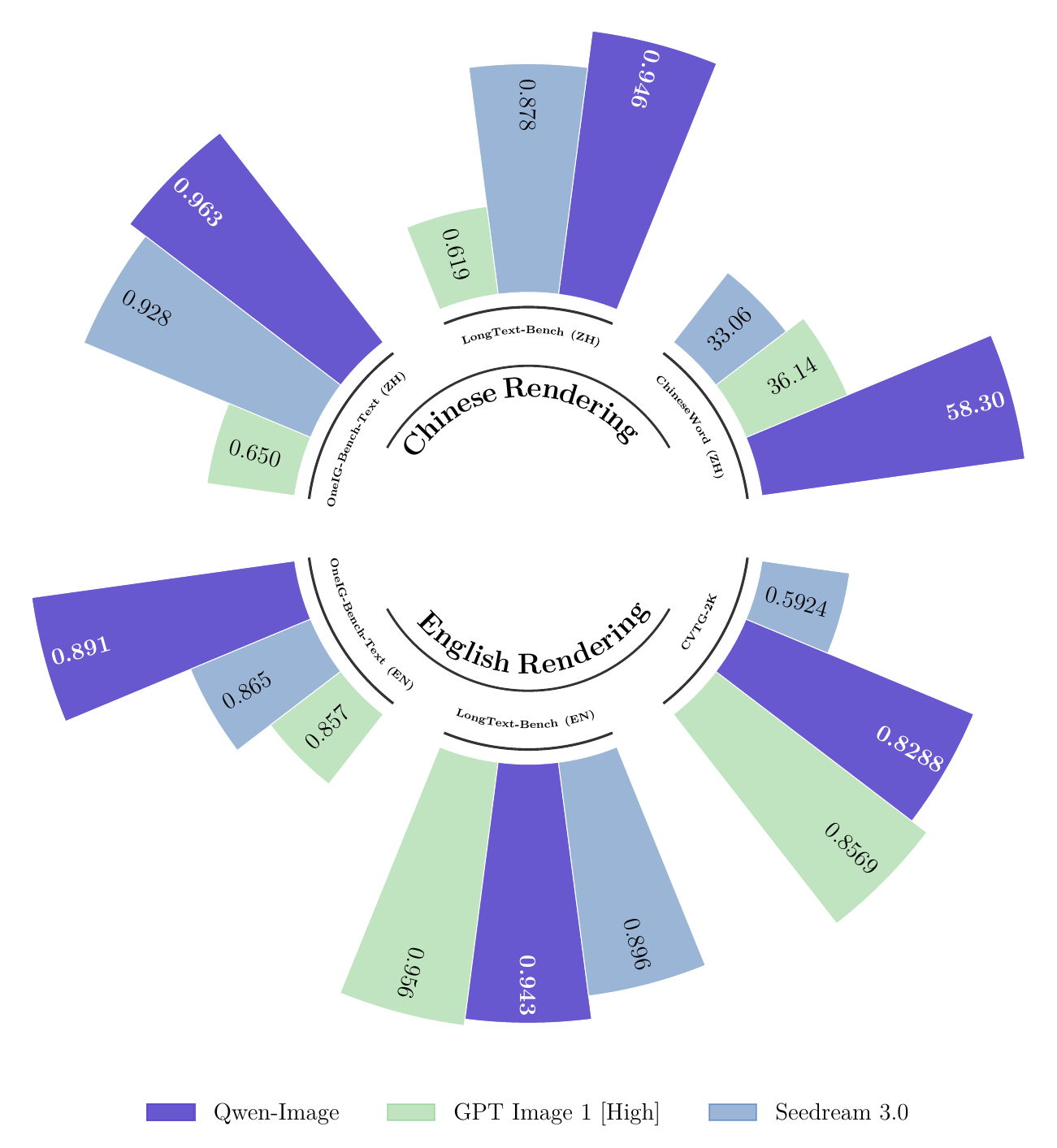}
    \caption{Text rendering benchmarks}
    \label{fig:text_radar}
  \end{subfigure}
  \vspace{-3mm}
  \caption{Qwen-Image exhibits strong general capabilities in both image generation and editing, while demonstrating exceptional capability in text rendering, especially Chinese.}
  \label{fig:two-images}
\end{figure}

\begin{figure}[p]
    \makebox[\linewidth]{
        \includegraphics[width=1.05\linewidth]{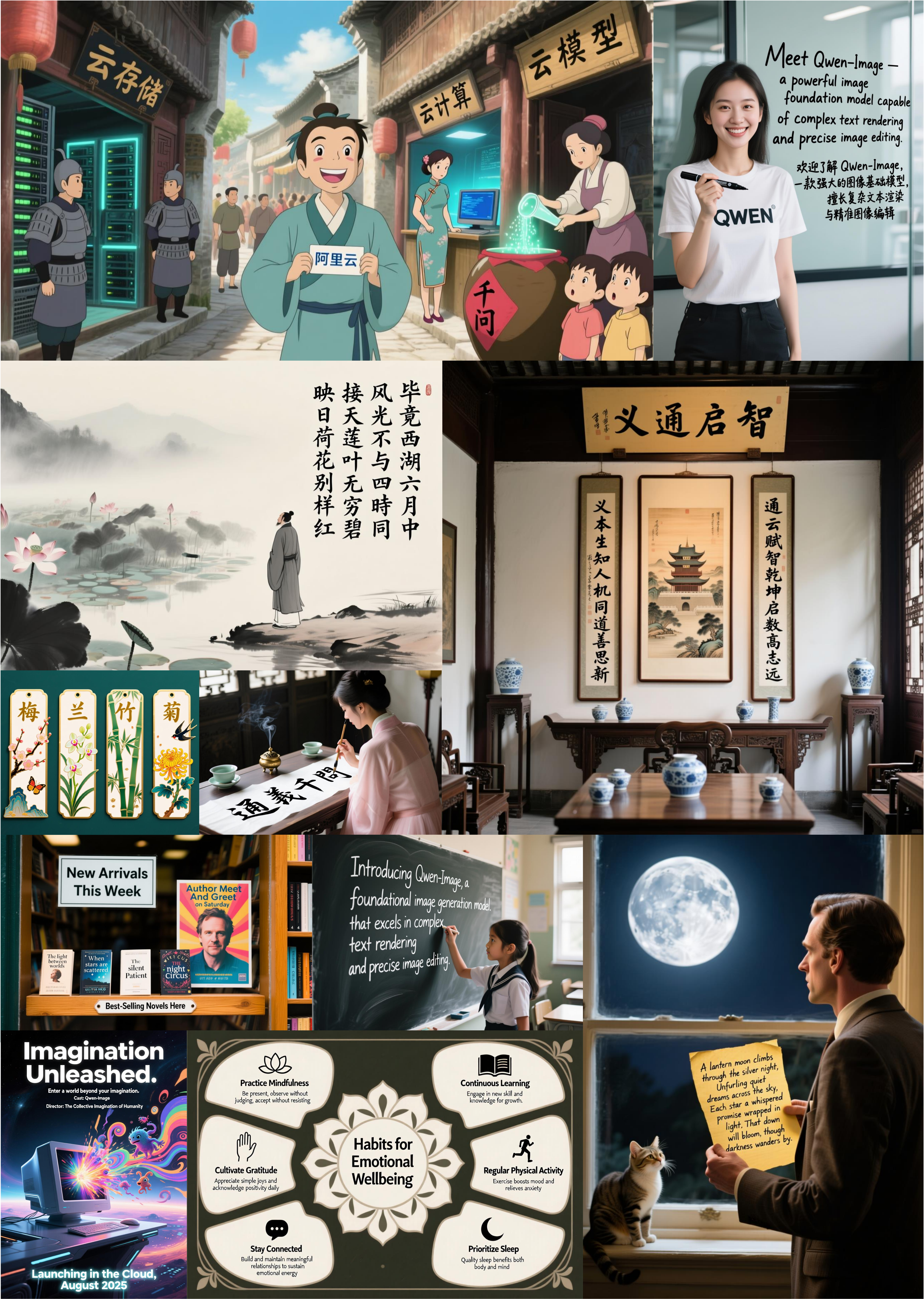}
    }
    \caption{Showcase of Qwen-Image in complex text rendering, including multi-line layouts, paragraph-level semantics, and fine-grained details. Qwen-Image supports both alphabetic languages (e.g., English) and logographic languages (e.g., Chinese) with high fidelity.
}
\end{figure}

\begin{figure}[p]
    \makebox[\linewidth]{
        \includegraphics[width=1.05\linewidth]{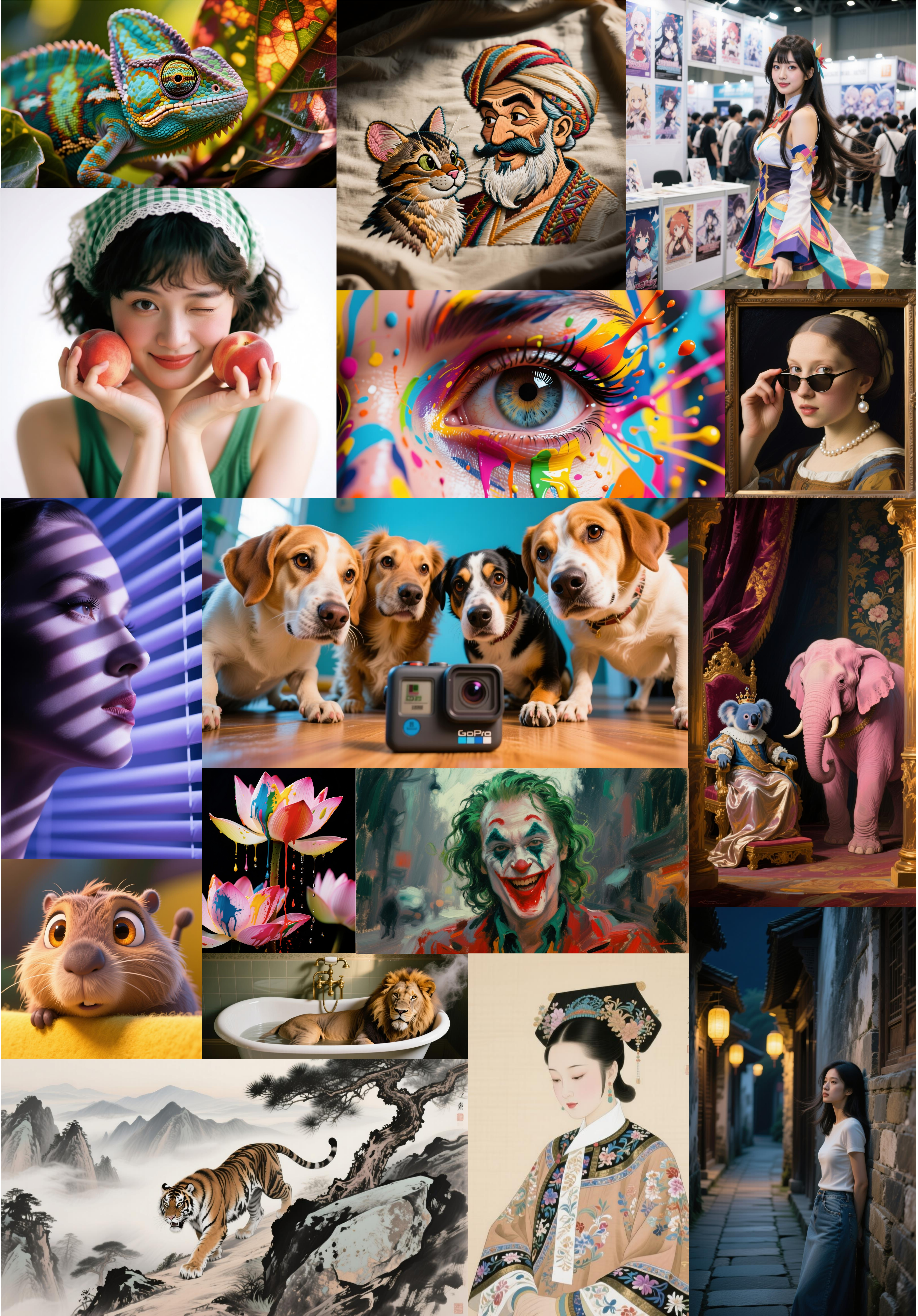}
    }
    \caption{Showcase of Qwen-Image in general image generation, supporting diverse artistic styles. }
\end{figure}

\begin{figure}[p]
    \makebox[\linewidth]{
        \includegraphics[width=1.05\linewidth]{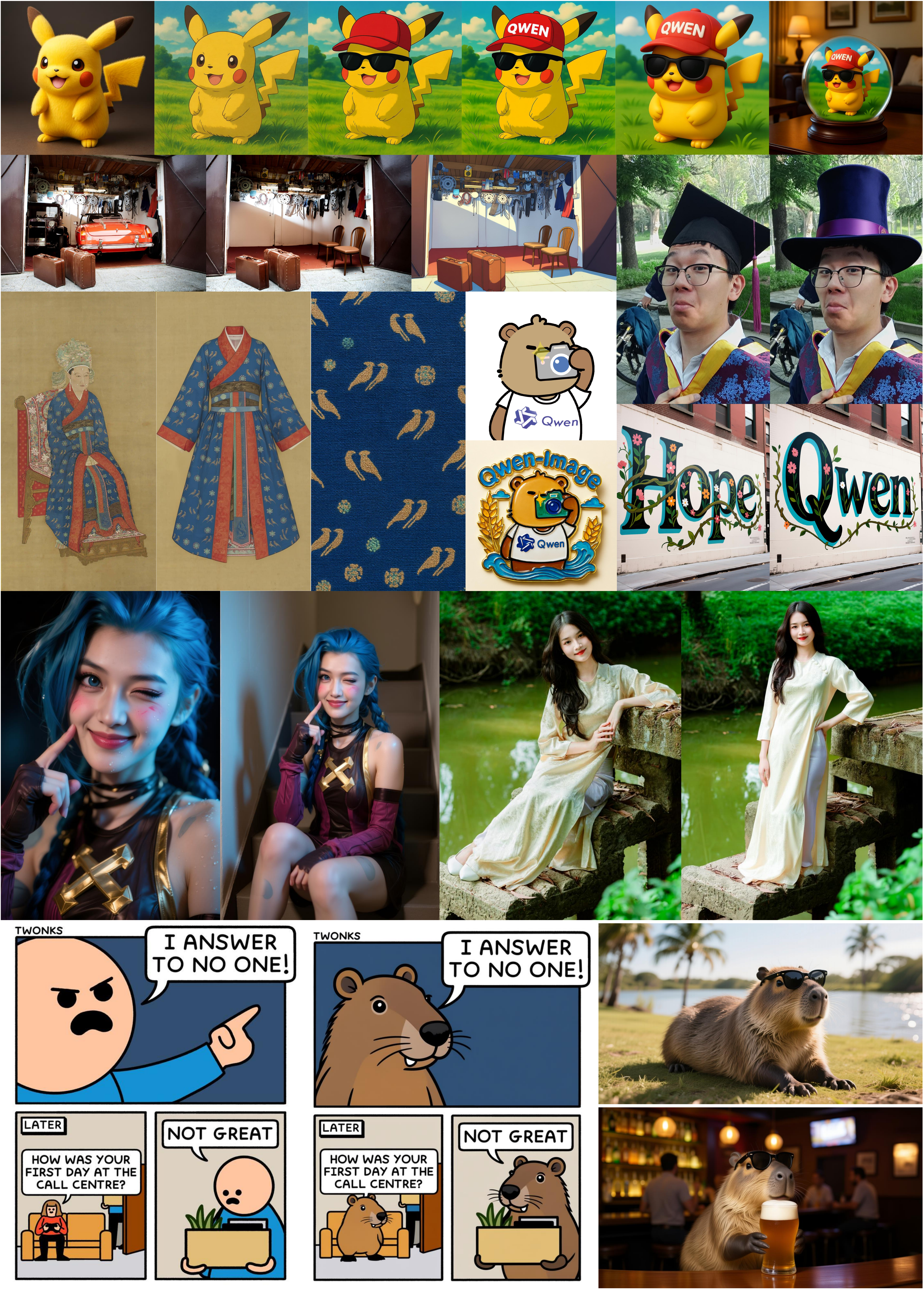}
    }
    \caption{Showcase of Qwen-Image in general image editing, including style transfer, text editing, background change, object addition, removal, and replacement, pose manipulation, and more.}
\end{figure}

\begin{figure}[p]
    \makebox[\linewidth]{
        \includegraphics[width=1.05\linewidth]{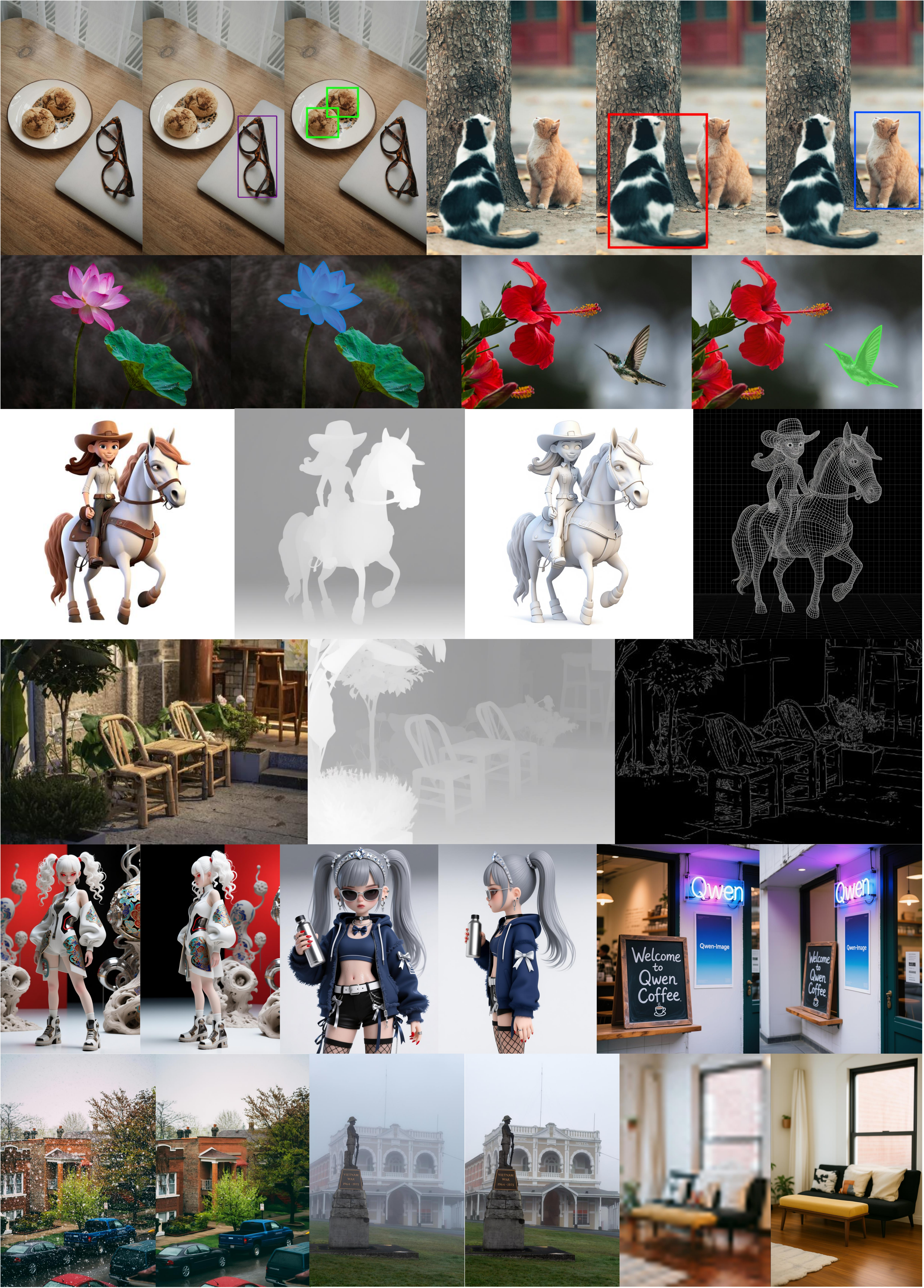}
    }
    \caption{Showcase of Qwen-Image in general image understanding tasks, including detection, segmentation, depth/canny estimation, novel view synthesis, and super resolution—tasks that can all be viewed as specialized forms of image editing.}
\end{figure}

\newpage

\section{Introduction}

Image generation models—encompassing both text-to-image generation (T2I)~\citep{rombach2021highresolution,wu2022nuwa,liang2022nuwa,openai2023dalle3,podell2023sdxl,chen2024pixartalpha,li2024hunyuandit,esser2024scaling,flux2024,gao2025seedream,gong2025seedream,cai2025hidream} and image editing (TI2I)~\citep{brooks2023instructpix2pix,zhang2023magicbrush,wang2025seededit,deng2025bagel,labs2025kontext,wu2025omnigen2,liu2025step1x,cai2025hidream,gptimage}—have emerged as a fundamental component of modern artificial intelligence, enabling machines to synthesize or modify visually compelling and semantically coherent content from text prompts. Over the past few years, remarkable progress has been achieved in this domain, particularly with the advent of diffusion-based architectures~\citep{ho2020denoising,liu2022flow} that enable high-resolution image generation while capturing fine-grained semantic details.

Despite these advances, two critical challenges persist. First, for text-to-image generation, aligning model outputs with complex, multifaceted prompts remains a significant hurdle. Our evaluation reveals that even state-of-the-art commercial models such as GPT Image 1~\citep{gptimage} and Seedream 3.0~\citep{gao2025seedream} struggle when faced with tasks requiring multi-line text rendering, non-alphabetic languages rendering (e.g., Chinese), localized text insertions, or seamless integration of text and visual elements. Second, for image editing, achieving precise alignment between the edited output and the original image poses dual challenges: (i) visual consistency, where only targeted regions should be modified while preserving all other visual details (e.g., changing hair color without altering facial details) and (ii) semantic coherence, where global semantics must be preserved during structural changes (e.g., modifying a person's pose while maintaining identity and scene coherence).

In this work, we introduce Qwen-Image, a novel image generation model within the Qwen series, designed to overcome these challenges through comprehensive data engineering, progressive learning strategies, enhanced multi-task training paradigms, and scalable infrastructure optimization.

To address the challenge of complex prompt alignment, we develop a robust data pipeline encompassing large-scale collection, annotation, filtering, synthetic augmentation, and class balancing. We further adopt a curriculum learning strategy, starting from basic text rendering tasks and progressively advancing to paragraph-level and layout-sensitive descriptions. This approach significantly enhances the model's ability to follow diverse languages, especially logographic languages like Chinese.

To address the challenge of image alignment, we propose an enhanced multi-task learning framework that seamlessly integrates T2I,I2I and TI2I objectives within a shared latent space. Specifically, the input image is encoded into two distinct yet complementary feature representations: semantic features are extracted via Qwen-VL~\citep{Qwen2.5-VL}, capturing high-level scene understanding and contextual meaning, while reconstructive features are obtained through the VAE encoder, preserving low-level visual details. Both sets of features are then jointly fed into the MMDiT architecture~\citep{esser2024scaling} as conditioning signals. This dual-conditioning design enables the model to simultaneously maintain semantic coherence and visual consistency.

To ensure training efficiency and stability at scale, we design a Producer-Consumer framework leveraging TensorPipe for distributed data loading and preprocessing. The Producer handles preprocessing tasks such as VAE encoding and data I/O, while the Consumer focuses on distributed model training using the Megatron~\citep{shoeybi2019megatron} framework. We also implement extensive monitoring tools to ensure reliable convergence and debugging capabilities throughout large-scale training.

Qwen-Image demonstrates significant advances in both generating high-quality images from complex textual prompts and performing accurate, context-aware image editing. The model is capable of interpreting intricate linguistic structures and producing visually compelling outputs that align with both semantic intent and visual constraints. To validate its effectiveness, we evaluate Qwen-Image across a diverse set of tasks, including text-to-image generation and image editing.

The key contributions of Qwen-Image can be summarized as follows:
\begin{itemize}
 \item \textbf{Superior Text Rendering:} Qwen-Image excels at complex text rendering, including multi-line layouts, paragraph-level semantics, and fine-grained details. It supports both alphabetic languages (e.g., English) and logographic languages (e.g., Chinese) with high fidelity.
 \item \textbf{Consistent Image Editing:} Through our enhanced multi-task training paradigm, Qwen-Image achieves exceptional performance in preserving both semantic meaning and visual realism during editing operations.
 \item \textbf{Strong Cross-Benchmark Performance:} Evaluated on multiple benchmarks, Qwen-Image consistently outperforms existing models across diverse generation and editing tasks, establishing a strong foundation model for image generation.
\end{itemize}

\section{Model}
\label{sec:model}
In this section, we present the architectural design of the Qwen-Image model, along with a comprehensive overview of the training data and training details.

\begin{figure}[t]
\centering
\includegraphics[width=1\linewidth]{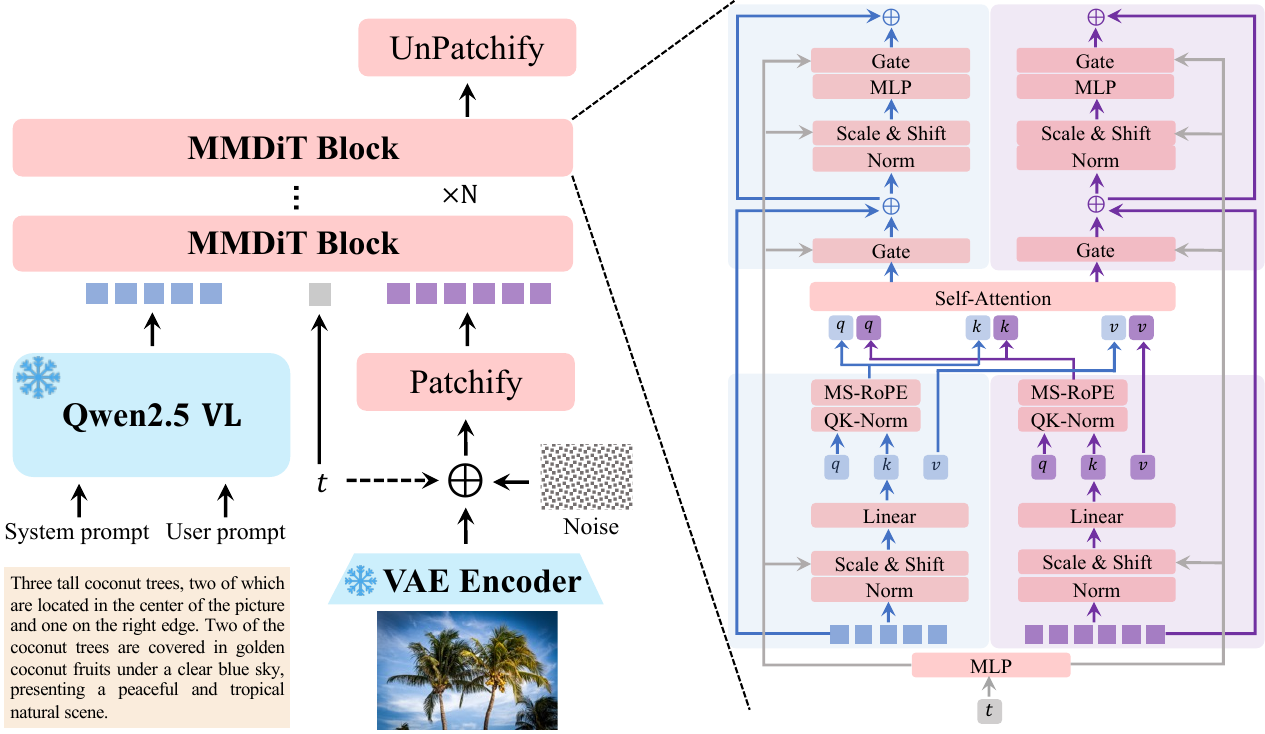}
   \caption{Overview of the Qwen-Image architecture. It adopts a standard double-stream MMDiT architecture. The input representations are provided by a frozen Qwen2.5-VL and a VAE encoder. The model employs RMSNorm~\citep{zhang2019root} for QK-Norm, while all other normalization layers use LayerNorm. Additionally, we design a new positional encoding scheme, MSRoPE~(Multimodal Scalable RoPE), to jointly encode positional information for both image and text modalities.}
\label{fig:arc}
\end{figure}

\subsection{Model Architecture}
\label{sec:model_arch}  

As shown in Figure~\ref{fig:arc}, the Qwen-Image architecture is built upon three core components that work in concert to enable high-fidelity text-to-image generation. 
First, a \textbf{Multimodal Large Language Model~(MLLM)} serves as the condition encoder, responsible extracting feature from  textual inputs. 
Second, a \textbf{Variational AutoEncoder~(VAE)} acts as the image tokenizer, compressing input images into compact latent representations and decoding them back during inference. 
Third, a \textbf{Multimodal Diffusion Transformer~(MMDiT)} functions as the backbone diffusion model, modeling the complex joint distribution between noise and image latents under text guidance.  While this section outlines their general roles, the specific model choices and architectural details are elaborated in the following sections.

\subsection{Multimodal Large Language Model}
\label{subsec:qwen2d5-vl}
Qwen-Image employs the Qwen2.5-VL model~\citep{Qwen2.5-VL} as the feature extraction module for textual inputs, owing to three key reasons: (1) The language and visual spaces of Qwen2.5-VL have already been aligned, which makes it more suitable for text-to-image tasks compared to language-based models like Qwen3~\citep{yang2025qwen3};
(2) Qwen2.5-VL retains strong language modeling capabilities, without significant degradation compared to language models;
(3) Qwen2.5-VL supports multimodal inputs, thereby enabling Qwen-Image to unlock a broader range of functionalities, e.g., image editing~\citep{labs2025kontext}.
Let $x$ and $y$ denote the image and textual inputs, respectively. Given the user inputs, such as prompts and images, we adopt the Qwen2.5-VL model to extract features.
To better guide the model in generating the refined representation latent, while accounting for the varying input modalities across different tasks, we design distinct system prompts tailored for \textbf{pure text input} and \textbf{text-and-image input}, respectively.
We illustrate the system template in Figure~\ref{sys_prompt_t2i} and Figure~\ref{sys_prompt_ti2i}. 
Finally, we utilize the latent of the last layer's hidden state from Qwen2.5-VL language model backbone as the representation of the user input.

\begin{figure}[t]
\begin{AcademicBox}[\footnotesize System Prompt for T2I task]
<|im\_start|>system \\ 
Describe the image by detailing the color, quantity, text, shape, size, texture, spatial relationships of \\ 
the objects and background: <|im\_end|> \\
<|im\_start|>user \\
\textbf{\textcolor{red}{<|user\_text|>}}<|im\_end|> \\
<|im\_start|>assistant
\end{AcademicBox}
\caption{System prompt for Text-to-Image generation task, where \textbf{\textcolor{red}{<|user\_text|>}} is the user input prompt.\looseness=-1}
\label{sys_prompt_t2i}
\end{figure}

\subsection{Variational AutoEncoder}
\label{subsec:vae}
A strong VAE representation is crucial for building a powerful image foundation model. Current image foundation models typically train an image VAE with 2D convolutions on massive image datasets to obtain a high-quality image representation. In contrast, our work aims to develop a more general visual representation compatible with both images and videos. However, existing joint image-video VAEs, such as Wan-2.1-VAE~\citep{wan2025wan}, typically suffer a performance trade-off that results in degraded image reconstruction capabilities. To this end, we leverage a single-encoder, dual-decoder architecture. This design utilizes a shared encoder compatible with both images and videos, alongside separate, specialized decoders for each modality, which enables our image foundation model to serve as a backbone for future video models. Specifically, we adopt the architecture of Wan-2.1-VAE, freeze its encoder, and exclusively fine-tune the image decoder.

To enhance reconstruction fidelity, particularly for small texts and fine-grained details, we train the decoder on an in-house corpus of text-rich images. The dataset consists of real-world documents (PDFs, PowerPoint slides, posters) alongside synthetic paragraphs, covering both alphabetic (e.g., English) and logographic (e.g., Chinese) languages. During training, we observe that: (1) Balancing reconstruction loss with perceptual loss effectively reduces grid artifacts, which are often observed in repetitive textures like bushes. (2) As reconstruction quality increases, adversarial loss becomes ineffective because the discriminator is unable to provide effective guidance. Based on these observations, we use only reconstruction and perceptual losses, dynamically adjusting their ratio during fine-tuning. Interestingly, we find that finetuning only the decoder can effectively enhance details and improve the rendering of small text, thereby building a solid foundation for Qwen-Image's text rendering abilities. Quantitative and qualitative results are presented in Section~\ref{sec:vae_exp}.

\subsection{Multimodal Diffusion Transformer}
\label{subsec:mmdit}
Qwen-Image adopts Multimodal Diffusion Transformer~(MMDiT)~\citep{esser2024scaling} to jointly model text and images. This approach has proven effective in a range of works, such as the FLUX~\citep{flux2024,labs2025kontext} series and the Seedream~\citep{gong2025seedream,gao2025seedream} series.

\begin{figure}[t]
    \centering
    \includegraphics[width=0.9\linewidth]{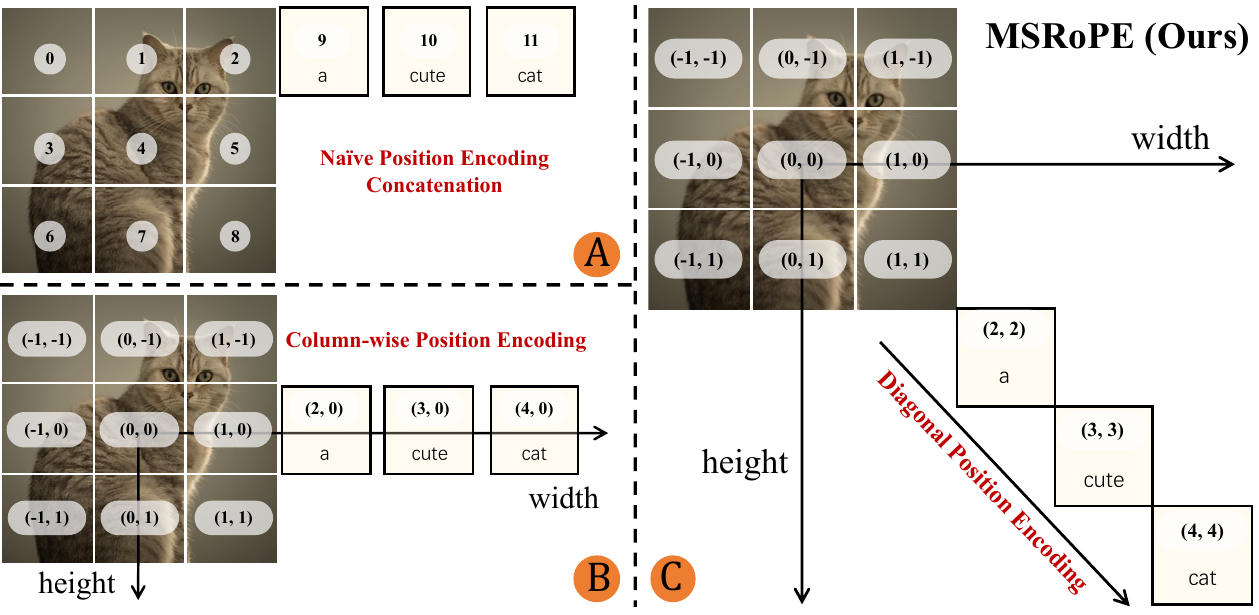}
    \caption{Comparison of different image-text joint positional encoding strategies. We design the Multimodal Scalable RoPE~(MSRoPE) strategy, which starts encoding from the image center and positions textual encodings along the diagonal of the grid, enabling better image resolution scaling and improved text-image alignment.}
    \label{fig:msrope}
\end{figure}

Within each block, we introduce a novel positional encoding method: Multimodal Scalable RoPE~(MSRoPE). 
As illustrated in Figure~\ref{fig:msrope}, we compare various text-image joint positional encoding strategies. In the traditional MMDiT block, text tokens are directly concatenated after the flattened image positional embeddings. 
Furthermore, Seedream 3.0~\citep{gao2025seedream} introduces Scaling RoPE, where the image positional encoding is shifted to the central region of the image, and text tokens are considered as 2D tokens with a shape of [1, L].
Then, 2D RoPE~\citep{heo2024rotary} is used for image-text joint positional encoding.
Although this adjustment facilitates resolution scale training, certain rows of positional encodings for text and image, e.g., the 0-th middle row in Figure~\ref{fig:msrope} (B), become isomorphic, making it harder for the model to distinguish between text tokens and the image latent tokens in the 0-th middle row.
Yet, it is also non-trivial to determine an appropriate image row to concatenate the text tokens.
To address the aforementioned challenges, we introduce Multimodal Scalable RoPE~(\textbf{MSRoPE}). In this approach, text inputs are treated as 2D tensors with identical position IDs applied across both dimensions. As depicted in Figure~\ref{fig:msrope} (C), the text is conceptualized as being concatenated along the diagonal of the image. This design allows MSRoPE to leverage resolution scaling advantages on the image side while maintaining functional equivalence to 1D-RoPE on the text side, thereby obviating the need to determine the optimal positional encoding for text.
We show the architecture and configuration of Qwen-Image in Table~\ref{tab:hyper-param-qwen-image-arch}.

\begin{table}[t]
    \centering
    \caption{Configuration of Qwen-Image architecture.}
    \begin{tabular}{l|cc|cc|c}
    \toprule
    \bf \multirow{2}{*}{Configuration} & \multicolumn{2}{c|}{\textbf{VLM}} & \multicolumn{2}{c|}{\textbf{VAE}} & \bf \multirow{2}{*}{MMDiT} \\
    \cmidrule{2-5}
    & ViT & LLM & Enc & Dec & \\
    \midrule
    \# Layers & 32 & 28 & 11 & 15 & 60 \\
    \# Num Heads~(Q / KV) & 16 / 16 & 28 / 4 & - & - & 24 / 24 \\
    Head Size & 80 & 128 & - & - & 128 \\
    Intermediate Size & 3,456 & 18,944 & - & - & 12,288 \\
    Patch / Scale Factor & 14 & - & 8x8 & 8x8 & 2 \\
    Channel Size & - & - & 16 & 16 & - \\
    \midrule
    \# Parameters & \multicolumn{2}{c|}{7B} & 54M & 73M & 20B \\
    \bottomrule
    \end{tabular}
    
    \label{tab:hyper-param-qwen-image-arch}
\end{table}

\section{Data}

\subsection{Data Collection}
\label{sec:date_collection}

\begin{figure*}[t]
\centering
\includegraphics[width=\linewidth]{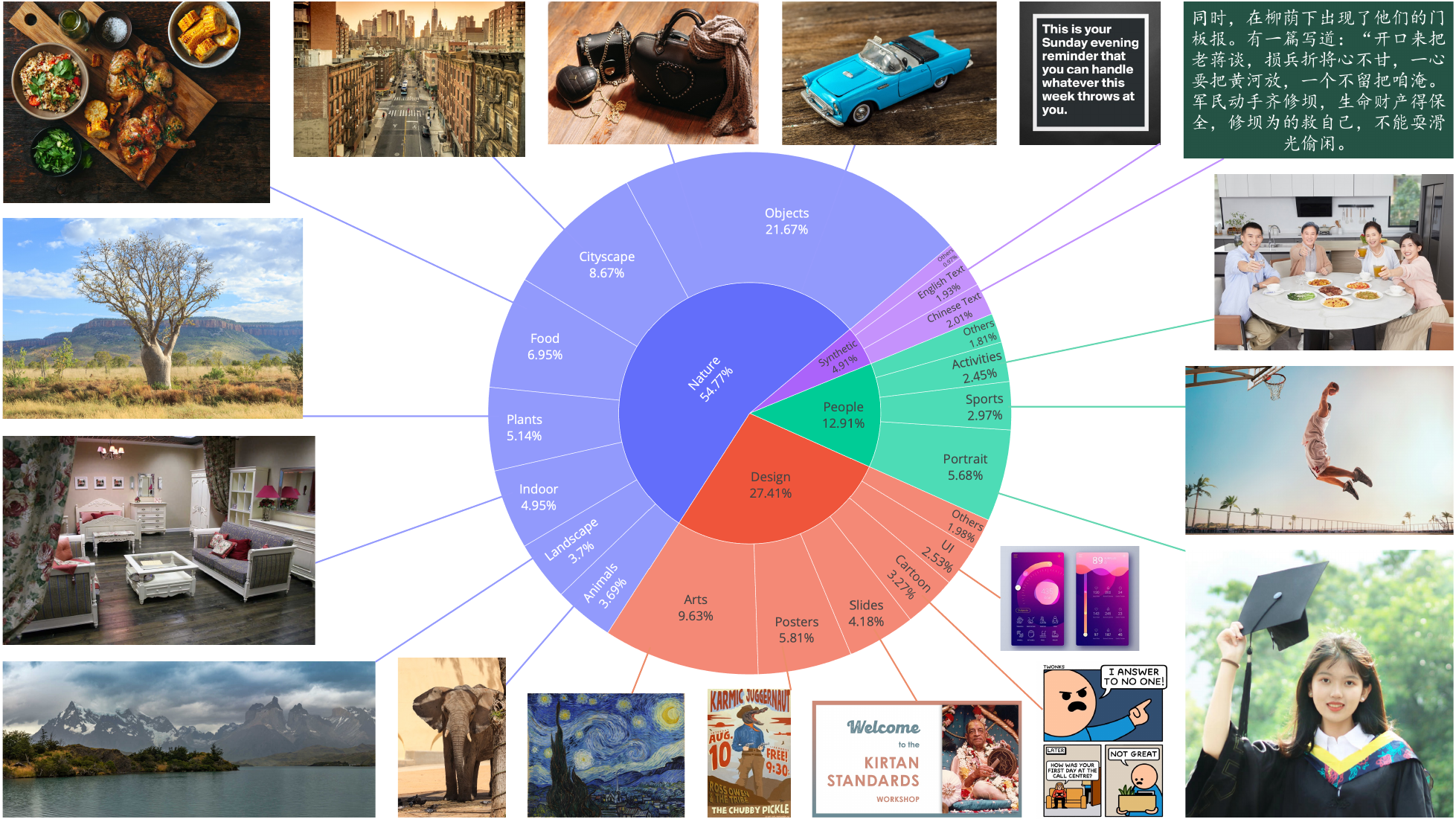}
   \caption{
      Overview of Data Collection.
      The dataset includes four main categories: Nature (general-purpose generation), People (human-centric generation), Design (artistic styles, text rendering, and complex layouts), and Synthetic Data (text rendering enhancement). 
      Our collection strategy balances diversity and quality while training, ensuring broad coverage and high-fidelity annotations to support robust model training.
   }
\label{fig:data_collection}
\end{figure*}

We systematically collected and annotated billions of image-text pairs to support the training of our image generation model. 
Rather than focusing solely on the scale of raw dataset, we prioritize data quality and balanced data distribution, aiming to construct a well-balanced and representative dataset that closely mirrors real-world scenarios. 
As illustrated in Figure~\ref{fig:data_collection}, the dataset is organized into four primary domains: {Nature}, {Design}, {People} and {Synthetic Data}.

Among these categories, {Nature} category constitutes the largest proportion, accounting for approximately 55\% of the dataset. 
This category includes diverse subcategories such as {Objects}, {Landscape}, {Cityscape}, {Plants}, {Animals}, {Indoor}, and {Food} categories. 
Additionally, content that does not clearly belong to the {People} or {Design} categories is also classified under {Nature} category.
This broad, general-purpose category serves as a crucial foundation for improving the model's ability to generate realistic and diverse natural images.

The second largest category is the {Design} category, comprising around 27\% of the dataset. 
It primarily includes structured visual content such as Posters, User Interfaces, and Presentation Slides, as well as various forms of art including paintings, sculptures, art crafts, and digital arts. 
These types of data often contain rich textual elements, complex layouts, and design-specific visual semantics. 
This category is particularly important for enhancing the model's capabilities in following intricate prompts about artistic styles, text rendering, and layout design.

Next, the {People} category makes up about 13\% of the dataset, encompassing subcategories such as {Portrait}, {Sports}, and {Human Activities}. 
It comprises a wide range of human-related images, including {Portrait}, {Sports}, {Activities}, and so on.
This category is essential for improving the model's ability to generate realistic and diverse human images, to ensure satisfactory user experiences and practical applicability.\looseness=-1

Finally, the {Synthetic Data} category accounts for approximately 5\% of the dataset. 
It is important to clarify that the synthetic data discussed here does not include images generated by other AI models, but rather data synthesized through controlled text rendering techniques (described in \cref{sec:date_synthesis}). 
This excludes images synthesized by other AI models, which often introduce significant risks such as visual artifacts, text distortions, biases, and hallucinations.
We adopt a conservative stance toward such data, as training on low-fidelity or misleading images may weaken the model's generalization capabilities and undermine its reliability.

\begin{figure*}[t]
\centering
\includegraphics[width=\linewidth]{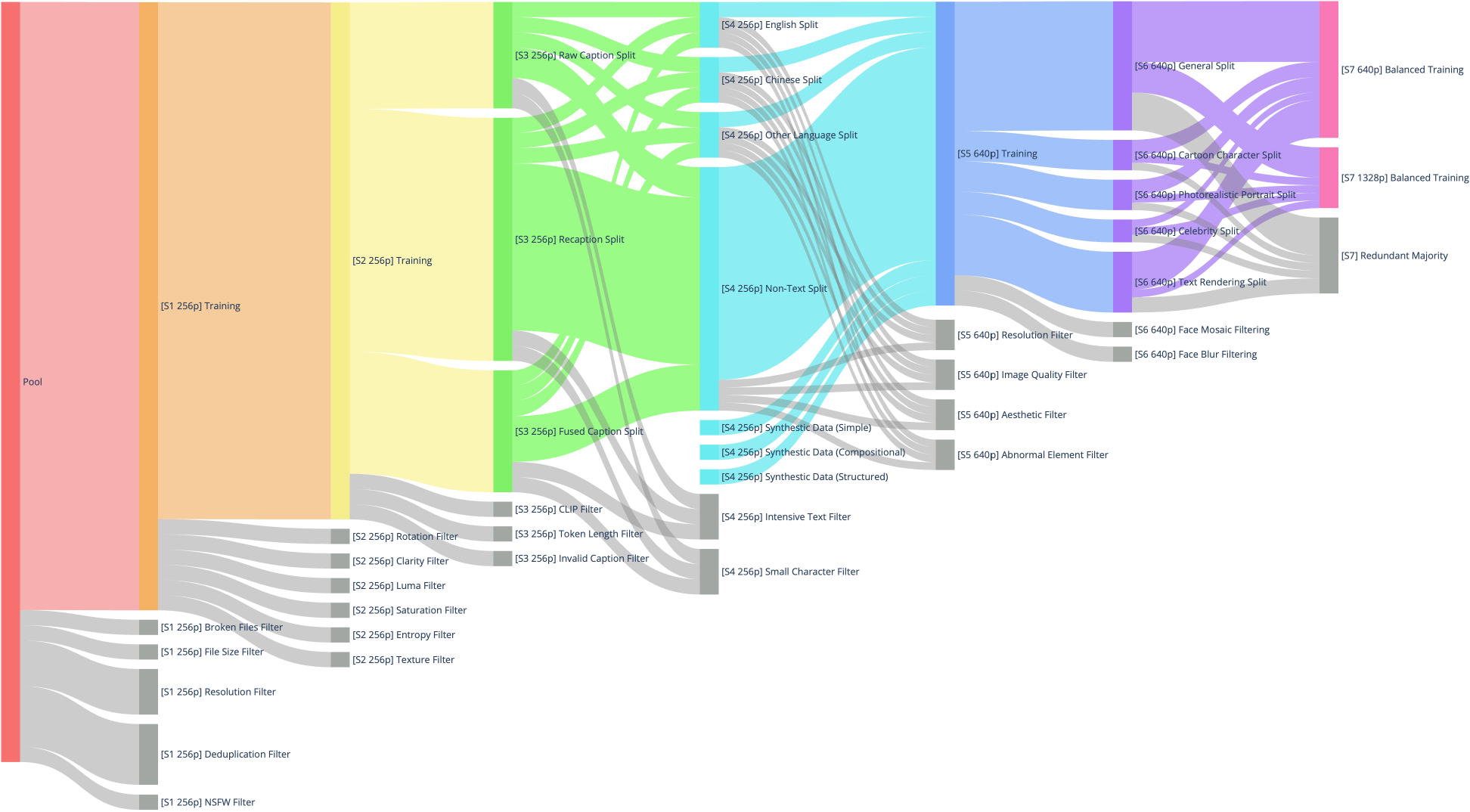}
\caption{
      Overview of the multi-stage data filtering pipeline. 
      Our filtering strategy consists of seven sequential stages (S1–S7), each targeting specific aspects of data quality, alignment, and diversity. 
      From initial pre-training data curation to high-resolution refinement and multi-scale training, the pipeline progressively improves dataset quality while maintaining semantic richness and distributional stability.
   }

\label{fig:data_filtering}
\end{figure*}

\subsection{Data Filtering}
\label{sec:data_filtering}

To curate high-quality training data throughout the iterative development of our image generation model, we propose a multi-stage filtering pipeline comprising seven sequential stages, as depicted in \cref{fig:data_filtering}. 
These stages are progressively applied throughout the training process, with data distributions continuously refined over time. 
Notably, synthetic data are introduced from Stage 4, when the foundational model has reached a certain level of stability. 
The following section presents a comprehensive description of each stage.

\paragraph{Stage 1: Initial Pre-training Data Curation}
At this early stage, the model is trained on images resized to 256p (256×256 pixels with various aspect ratios including 1:1, 2:3, 3:2, 3:4, 4:3, 9:16, 16:9, 1:3 and 3:1). 
To improve data quality, a series of filters are applied to remove low-quality or irrelevant images. 
The Broken Files Filter identifies and discards corrupted or partially damaged files (e.g., truncated images), many of which are also associated with abnormally small file sizes, as detected by the File Size Filter. 
The Resolution Filter removes images with original resolutions below 256p. 
The Deduplication Filter eliminates duplicate or near-duplicate image-text pairs. 
Furthermore, the NSFW Filter is applied to exclude content containing sexual, violent, or other offensive material.

\begin{figure*}[t]
\centering
\includegraphics[width=\linewidth]{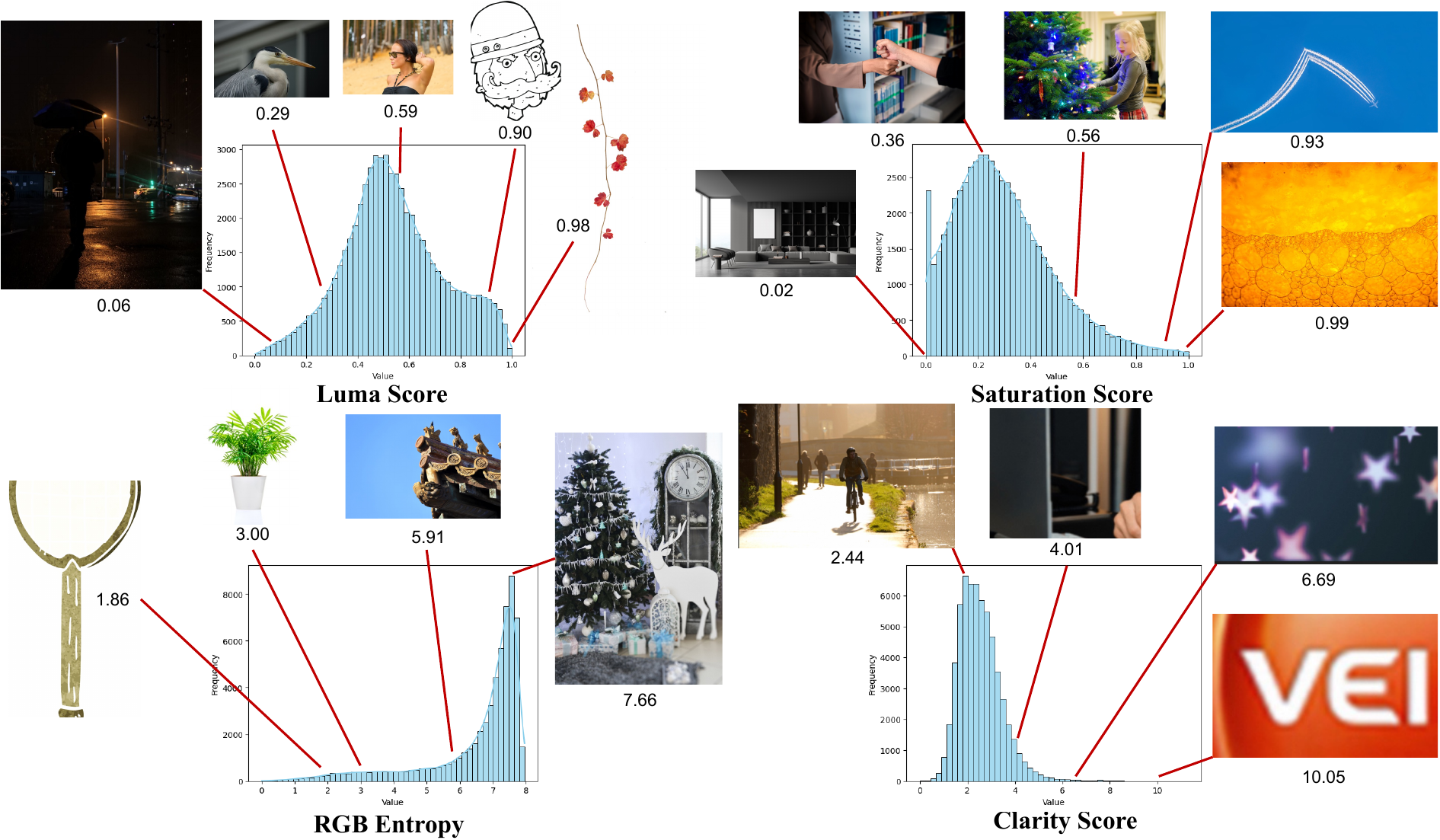}
   \caption{
      Examples of some filtering operators used in our pipeline.
      Extreme values in these operators often indicate atypical or low-quality images. 
      For instance, images with excessively high Luma score typically contain large areas of white or overexposed regions.
   }
\label{fig:data_filter_operator}
\end{figure*}

\paragraph{Stage 2: Image Quality Enhancement}
In this stage, we focus on systematically improving the image quality of the dataset. 
The Rotation Filter removes images with significant rotation or flipping, as indicated by the EXIF metadata. 
The Clarity Filter discards blurry or out-of-focus images, ensuring that only sharp and clear images are retained.
The Luma Filter excludes images that are excessively bright or dark, while the Saturation Filter eliminates images with unnaturally high color saturation, which often suggests artificial rendering or unrealistic digital manipulations. 
Furthermore, the Entropy Filter identifies and removes images with significantly low entropy, typically characterized by large uniform regions or minimal visual content. 
Finally, the Texture Filter is employed to discard images with overly complex textures, which are often associated with noise or non-semantic patterns.
\cref{fig:data_filter_operator} illustrates some examples of the filtering operators used in this stage.

\paragraph{Stage 3: Image-Text Alignment Improvement}
This stage focuses on improving the alignment between textual descriptions and visual content. 
To balance the training data distribution, the dataset is divided into three splits based on the source of captions: Raw Caption Split, Recaption Split, and Fused Caption Split.
Raw Caption Split includes captions provided by websites as well as metadata such as titles or tags originally associated with the images.
Although raw captions may introduce noise, they contribute to the model's robustness in handling short text inputs and serve as a vital source of real-world knowledge (e.g., plant names, cartoon IPs) often absent in datasets with synthesized captions.
Recaption Split consists of captions generated by the most advanced Qwen-VL Captioner~\citep{Qwen2.5-VL}, which provides more descriptive and structured annotations.
Due to model limitations, not all IPs can be accurately identified.
Fused Caption Split combines both raw captions and synthesized captions, offering a blend of general knowledge and detailed descriptions.
To further improve alignment, we applied both the Chinese CLIP~\citep{chinese_clip} Filter and the SigLIP 2~\citep{tschannen2025siglip} Filter to remove mismatched image-text pairs from the Raw Caption Split. 
Additionally, a Token Length Filter was employed to eliminate excessively long captions, and the Invalid Caption Filter discards captions with abnormal content, such as ``Sorry, I cannot provide a caption for this image.''.

\paragraph{Stage 4: Text Rendering Enhancement}
In this stage, we focus on improving the model's capability in rendering text within images, which is crucial for generating images with high textual fidelity.
To this end, we categorize the dataset based on the presence and language of text within images.
Specifically, the dataset from Stage 3 is divided into four splits: English Split, Chinese Split, Other Language Split, and Non-Text Split, to ensure balanced training across different linguistic contexts.
To address challenges such as low-frequency characters, mixed-language scenarios, and font diversity, we incorporate synthetic text rendering data, which are generated using the strategies described in \cref{sec:date_synthesis}.
Moreover, the Intensive Text Filter and the Small Character Filter are applied to remove images with overly dense or excessively small text, as such cases are challenging to annotate accurately and difficult to render legibly.

\paragraph{Stage 5: High-Resolution Refinement}
In this stage, the model transitions to training with images at 640p resolution, accompanied by further dataset refinement to ensure both high quality and aesthetic appeal. 
The Image Quality Filter is applied to eliminate images with quality defects such as overexposure, underexposure, blur, or compression artifacts. 
The Resolution Filter ensures that all images meet the minimum resolution requirements.
The Aesthetic Filter is employed to exclude images with poor composition or visual appeal.
Finally, the Abnormal Element Filter removes images containing watermarks, QR codes, barcodes, or other elements that can interfere with viewing.

\paragraph{Stage 6: Category Balance and Portrait Augmentation}
After identifying underperformed categories through careful error analysis, this stage involves recategorizing the dataset into three primary categories: General, Portrait, Text Rendering, to facilitate category-based rebalancing during training.
Keyword-based retrieval and image retrieval techniques are employed to augment the dataset with targeted patches, to enhance coverage of underrepresented categories.
To further improve the model's ability to generate high-quality portraits, we first retrieve photorealistic portraits, cartoon characters, and celebrity images from the People category.
Synthesized captions are then generated to emphasize character-specific details, such as facial features, expressions, and clothing, as well as contextual elements such as background, lighting, and mood.
This approach aims to enhance both the quality of generated images and the model's instruction-following capability.
Additional filters are applied to remove images with face mosaics or blurs, to avoid potential privacy issues and ensure the model's robustness in handling human subjects.

\paragraph{Stage 7: Balanced Multi-Scale Training}

In the final stage, the model is trained jointly on images with resolutions of 640p and 1328p.
Imposing a strict resolution threshold of 1328p would lead to significant data loss and distort the underlying data distribution. 
To improve training efficiency and ensure balanced data distribution, we design a hierarchical taxonomy system for image categorization, inspired by the design principles of WordNet~\citep{10.1145/219717.219748}.
All images in Stage 6 are classified according to this hierarchical taxonomy.
Within each category, we retain only images with the highest quality and aesthetic appeal.
Furthermore, a specialized resampling strategy is employed to balance data containing text rendering, addressing the long-tail distribution of token frequencies.
Such balanced multi-scale training allows the model to retain previously learned general knowledge and ensure stable convergence while adapting to higher-resolution inputs, thereby improving detail generation without sacrificing robustness.\looseness=-1

\subsection{Data Annotation} 
\label{sec:data_annotation}

\begin{figure}[t]
\small
\begin{AcademicBox}[\footnotesize Qwen-Image Annotation Prompt]
\begin{verbatim}
# Image Annotator
You are a professional image annotator. Please complete the following tasks based on the input image.

## Step 1: Create Image Caption
1. Write the caption using natural, descriptive text without structured formats or rich text.
2. Enrich caption details by including: object attributes, vision relations between objects, and 
environmental details.
3. Identify the text visible in the image, without translation or explanation, and highlight 
it in the caption with quotation marks.
3. Maintain authenticity and accuracy, avoid generalizations.

## Step 2: Image Quality Assessment
1. Image Type Identification: Return the image type based on its source and usage.
2. Image Style Identification: Return the image style based on its overall visual characteristics.
3. Watermark Detection: Detect watermarks in the image. Return the detected watermarks in a list format.
4. Abnormal Element Detection: Check if any elements affect viewing, such as QR codes or mosaics.

## Sample Output Format
```json
{
   "Caption": "...",
   "Image Type": "...",
   "Image Style": "...",
   "Watermark List": [],
   "Abnormal Element": "yes/no",
}
```
\end{verbatim}
\end{AcademicBox}
\caption{
   Example of the annotation prompt used in Qwen-Image.
}
\label{fig:qwen_image_prompt}
\end{figure}

In our data annotation pipeline, we utilize a capable image captioner (e.g., Qwen2.5-VL) to generate not only comprehensive image descriptions, but also structured metadata that captures essential image properties and quality attributes.

Instead of treating captioning and metadata extraction as independent tasks, we designed an annotation framework in which the captioner concurrently describes visual content and generates detailed information in a structured format, such as JSON.
Critical details such as object attributes, spatial relationships, environmental context, and verbatim transcriptions of visible text are captured in the caption, while key image properties like type, style, presence of watermarks, and abnormal elements (e.g., QR codes or facial mosaics) are reported in a structured format.
With the help of advanced captioner, this methodology transcends traditional image captioning and generate both detailed image descriptions and structured metadata in a single pass, as shown in \cref{fig:qwen_image_prompt}.
The annotation pipeline is designed to be efficient and scalable, allowing us to process large-scale datasets without relying on additional models or post-processing steps.
In practice, we further refine the above pipeline by integrating initial annotations with expert rules and lightweight classification models for critical tasks, such as watermark verification and content filtering.

Overall, our pipeline not only provides deep insights into image content, but also supports advanced data curation, providing a solid foundation for training robust and reliable image generation models.

\subsection{Data Synthesis}
\label{sec:date_synthesis}

\begin{figure*}[t]
\centering
\includegraphics[width=\linewidth]{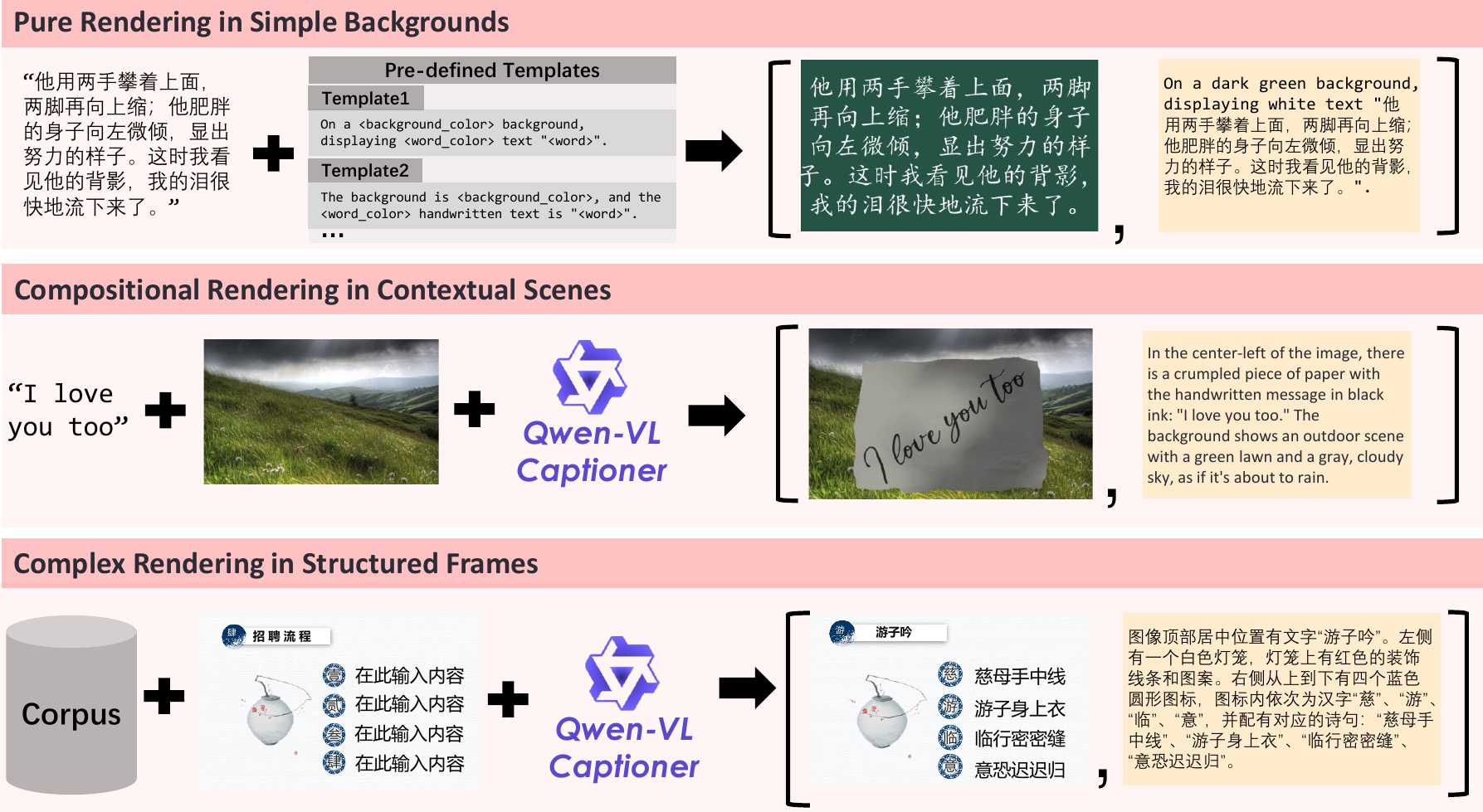}
   \caption{
      Overview of Data Synthesis. 
      We designed three rendering strategies—Pure Rendering, Compositional Rendering , and Complex Rendering —to generate text-only data, text-in-context data, and complex-layout data, respectively.
   }
\label{fig:data_synthesis}
\end{figure*}

Given the long-tail distribution of textual content in real-world images, particularly for non-Latin languages such as Chinese, where numerous characters exhibit extremely low frequency, relying solely on naturally occurring text is insufficient to ensure adequate exposure to these rare characters during model training.
To address this challenge and improve the robustness of text rendering across diverse contexts, we propose a multi-stage text-aware image synthesis pipeline. 
This pipeline integrates three complementary strategies: Pure Rendering, Compositional Rendering, and Complex Rendering. 
The details of each strategy are elaborated below.

\paragraph{Pure Rendering in Simple Backgrounds}
This strategy represents the most straightforward and effective method for training the model to recognize and generate characters (e.g., English and Chinese characters).
Text paragraphs are extracted from large-scale high-quality corpora and rendered onto clean backgrounds using dynamic layout algorithms that adapt font size and spacing based on canvas size. 
To ensure high-quality synthesized samples, a rigorous quality control mechanism is employed: if any character within a paragraph cannot be rendered due to limitations (e.g., font unavailability or rendering errors), the entire paragraph is discarded.
This strict filtering guarantees that only fully valid and legible samples are included in the training dataset, thereby maintaining high fidelity in character-level text rendering.

\paragraph{Compositional Rendering in Contextual Scenes}
This strategy focuses on embedding synthetic text into realistic visual contexts, mimicking its appearance in everyday environments. 
Text is simulated as being written or printed onto various physical media, such as paper or wooden boards, and then seamlessly composited into diverse background images to create visually coherent scenes. 
We employ the Qwen-VL Captioner to generate descriptive captions for each synthesized image, capturing contextual relationships between the text and its surrounding visual elements. 
This approach significantly improves the model's ability to comprehend and generate text within real-world scenarios.

\paragraph{Complex Rendering in Structured Templates}
To improve the model's capacity to follow complex, structured prompts involving layout-sensitive content, we propose a synthesis strategy based on programmatic editing of pre-defined templates, such as PowerPoint slides or User Interface Mockups. 
A comprehensive rule-based system is designed to automate the substitution of placeholder text while maintaining the integrity of layout structure, alignment, and formatting. 
These synthetic examples are crucial for helping model understand and execute detailed instructions that involve multi-line text rendering, precise spatial layouts, and control over text font and color.

In conclusion, our data synthesis framework systematically addresses the challenges associated with the scarcity and imbalance of textual content in natural image datasets. 
By integrating multiple rendering strategies that span simplicity, realism, and structural complexity, the framework synthesize comprehensive and diverse training data. 
Robustness across various text rendering tasks is achieved, thereby enhancing the model's ability to generate high-quality images that accurately follow complex user prompts about text rendering.

\section{Training}

\subsection{Pre-training}

We adopt a flow matching training objective to pre-train Qwen-Image, which facilitates stable learning dynamics via ordinary differential equations~(ODEs) while preserving equivalence to the maximum likelihood objective.
Formally, let $x_0$ denote the latent of the input image. 
The latent representation $z$ is obtained by encoding $x_0$ through the variational autoencoder~(VAE) encoder $\mathcal{E}$, i.e., $z = \mathcal{E}(x)$, where $\mathcal{E}: x \mapsto z$. 
Next, a random noise vector $x_1$ is sampled from the standard multivariate normal distribution, i.e., $x_1 \sim \mathcal{N}(0, \mathbf{I})$. 
For a user input $\boldsymbol{S}$, which may comprise text or prompt combined with image, the guidance latent $h$ is obtained from an MLLM $\phi$, i.e., $h = \phi(\boldsymbol{S})$, where $\phi: \mathcal{S} \mapsto h$. 
In addition, a diffusion timestep $t$ is sampled from a logit-normal distribution with $t \in [0, 1]$.
According to Rectified Flow~\citep{liu2022flow,esser2024scaling}, the intermediate latent variable at timestep $t$ and its corresponding velocity $v_t$ can be calculated as:
\begin{equation}
\left\{
\begin{aligned}
& x_t = tx_0 + (1-t)x_1 \\
& v_t = \frac{dx_t}{dt} = x_0 - x_1
\end{aligned}.
\right.
\label{equ:flow_match}
\end{equation}

Then, the model is trained to predict the target velocity, and the loss function is defined as the mean squared error~(MSE) between the predicted output $f_{\theta}(x_t, t)$ and the ground truth velocity $v_t$:
\begin{equation}
    \mathcal{L} = \mathbb{E}_{(x_0, h)\sim\mathcal{D}, x_1, t} \left\| v_{\theta}(x_t,t,h) - v_t \right\|^2,
\end{equation}
where $v_{\theta}(x_t,t,h)$ is velocity predicted by the model and $\mathcal{D}$ denotes the training dataset.

\subsubsection{Producer-Consumer Framework}
To ensure both high throughput and training stability when scaling to large-scale GPU clusters, we adopt a Ray~\citep{moritz2018ray}-inspired Producer–Consumer framework that decouples data preprocessing from model training.
This design enables both stages to operate asynchronously and at optimal efficiency, while also supporting on-the-fly updates to the data pipeline without interrupting the ongoing training process.
On the \textbf{Producer} side, raw image-caption pairs are first filtered according to our pre-defined criteria, such as image resolution and detection operators.
The selected data is then encoded into latent representations using MLLM models~(e.g., Qwen2.5 VL) and VAE.
Processed images are subsequently grouped by resolution in fast-access cache buckets and stored in a shared, location-aware store, allowing consumers to fetch them immediately without waiting in line.
The connection between the Producer and Consumer is achieved by employing a specific HTTP transport layer, which natively supports the RPC semantics required for asynchronous, zero-copy scheduling between the two endpoints.
The \textbf{Consumer} is deployed on GPU-dense clusters and is dedicated exclusively to model training.
By offloading all data processing to the Producer, the Consumer nodes can devote their entire compute budget to training the MMDiT model.
The MMDiT parameters are distributed across these nodes under a 4-way tensor-parallel layout, and every data-parallel group asynchronously pulls pre-processed batches directly from the Producer.
Additional consumer-side optimizations are detailed in the following section.

\subsubsection{Distributed Training Optimization}
Given the large parameter size of the Qwen-Image model, using FSDP~\citep{zhao2023pytorch} alone is insufficient to fit the model on each GPU. 
Therefore, we leverage Megatron-LM~\citep{shoeybi2019megatron,korthikanti2023reducing} for training and apply the following optimizations:

\paragraph{Hybrid Parallelism Strategy} We adopted a hybrid parallelism strategy, combining data parallelism and tensor parallelism, to efficiently scale training across large GPU clusters. Specifically, to implement tensor parallelism, we built the MMDiT model using the Transformer-Engine library\footnote{\url{https://github.com/NVIDIA/TransformerEngine}}, which allows seamless and automatic switching between different degrees of tensor parallelism. 
Furthermore, for the multi-head self-attention blocks, we employ the head-wise parallelism to reduce the synchronization and communication overhead compared to tensor parallelism along the head dimension.

\paragraph{Distributed Optimizer and Activation Checkpointing}
To alleviate GPU memory pressure with minimal recomputation overhead during backpropagation, we experimented with both distributed optimizers and activation checkpointing. However, activation checkpointing introduces substantial computational overhead in the backward pass, which can significantly degrade training speed. Through empirical comparison with the 256 multi-resolution image training setup, we observed that enabling activation checkpointing reduced per-GPU memory consumption by 11.3\%~(from 71GB to 63GB per GPU), but at the cost of increasing per-iteration time by 3.75$\times$~(from 2s to 7.5s per iteration). Based on this trade-off, we ultimately opted to disable activation checkpointing and rely solely on distributed optimizers. During training, all-gather operations are performed in bfloat16, while gradient reduce-scatter operations utilize float32, ensuring both computational efficiency and improved numerical stability.

\subsubsection{Training Strategy}
\label{subsubsec:train_strategy}
We adopt a multi-stage pre-training strategy aimed at progressively enhancing data quality, image resolution, and model performance. Throughout different training stages, we integrate various training strategies to optimize the learning process. These training strategies are listed below:

\paragraph{Enhancing Resolution: From Low Resolution to High Resolution}
This strategy progressively upscales the multi-resolution, multi-aspect ratio inputs, starting from an initial resolution of 256×256 pixels (with various aspect ratios including 1:1, 2:3, 3:2, 3:4, 4:3, 9:16, 16:9, 1:3 and 3:1), then increasing to 640×640 pixels, and finally reaching 1328×1328 pixels. By enhancing image resolution, the model can capture more detailed features, leading to better performance. Richer feature spaces facilitate improved generalization to unseen data. For example, transitioning from low-resolution to high-resolution flower images allows the model to discern finer details such as petal textures and color gradients.

\paragraph{Integrating Textual Rendering: From Non-text to Text}
To address the limited availability of textual content in conventional vision datasets and the consequent suboptimal glyph generation performance, particularly for Chinese characters, we progressively introduce images containing rendered text superimposed on natural backgrounds. This strategy enables the model to initially learn general visual representations and subsequently acquire text rendering capability.

\paragraph{Refining Data Quality: From Massive Data to Refined Data}
During the early stages of pre-training, we utilize large-scale datasets to enable the model to acquire fundamental visual generation capabilities. As training progresses, we gradually employ increasingly stringent data filtering mechanisms to select higher-quality data. This progressive data refinement ensures that only the most relevant and high-quality samples are leveraged to ensure the training efficiency and model performance.

\paragraph{Balancing Data Distribution: From Unbalanced to Balanced}
Throughout the training process, we progressively balance the dataset with respect to domain and image resolution distributions. This adjustment mitigates the risk of the model overfitting to particular domains or resolutions, which could otherwise compromise the fidelity and fine-grained details of generated images in underrepresented settings. By maintaining a more uniform data distribution, we promote robust generalization across diverse domains and resolutions.

\paragraph{Augmenting with Synthetic Data: From Real-World Data to Synthetic Data}
Certain data distributions, such as surrealistic styles or high-resolution images containing extensive textual content, are underrepresented or even absent in real-world datasets. Additionally, the availability of some high-quality data samples is inherently limited. To address these gaps, we employ data synthesis techniques to generate supplementary samples, thereby enriching the dataset and ensuring more comprehensive coverage of diverse visual domains. This augmentation strategy enhances the model's ability to generalize and perform robustly across a wider range of scenarios.

\subsection{Post-training}
In this section, we present the post-training framework for Qwen-Image, which consists of two stages: supervised fine-tuning~(SFT) and reinforcement learning~(RL)~\citep{kaelbling1996reinforcement}.

\subsubsection{Supervised Fine-Tuning (SFT)}
During the SFT stage, we construct a hierarchically organized dataset of semantic categories and employ meticulous human annotation to address specific shortcomings of the model. We require selected images to be clear, rich in detail, bright, and photorealistic. This approach is designed to guide the model towards producing content with greater realism and finer details.

\subsubsection{Reinforcement Learning~(RL)}

We employ two distinct RL strategies: Direct Preference Optimization~(DPO)~\citep{rafailov2023direct} and Group Relative Policy Optimization~(GRPO)~\citep{shao2024deepseekmath}.
DPO excels at flow-matching~(one step) online preference modeling and is computationally efficient, whereas GRPO performs on-policy sampling during training and evaluates each trajectory with a reward model.
To leverage the scalability advantages of offline preference learning, we conduct relative large-scale RL with DPO and reserve GRPO for small fine-grained RL refinement.
Details of both algorithms are provided below.

\subsubsubsection{\textbf{(A) Direct Preference Optimization~(DPO)}}

\paragraph{Data Preparation}
For DPO training data, given the same prompt, multiple images are generated with different random initialization seeds. Human annotators are then tasked with selecting the best and the worst images from these candidates. The data is divided into two categories: prompts associated with reference (gold) images, and prompts without reference images. For data with reference images, annotators first compare the generated outputs to the reference. If there is a significant discrepancy, the annotators are instructed to designate the worst generation as the rejected sample. For prompts without reference images, annotators are asked to select the best and worst samples among the generated images, or to indicate if all generated results are of unsatisfactory quality.

\paragraph{Algorithm}

Given the text hidden state $h$, chosen generated image~(or golden image) $x_0^{win}$ and rejected generated image $x_0^{lose}$, we sample timestep $t\sim (0, 1)$ to construct the input latent variable $x_t^{win}$ and $x_t^{lose}$ as well as their corresponding velocity $v_t^{win}$ and $v_t^{lose}$ following Equation~\ref{equ:flow_match}.
Then, inspired by previous work~\citep{wallace2024diffusion}, we construct the DPO objective based on the flow matching training criterion, which can be formulated as follows:

\begin{equation}
\label{equ:fm_dpo_full}
\left\{
\begin{aligned}
& \text{Diff}_{\text{policy}} = 
     \left( \big\| v_\theta(x_t^{win}, h, t) - v_t^{win} \big\|_2^2 - \big\| v_\theta(x_t^{lose}, h, t) - v_t^{lose} \big\|_2^2 \right) \\
& \text{Diff}_{\text{ref}} = 
     \left( \big\| v_{\text{ref}}(x_t^{win}, h, t) - v_t^{win} \big\|_2^2 - \big\| v_{\text{ref}}(x_t^{lose}, h, t) - v_t^{lose} \big\|_2^2 \right) \\
& \mathcal{L}_{DPO} = -\mathbb{E}_{h, (x_0^{win}, x_0^{lose}) \sim \mathcal{D},\, t \sim \mathcal{U}(0, 1)}
\Bigg[
      \log \sigma\Big( -\beta\, (\text{Diff}_{\text{policy}} - \text{Diff}_{\text{ref}}) \Big)
\Bigg],
\end{aligned}
\right.
\end{equation}
where $\text{Diff}_{\text{policy}}$ and $\text{Diff}_{\text{ref}}$ denote the preference differences computed by the policy model and the reference model, respectively, $\beta$ is a scaling parameter, and $\sigma(\cdot)$ denotes the sigmoid function.

\subsubsubsection{\textbf{(B) Group Relative Policy Optimization~(GRPO)}}

\paragraph{Algorithm}

After training with DPO, we perform further fine-grained training using GRPO following the Flow-GRPO~\citep{liu2025flowgrpo} framework.
Given text hidden state ${h}$, the flow model predicts a group of $G$ images $\{{x}_{0}^i\}_{i=1}^{G}$ and the corresponding trajectory $\{{x}_{T}^i,{x}_{T-1}^i,...,{x}_{0}^i\}_{i=1}^{G}$.
Within each group, the advantage function can be formulated as:
\begin{equation}
    A_{i}= \frac{R({x}_{0}^i,{h})-mean(\{R({x}_{0}^i,{h})\}_{i=1}^G)}{std(\{R({x}_{0}^i,{h})\}_{i=1}^G)}\mbox{,}
\end{equation}

where $R$ is the reward model. Then, the training objective of GRPO is:
\begin{equation}
\begin{split}
    \mathcal{L}_{\text{GRPO}}(\theta)=&\mathbb{E}_{{h}\sim\mathcal{D},\{{x}_{T}^i,...,{x}_{0}^i\}_{i=1}^{G}\sim \pi_{\theta}} \\
    &\frac{1}{G}\sum_{i=1}^G\frac{1}{T}\sum_{t=0}^{T-1}\left(\min(r_{t}^{i}(\theta)A_i,\text{clip}(r_{t}^{i}(\theta),1-\epsilon,1+\epsilon)A_i) - \beta D_{KL}(\pi_{\theta}||\pi_{\text{ref}}) \right),\label{loss_grpo}
\end{split}
\end{equation}
where $r_{t}^{i}(\theta) = \frac{p_{\theta}({x}_{t-1}^{i}|{x}_{t}^{i},{h})}{p_{\theta_{\text{old}}}({x}_{t-1}^{i}|{x}_{t}^{i},{h})}$. 

When sampling the trajectories $\{{x}_{T}^i,...,{x}_{0}^i\}_{i=1}^{G}\sim \pi_{\theta}$, we have $\text{d}{x}_t = {v}_t \text{d}t$ for flow-matching sampling~(following Eq.~\ref{equ:flow_match}), where ${v}_t = {v}_{\theta}({x}_t,t,h)$ is the predicted velocity. However, this sampling strategy exhibits no randomness, which is not suitable for exploration. Thus, we reformulate the sampling process as an SDE process for more randomness.
The SDE sampling process can be written as:
\begin{equation}
    \text{d}{x}_t = \left({v}_t+\frac{\sigma_t^2}{2t}({x}_t+(1-t){v}_t)\right) \text{d}t + \sigma_t\text{d}{w}\mbox{,}
\end{equation}
where $\sigma_t$ denotes the magnitude of randomness. Using Euler-Maruyama discretization, we have

\begin{equation}
    {x}_{t+\Delta t} = {x}_t + \left[{v}_\theta({x}_t,t,h)+\frac{\sigma_t^2}{2t}({x}_t+(1-t){v}_{\theta}({x}_t,t,h))\right]\Delta t+\sigma_t\sqrt{\Delta t} \epsilon\mbox{.}
\end{equation}

We use the above equation for sampling the trajectories. The KL-divergence in Eq.~\eqref{loss_grpo} can be solved in a closed form
\begin{equation}
    D_{KL}(\pi_{\theta}||\pi_{\text{ref}}) = \frac{\Delta t}{2} \left(\frac{\sigma_t(1-t)}{2t}+\frac{1}{\sigma_t}\right)^2 ||{v}_{\theta}({x}_t,t,h)-{v}_{\text{ref}}({x}_t,t,h)||^2\mbox{.}
\end{equation}

\begin{figure}[t]
\centering
\includegraphics[width=1\linewidth]{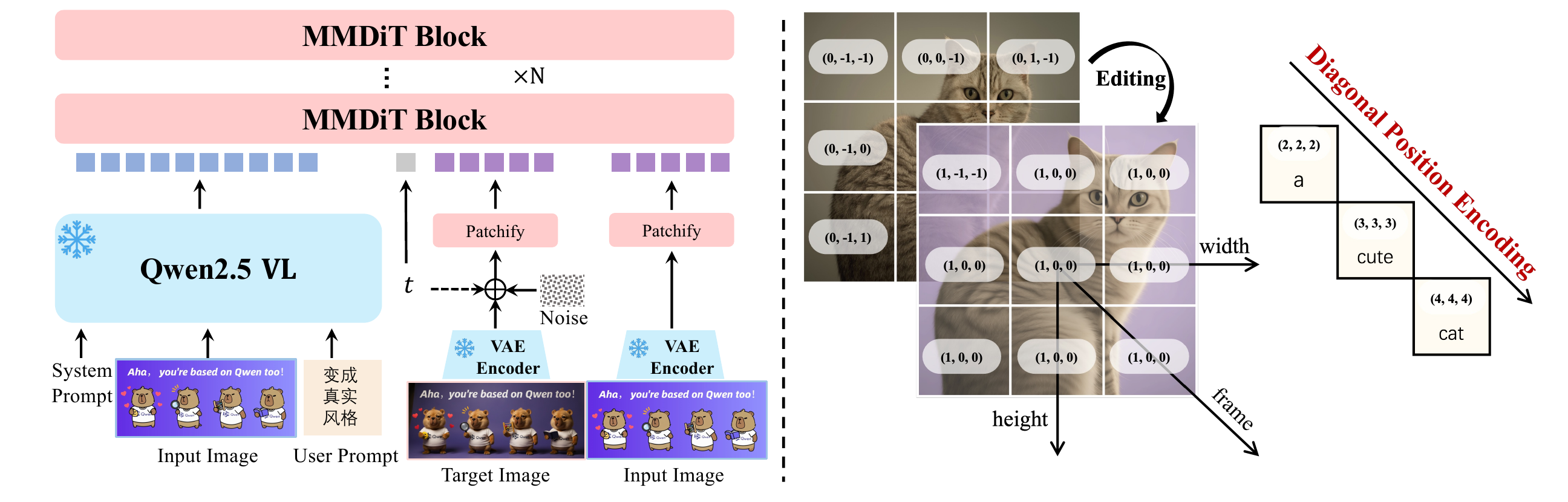}
\caption{Overview of the Image Editing (TI2I) task. Left: Illustration of the TI2I task training input format. The user prompt is ``Turn into realistic style'' in English. Right: The corresponding modification to MSRoPE for TI2I, where a new dimension~(frame) is introduced to distinguish between the images before and after editing.}
\label{fig:ti2i_task}
\end{figure}

\begin{figure}[t]
\begin{AcademicBox}[\footnotesize System Prompt for TI2I task]
<|im\_start|>system \\
Describe the key features of the input image (color, shape, size, texture, objects, background), then \\
explain how the user's text instruction should alter or modify the image. Generate a new image that \\
meets the user's requirements while maintaining consistency with the original input where appropriate. <|im\_end|> \\
<|im\_start|>user \\
<|vision\_start|>\textbf{\textcolor{red}{<|user\_image|>}}<|vision\_end|>\textbf{\textcolor{red}{<|user\_text|>}}<|im\_end|>\\
<|im\_start|>assistant
\end{AcademicBox}
\caption{System prompt for Image Editing~(TI2I) task, where \textbf{\textcolor{red}{<|user\_image|>}} is the user input image and \textbf{\textcolor{red}{<|user\_text|>}} is the user input prompt.}
\label{sys_prompt_ti2i}
\end{figure}

\subsection{Multi-task training}
In addition to text-to-image~(T2I) generation, we extend our base model to explore multi-modal image generation tasks that incorporate both text and image inputs, including instruction-based image editing~\citep{wang2025seededit}, novel view synthesis~\citep{wang2024crm}, and computer vision tasks such as depth estimation~\citep{bochkovskii2024depthpro}. 
We can broadly regard these as general image editing tasks.
Building on the capabilities of Qwen2.5-VL, our model natively supports image inputs: visual patches extracted from the user-provided image are encoded by a Vision Transformer (ViT) and concatenated with text tokens to form the input sequence. We design the system prompt shown in Figure~\ref{sys_prompt_ti2i} and extract both the input image and textual instructions as inputs to the text stream of Qwen-Image MMDiT.

Inspired by prior work~\citep{labs2025kontext}, which demonstrates that incorporating VAE embeddings helps maintain character and scene consistency, we additionally feed the VAE-encoded latent representation of the input image into the image stream, concatenating it with the noised image latent along the sequence dimension. To enable the model to distinguish between multiple images, we extend MSRoPE by introducing an additional frame dimension, in addition to the height and width used to locate image patches within a single image (see the right part of Figure~\ref{fig:ti2i_task}). Empirically, we find that providing the visual semantic embeddings from the MLLM enables better instruction following, while introducing pixel-level VAE embeddings further enhances the model's ability to preserve visual fidelity and maintain structural consistency with the user-provided image.

\section{Experiments}
\label{sec:exp}

\begin{figure}[t]
    \centering
  \centering
  \begin{subfigure}[b]{0.48\textwidth}
    \centering
    \includegraphics[width=\textwidth]{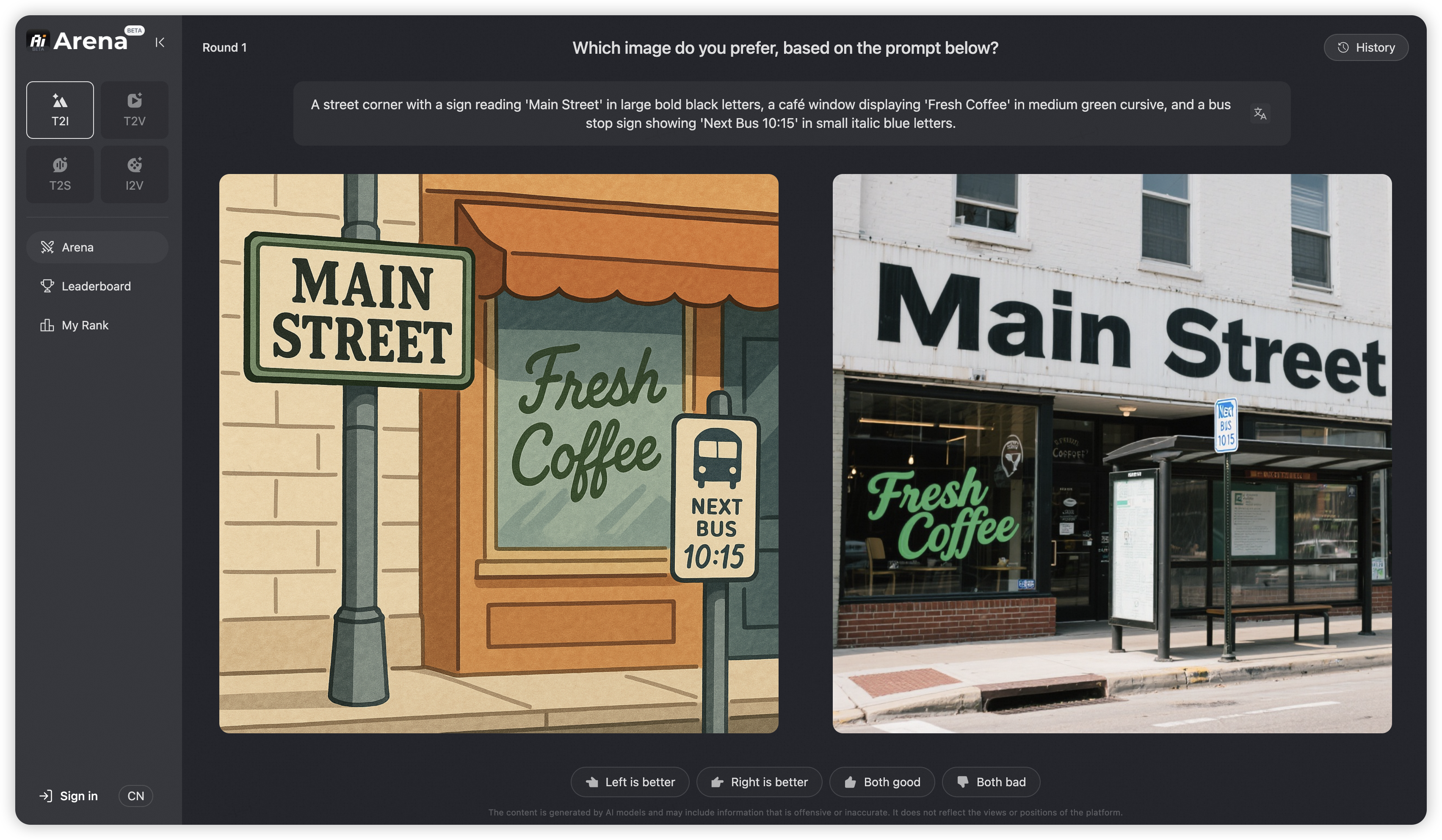}
    \caption{Front end of \href{https://aiarena.alibaba-inc.com}{AI Arena} platform}
    \label{fig:radar}
  \end{subfigure}
  \hspace{1em}
  \begin{subfigure}[b]{0.48\textwidth}
    \centering
    \includegraphics[width=\textwidth]{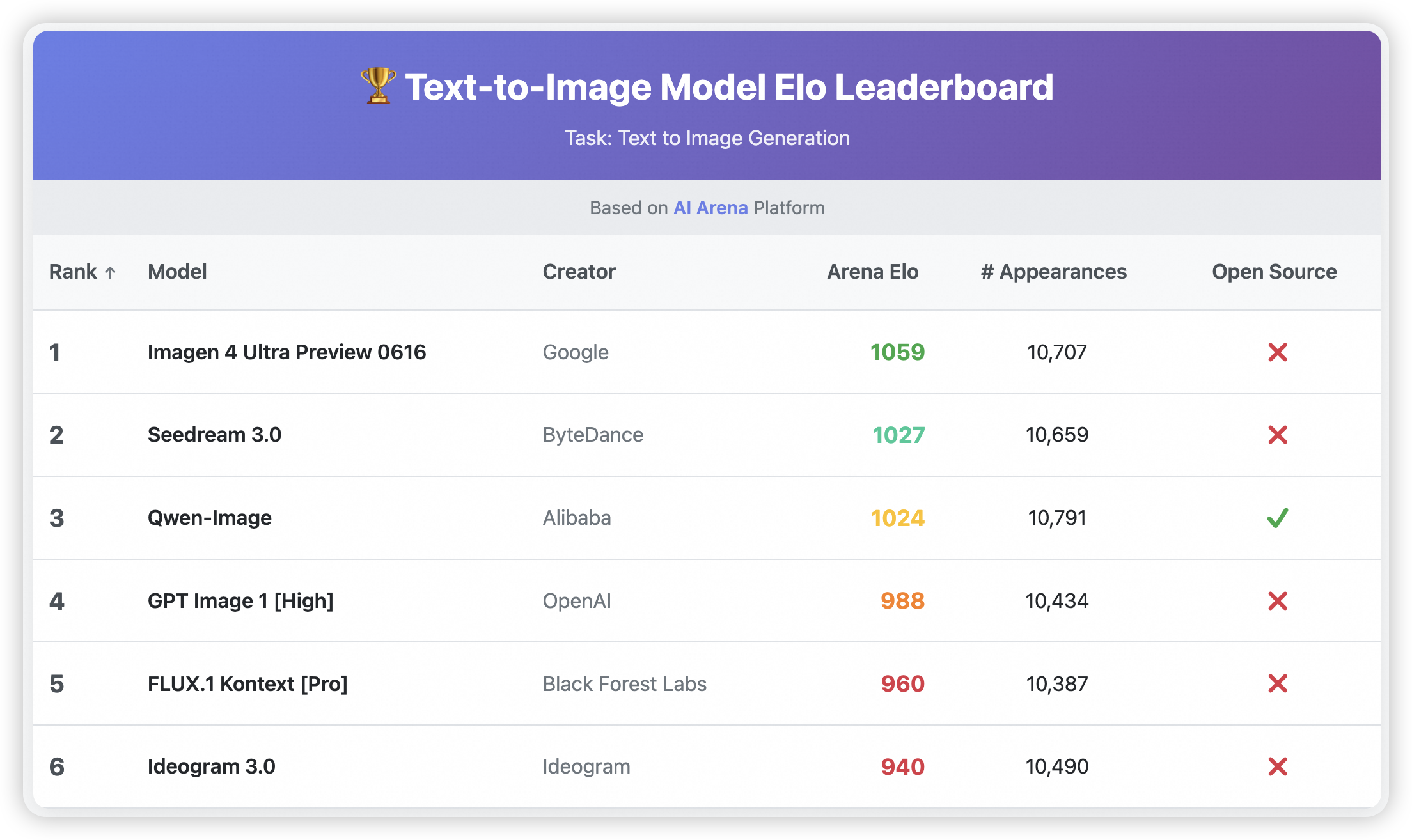}
    \caption{Text-to-Image \href{https://chat.qwen.ai/s/deploy/823627bf-7c07-4d88-824b-987d920f7bf3}{Elo Leaderboard}}
    \label{fig:text_radar}
  \end{subfigure}
  \vspace{-3mm}
    \caption{
        Comparison of Qwen-Image and leading closed-source APIs on the \href{https://aiarena.alibaba-inc.com}{AI Arena} platform.
        Users can compare images generated by two anonymous models based on the prompt and choose which one is better, both are good, or both are bad.
        \href{https://chat.qwen.ai/s/deploy/823627bf-7c07-4d88-824b-987d920f7bf3}{ELo Leaderboard} is powered by Qwen3-Coder~\citep{qwen3-coder} from \href{chat.qwen.ai}{chat.qwen.ai}.
    }
    \label{fig:aiarena_merge}
\end{figure}

\subsection{Human Evaluation}

To comprehensively evaluate the general image generation capabilities of Qwen-Image and objectively compare it with state-of-the-art closed-source APIs, we have developed AI Arena\footnote{\url{https://aiarena.alibaba-inc.com}}, an open benchmarking platform built upon the Elo rating system~\citep{elo1978rating}, as shown in Figure~\ref{fig:aiarena_merge}.

AI Arena serves as a fair and dynamic open competition platform. 
In each round, two images generated by randomly selected models using the same prompt are anonymously presented to users for pairwise comparison. 
Users vote for the superior image, and the results are used to update personal and global leaderboards through the Elo algorithm, allowing developers, researchers, and users to holistically assess the performance of models.
Specifically, we curated about 5,000 diverse prompts that span a wide range of evaluation dimensions, including subject, style, photographic perspective, and more. 
More than 200 evaluators from various professional backgrounds were invited to participate in the assessment process. 

AI Arena is now open to the public. Anyone can participate in the comparison of different models. 
In the future, the platform will further expand from text-to-image generation to various multimodal generation tasks such as image editing, text-to-audio, text-to-video, and image-to-video generation. 
The platform strictly adheres to the standards of objectivity and independence, and will detect and eliminate cheating or invalid data through various techniques.

We selected five state-of-the-art closed-source APIs as competitors for Qwen-Image in the arena: Imagen 4 Ultra Preview 0606~\citep{imagen4}, Seedream 3.0~\citep{gao2025seedream}, GPT Image 1 [High]~\citep{gptimage}, FLUX.1 Kontext [Pro]~\citep{labs2025kontext}, and Ideogram 3.0~\citep{ideogram3.0}. 
To date, each model has participated in at least 10,000 pairwise comparisons, ensuring robustness and fairness in the evaluation. 
Considering that most closed-source APIs do not reliably support Chinese text generation, we excluded prompts involving Chinese text to maintain objectivity in the comparative results.
                          
As shown in~\cref{fig:aiarena_merge}, Qwen-Image, as the only open-source image generation model, ranks third in the AI Arena. 
Although Qwen-Image trails the leading Imagen 4 Ultra Preview 0606 by \~30 Elo points, it demonstrates a significant advantage of over 30 Elo points compared to models such as GPT Image 1 [High] and FLUX.1 Kontext [Pro].
These results establish Qwen-Image as a powerful open-source image generation model, providing strong performance and broad utility for developers, researchers, and users.\looseness=-1

\subsection{Quantitative Results}
\label{sec:quantitative}
To comprehensively evaluate the visual generation capabilities of our model, we first report its performance on VAE reconstruction in~\cref{sec:vae_exp}, which serves to demonstrate the upper bound of the model's generation quality. We further conduct evaluations on two fundamental visual generation tasks, text-to-image (T2I) in ~\cref{sec:t2i_eval} and image editing (TI2I) in ~\cref{sec:ti2i_eval}, to provide a thorough assessment of the model's foundational generative abilities.

\subsubsection{Performance of VAE Reconstruction}
\label{sec:vae_exp}

We quantitatively evaluate several state-of-the-art image tokenizers, reporting Peak Signal-to-Noise Ratio (PSNR) and Structural Similarity Index Measure (SSIM) to assess reconstruction quality. All compared VAEs operate at an 8x8 compression rate using a latent channel dimension of 16. Notably, FLUX-VAE~\citep{flux2024}, Cosmos-CI-VAE~\citep{agarwal2025cosmos}, and SD-3.5-VAE~\citep{esser2024scaling} are image tokenizers, while Wan2.1-VAE~\citep{wan2025wan}, Hunyuan-VAE~\citep{kong2024hunyuanvideo}, and Qwen-Image-VAE function as joint image and video tokenizers. For a fair comparison, we report effective image parameters (see "Image Params" in Table~\ref{tab:table_vae}). This accounts for converting the 3D convolutions in joint models to equivalent 2D convolutions for image processing. Following prior art, evaluation is conducted on the ImageNet-1k~\citep{deng2009imagenet} validation set at 256x256 resolution for general domain performance. To further evaluate reconstruction capabilities on small texts, we also include reconstruction results on an in-house text-rich corpus covering diverse text sources (PDFs, PPT slides, posters, and synthetic texts) and languages. For improved numerical precision, we evaluate tokenizers using float32. As shown in Table~\ref{tab:table_vae}, Qwen-Image-VAE achieves state-of-the-art reconstruction performance across all evaluated metrics. Significantly, when processing images, Qwen-Image-VAE activates only 19M parameters in the encoder and 25M in the decoder, achieving an optimal balance between reconstruction quality and computational efficiency.

\begin{table}[t]
\centering
\caption{Quantitative Evaluation results of VAE.}
\scalebox{0.86}{
\begin{tabular}{l|cccc|cccc}
\toprule
\multirow{2}{*}{\textbf{Model}} & 
\multicolumn{2}{c}{\textbf{\# Params}} & 
\multicolumn{2}{c|}{\textbf{\# Image Params}} & 
\multicolumn{2}{c}{\textbf{Imagenet\_256x256}} & 
\multicolumn{2}{c}{\textbf{Text\_256x256}} \\
\cmidrule(lr){2-3} \cmidrule(lr){4-5} \cmidrule(lr){6-7} \cmidrule(lr){8-9}
& {Enc} & {Dec} & {Enc} & {Dec} & {PSNR($\uparrow$)} & {SSIM($\uparrow$)} & {PSNR($\uparrow$)} & {SSIM($\uparrow$)} \\
\midrule
Wan2.1-VAE \citep{wan2025wan}       & 54M & 73M & 19M & 25M & 31.29 & 0.8870 & 26.77 & 0.9386 \\
Hunyuan-VAE \citep{kong2024hunyuanvideo}   & 100M & 146M & 34M & 50M & 33.21 & 0.9143 & 32.83 & 0.9773  \\
FLUX-VAE \citep{flux2024}     & 34M & 50M & 34M & 50M & 32.84 & 0.9155 & 32.65 & 0.9792  \\
Cosmos-CI-VAE \citep{agarwal2025cosmos}   & 31M & 46M & 31M & 46M & 32.23 & 0.9010 & 30.62 & 0.9664  \\
SD-3.5-VAE \citep{esser2024scaling}   & 34M & 50M & 34M & 50M & 31.22 & 0.8839 & 29.93 & 0.9658 \\
\midrule
\bf Qwen-Image-VAE & 54M & 73M & 19M & 25M & \textbf{33.42} & \textbf{0.9159} & \textbf{36.63} & \textbf{0.9839}  \\
\bottomrule
\end{tabular}
}
~\label{tab:table_vae}
\end{table}

\subsubsection{Performance of Text-to-Image Generation}~\label{sec:t2i_eval}

We evaluate Qwen-Image’s performance on the text-to-image (T2I) task from two perspectives: general generation capability and text rendering capability. 
To assess the model's general generation performance, we conduct evaluations on four publicly available benchmarks — DPG~\citep{hu2024ella}, GenEval~\citep{ghosh2023geneval}, OneIG-Bench~\citep{chang2025oneig}, and TIIF~\citep{wei2025tiif}. 
These benchmarks provide a comprehensive measurement of the model’s ability to generate high-quality and semantically consistent images from text prompts. 
To further evaluate the model’s text rendering capability, we separately evaluate its performance on English and Chinese text generation. 
For English text rendering, we use the CVTG-2K~\citep{du2025textcrafter} benchmark, which is specifically designed to assess the readability of the rendered English text. 
To address the lack of standardized evaluation for Chinese text rendering, we introduce a new benchmark named ChineseWord, which evaluates the model’s ability to render Chinese characters, allowing us to systematically assess the model's text rendering performance. 
Furthermore, to fully evaluate Qwen-Image's ability to accurately render long texts, we conducted an evaluation on LongText-Bench~\citep{geng25xomni}, a benchmark designed to evaluate the performance on rendering longer texts in both English and Chinese.

\begin{table}[!h]
\centering
\caption{Quantitative evaluation results on DPG~\citep{hu2024ella}.}
\small
\begin{adjustbox}{width=\textwidth}
\begin{tabular}{l|ccccc|c}
\toprule
\textbf{Model}           & \bf Global & \bf Entity & \bf Attribute & \bf Relation & \bf Other & \bf Overall$\uparrow$ \\
\midrule
SD v1.5~\citep{rombach2021highresolution}           & 74.63  & 74.23  & 75.39     & 73.49    & 67.81 & 63.18    \\
PixArt-$\alpha$~\citep{chen2024pixartalpha}         & 74.97  & 79.32  & 78.60      & 82.57    & 76.96 & 71.11    \\
Lumina-Next~\citep{zhuo2024luminanext}      & 82.82  & 88.65  & 86.44     & 80.53    & 81.82 & 74.63    \\
SDXL~\citep{podell2023sdxl}             & 83.27  & 82.43  & 80.91     & 86.76    & 80.41 & 74.65    \\
Playground v2.5~\citep{li2024playground}  & 83.06  & 82.59  & 81.20      & 84.08    & 83.50  & 75.47    \\
Hunyuan-DiT~\citep{li2024hunyuandit}      & 84.59  & 80.59  & 88.01     & 74.36    & 86.41 & 78.87    \\
Janus~\citep{wu2025janus}            & 82.33  & 87.38  & 87.70      & 85.46    & 86.41 & 79.68    \\
PixArt-$\Sigma$~\citep{chen2024pixartsigma}         & 86.89  & 82.89  & 88.94     & 86.59    & 87.68 & 80.54    \\
Emu3-Gen~\citep{wang2024emu3}         & 85.21  & 86.68  & 86.84     & 90.22    & 83.15 & 80.60     \\
Janus-Pro-1B~\citep{chen2025janus}     & 87.58  & 88.63  & 88.17     & 88.98    & 88.30  & 82.63    \\
DALL-E 3~\citep{openai2023dalle3}         & 90.97  & 89.61  & 88.39     & 90.58    & 89.83 & 83.50     \\
FLUX.1 [Dev]~\citep{flux2024}       & 74.35  & 90.00     & 88.96     & 90.87    & 88.33 & 83.84    \\
SD3 Medium~\citep{esser2024scaling}       & 87.90   & 91.01  & 88.83     & 80.70     & 88.68 & 84.08    \\
Janus-Pro-7B~\citep{chen2025janus}     & 86.90   & 88.90   & 89.40      & 89.32    & 89.48 & 84.19    \\
HiDream-I1-Full~\citep{cai2025hidream}          & 76.44  & 90.22  & 89.48     & 93.74    & 91.83 & 85.89    \\
Lumina-Image 2.0~\citep{qin2025lumina} & -      & 91.97  & 90.20      & \textbf{94.85}    & -     & 87.20     \\
Seedream 3.0~\citep{gao2025seedream} &  \textbf{94.31}     & \textbf{92.65}  & 91.36      & 92.78    & 88.24     & 88.27     \\
GPT Image 1 [High]~\citep{gptimage} &  88.89     & 88.94  & 89.84      & 92.63    & 90.96     & 85.15     \\
\midrule
\bf Qwen-Image & 91.32 & 91.56 & \textbf{92.02} & 94.31 & \textbf{92.73} & \textbf{88.32} \\

\bottomrule
\end{tabular}\label{tab:dpg}
\end{adjustbox}
\end{table}

\paragraph{DPG} Table~\ref{tab:dpg} shows the performance comparison on \textbf{DPG}~\citep{hu2024ella}. This benchmark consists of 1K dense prompts, enabling fine-grained assessment of different aspects of prompt adherence. In general, Qwen-Image achieves the highest overall score, indicating its superior prompt-following capability. In particular, Qwen-Image excels at interpreting prompts involving attributes and other facets, outperforming all other models in the comparison.

\begin{table}[!h]\centering
\caption{Quantitative Evaluation results on GenEval~\citep{ghosh2023geneval}.}
\begin{adjustbox}{width=\textwidth}
\begin{tabular}{l|cccccc|c}
\toprule
\multirow{2}{*}{\textbf{Model}} & \textbf{Single} & \textbf{Two} & \multirow{2}{*}{\textbf{Counting}} & \multirow{2}{*}{\textbf{Colors}} & \multirow{2}{*}{\textbf{Position}} & \textbf{Attribute} & \multirow{2}{*}{\textbf{Overall$\uparrow$}} \\
& \bf Object & \bf Object & & & & \bf Binding & \\
\midrule
Show-o~\citep{xie2024show} & 0.95 & 0.52 & 0.49 & 0.82 & 0.11 & 0.28 &0.53 \\
Emu3-Gen~\citep{wang2024emu3} & 0.98 & 0.71 & 0.34 & 0.81 & 0.17 &0.21 & 0.54 \\
PixArt-$\alpha$~\citep{chen2024pixartalpha}         & 0.98          & 0.50        & 0.44     & 0.80    & 0.08     & 0.07              & 0.48     \\
SD3 Medium~\citep{esser2024scaling}      & 0.98          & 0.74       & 0.63     & 0.67   & 0.34     & 0.36              & 0.62     \\
FLUX.1 [Dev]~\citep{flux2024}   & 0.98 & 0.81 & 0.74  & 0.79 & 0.22   & 0.45 & 0.66     \\
SD3.5 Large~\citep{esser2024scaling}     & 0.98          & 0.89       & 0.73     & 0.83   & 0.34     & 0.47              & 0.71     \\
JanusFlow~\citep{ma2025janusflow} &0.97 & 0.59 & 0.45 & 0.83 & 0.53 & 0.42 & 0.63 \\
Lumina-Image 2.0~\citep{qin2025lumina} & - & 0.87 & 0.67 & -      & - & 0.62 & 0.73     \\
Janus-Pro-7B~\citep{chen2025janus}     & 0.99          & 0.89       & 0.59     & 0.90    & 0.79     & 0.66              & 0.80      \\
HiDream-I1-Full~\citep{cai2025hidream}          & \bf 1.00             & 0.98       & 0.79     & 0.91   & 0.60      & 0.72              & 0.83     \\
GPT Image 1 [High]~\citep{gptimage}           & 0.99          & 0.92       & 0.85     & 0.92   & 0.75     & 0.61              & 0.84     \\
Seedream 3.0~\citep{gao2025seedream} & 0.99 & \bf 0.96 & 0.91 & \bf 0.93 & 0.47 & 0.80 &0.84 \\
\midrule
\bf Qwen-Image  & 0.99 & 0.92 & 0.89 & 0.88 &  0.76 & 0.77 & \underline{0.87} \\
\bf Qwen-Image-RL   & \bf 1.00 & 0.95  & \bf 0.93  & 0.92 & \bf 0.87 & \bf 0.83 & \bf 0.91 \\
\bottomrule
\end{tabular}\label{tab:geneval}
\end{adjustbox}
\end{table}

\paragraph{GenEval} Table~\ref{tab:geneval} presents a comparison of model performance on the \textbf{GenEval}~\citep{ghosh2023geneval} benchmark, which focuses on object-centric text-to-image generation using compositional prompts with diverse object attributes.
We separately assess the performance of both the SFT model and the RL-enhanced model against other leading foundation models. Notably, our base model already surpasses the state of the art, outperforming Seedream 3.0~\citep{gao2025seedream} and GPT Image 1 [High]~\citep{gptimage}. Following reinforcement learning~(RL) fine-tuning, our model achieves an even higher score of 0.91, making it the only foundation model on the leaderboard to exceed the 0.9 threshold. 
These results demonstrate the superior controllable generation capabilities of Qwen-Image.

\begin{table}[!h]
    \centering
    \caption{Quantitative evaluation results on OneIG-EN~\citep{chang2025oneig}. The overall score is the average of the five dimensions.}
    \resizebox{0.99\linewidth}{!}{
    \begin{tabular}{l|ccccc|c}
        \toprule
        \textbf{Model} & \textbf{Alignment}& \textbf{Text} & \textbf{Reasoning} & \textbf{Style}& \textbf{Diversity} & \textbf{Overall}$\uparrow$ \\
        \midrule
        Janus-Pro~\citep{chen2025janus} & 0.553  & 0.001  &   0.139     & 0.276 & 0.365 & 0.267\\
        BLIP3-o~\citep{chen2025blip3} & 0.711  & 0.013  &   0.223      & 0.361 & 0.229 & 0.307\\
        BAGEL~\citep{deng2025bagel} & 0.769  & 0.244  &   0.173    & 0.367 & 0.251& 0.361\\
        BAGEL+CoT~\citep{deng2025bagel} & 0.793 & 0.020  &   0.206    & 0.390 & 0.209 & 0.324\\
        Show-o2-1.5B~\citep{xie2025show} & 0.798 & 0.002 & 0.219& 0.317 & 0.186 & 0.304\\
        Show-o2-7B~\citep{xie2025show} & 0.817 & 0.002 & 0.226 & 0.317 & 0.177&0.308\\
        OmniGen2~\citep{wu2025omnigen2} & 0.804 & 0.680 & 0.271 & 0.377 & 0.242 &0.475\\       
        SD 1.5~\citep{rombach2021highresolution} & 0.565 & 0.010 & 0.207 & 0.383 & \textbf{0.429} &0.319\\
        SDXL~\citep{podell2023sdxl} & 0.688 & 0.029 & 0.237 & 0.332 & 0.296 &0.316\\
        SD3.5 Large~\citep{esser2024scaling} & 0.809 & 0.629 & 0.294 & 0.353 & 0.225&0.462 \\
        FLUX.1 [Dev]~\citep{flux2024} & 0.786 & 0.523 & 0.253 & 0.368 & 0.238 &0.434\\
        CogView4~\citep{Cogview4} & 0.786 & 0.641 & 0.246 & 0.353 & 0.205 &0.446\\
        SANA-1.5 1.6B (PAG)~\citep{xie2025sana1d5} & 0.762 & 0.054 & 0.209 & 0.387 & 0.222 &0.327\\
        SANA-1.5 4.8B (PAG)~\citep{xie2025sana1d5} & 0.765 & 0.069 & 0.217 & 0.401 & 0.216 &0.334\\
        Lumina-Image 2.0~\citep{qin2025lumina} & 0.819 & 0.106 & 0.270 & 0.354 & 0.216 & 0.353\\
        HiDream-I1-Full~\citep{cai2025hidream} & 0.829 & 0.707 & 0.317 & 0.347 & 0.186 &0.477\\
        Imagen3~\citep{Imagen3} & 0.843 & 0.343 & 0.313 & 0.359 & 0.188 &0.409\\
        Recraft V3~\citep{recraftv3} & 0.810 & 0.795 & 0.323 & 0.378 & 0.205 &0.502\\
        Kolors 2.0~\citep{Kolors2} & 0.820 & 0.427 & 0.262 & 0.360 & 0.300 &0.434\\
        Seedream 3.0~\citep{gao2025seedream} & 0.818 & 0.865 & 0.275 & 0.413 & 0.277 &0.530\\
        Imagen4~\citep{imagen4} & 0.857 & 0.805 & 0.338 & 0.377 & 0.199&0.515\\
        GPT Image 1 [High]~\citep{gptimage} & 0.851 & 0.857 & \textbf{0.345} & \textbf{0.462} & 0.151& 0.533\\
        \midrule
        \bf Qwen-Image & \textbf{0.882} & \textbf{0.891} & 0.306 & 0.418 & 0.197 & \textbf{0.539} \\
        \bottomrule
    \end{tabular}
    }
    \label{tab:oneig_en}
\end{table}

\begin{table}[!h]
    \centering
    \caption{Quantitative evaluation results on OneIG-ZH~\citep{chang2025oneig}. The overall score is the average of the five dimensions.}
    \resizebox{0.99\linewidth}{!}{
    \begin{tabular}{l|ccccc|c}
        \toprule
        \textbf{Model} &
        \textbf{Alignment} & 
        \textbf{Text} & 
        \textbf{Reasoning} & 
        \textbf{Style} & 
        \textbf{Diversity} &
        \textbf{Overall}$\uparrow$\\
        \midrule
        Janus-Pro~\citep{chen2025janus} & 0.324          & 0.148          & 0.104        & 0.264          & \textbf{0.358} & 0.240 \\
        BLIP3-o~\citep{chen2025blip3} & 0.608          & 0.092          & 0.213        & 0.369          & 0.233          & 0.303\\
        BAGEL~\citep{deng2025bagel} & 0.672          & 0.365          & 0.186 & 0.357          & 0.268          & 0.370\\
        BAGEL+CoT~\citep{deng2025bagel} & 0.719   & 0.127  &   0.219    & 0.385 & 0.197& 0.329\\
        Cogview4~\citep{Cogview4} & 0.700          & 0.193          & 0.236        & 0.348          & 0.214          & 0.338\\
        Lumina-Image 2.0~\citep{qin2025lumina} & 0.731          & 0.136          & 0.221        & 0.343          & 0.240           & 0.334\\
        HiDream-I1-Full~\citep{cai2025hidream} & 0.620          & 0.205          & 0.256        & 0.304          & 0.300            & 0.337\\
        Kolors 2.0~\citep{Kolors2} & 0.738          & 0.502          & 0.226        & 0.331          & 0.333        & 0.426 \\
        Seedream 3.0~\citep{gao2025seedream} & 0.793          & 0.928 & 0.281      & 0.397          & 0.243          & 0.528\\
        GPT Image 1 [High]~\citep{gptimage} & 0.812 & 0.650           & \textbf{0.300} & \textbf{0.449} & 0.159     & 0.474 \\
          \midrule
        \bf Qwen-Image & \textbf{0.825} & \textbf{0.963} & 0.267 & 0.405 & 0.279 & \textbf{0.548} \\
        \bottomrule
    \end{tabular}
    }
    \label{tab:oneig_zh}
\end{table}

\paragraph{OneIG-Bench} Table.~\ref{tab:oneig_en} and Table.~\ref{tab:oneig_zh} report the quantitative results on OneIG-Bench~\citep{chang2025oneig}, a comprehensive benchmark designed for fine-grained evaluation of T2I models across multiple dimensions. For a fair overall comparison, we average the scores across all dimensions to obtain the final overall score. In general, Qwen-Image achieves the highest overall score on both the Chinese and English tracks, demonstrating its strong general-purpose generation capability. Notably, it ranks first in the Alignment and Text categories, evidencing its superior prompt following and text rendering capabilities.

\begin{table}[!h]\centering
\caption{Quantitative evaluation results on TIIF Bench testmini~\citep{wei2025tiif}. The best result is in bold and the second best result is underlined.}
\renewcommand{\arraystretch}{1.7} 
\setlength{\tabcolsep}{3pt}

\centering
\begin{adjustbox}{width=\textwidth}
\begin{tabular}{l|cc|cccccccc|cccccccccccc|cc}
\toprule
\multirow{3}{*}{\textbf{Model}}
  & \multicolumn{2}{c|}{\multirow{2}{*}{\textbf{Overall}}}
  & \multicolumn{8}{c|}{\textbf{Basic Following}}
  & \multicolumn{12}{c|}{\textbf{Advanced Following}}
  & \multicolumn{2}{c}{\textbf{Designer}} \\

\cmidrule(lr){4-11} \cmidrule(lr){12-23} \cmidrule(lr){24-25}

& & &
  \multicolumn{2}{c}{Avg}                    
  & \multicolumn{2}{c}{Attribute}
  & \multicolumn{2}{c}{Relation}
  & \multicolumn{2}{c|}{Reasoning}
  & \multicolumn{2}{c}{Avg}                  
  & \multicolumn{2}{c}{\makecell{Attribute\\+Relation}}
  & \multicolumn{2}{c}{\makecell{Attribute\\+Reasoning}}
  & \multicolumn{2}{c}{\makecell{Relation\\+Reasoning}}
  & \multicolumn{2}{c}{Style}
  & \multicolumn{2}{c|}{Text}
  & \multicolumn{2}{c}{\makecell{Real\\World}} \\

& short & long &          
  short & long &          
  short & long &          
  short & long &          
  short & long &          
  short & long &          
  short & long &          
  short & long &          
  short & long &          
  short & long &          
  short & long &          
  short & long            
\\
\midrule

FLUX.1 [dev]~\citep{flux2024}  &{{71.09}} &71.78 &83.12 &78.65& 87.05 & 83.17 & 87.25 &80.39 &75.01 &72.39 &65.79 &68.54& 67.07 &73.69 &73.84 &73.34 &69.09 &71.59 & 66.67 & 66.67 &43.83 &52.83 &70.72 &71.47 \\
FLUX.1 [Pro]~\citep{flux2024} &67.32 &69.89 &79.08 &78.91 &78.83 &81.33 &82.82 &83.82 &75.57 &71.57 &61.10 &65.37 &62.32 &65.57 &69.84 &71.47 &65.96 &67.72 &63.00 &63.00 &35.83 &55.83 &71.80 &68.80 \\
DALL-E 3~\citep{openai2023dalle3} &74.96 &70.81 &78.72 &78.50 &79.50 &79.83 &80.82 &78.82 &75.82 &76.82 &73.39 &67.27 &73.45 &67.20 &72.01 &71.34 &63.59 &60.72 &89.66 &86.67 &66.83 &54.83 &72.93 &60.99 \\
SD 3~\citep{esser2024scaling}            &67.46 &66.09 &78.32 &77.75 &83.33 &79.83 &82.07 &78.82 &71.07 &74.07 &61.46 &59.56 &61.07 &64.07 &68.84 &70.34 &50.96 &57.84 &66.67 &76.67 &59.83 &20.83 &63.23 &67.34 \\
PixArt-$\Sigma$~\citep{chen2024pixartsigma} &62.00 &58.12 &70.66 &75.25 &69.33 &78.83 &75.07 &77.32 &67.57 &69.57 &57.65 &49.50 &65.20 &56.57 &66.96 &61.72 &66.59 &54.59 &83.33 &70.00 &1.83 &1.83 &62.11 &52.41 \\
Lumina-Next~\citep{zhuo2024luminanext} &50.93 &52.46 &64.58 &66.08 &56.83 &59.33 &67.57 &71.82 &69.32 &67.07 &44.75 &45.63 &51.44 &43.20 &51.09 &59.72 &44.72 &54.46 &70.00 &66.67 &0.00 &0.83 &47.56 &49.05 \\
Hunyuan-DiT~\citep{li2024hunyuandit} &51.38 &53.28 &69.33 &69.00 &65.83 &69.83 &78.07 &73.82 &64.07 &63.32 &42.62 &45.45 &50.20 &41.57 &59.22 &61.84 &47.84 &51.09 &56.67 &73.33 &0.00 &0.83 &40.10 &44.20 \\
Show-o~\citep{xie2024show} &59.72 &58.86 &73.08 &75.83 &74.83 &79.83 &78.82 &78.32 &65.57 &69.32 &53.67 &50.38 &60.95 &56.82 &68.59 &68.96 &66.46 &56.22 &63.33 &66.67 &3.83 &2.83 &55.02 &50.92 \\
LightGen~\citep{wu2025lightgen} &53.22 &43.41 &66.58 &47.91 &55.83 &47.33 &74.82 &45.82 &69.07 &50.57 &46.74 &41.53 &62.44 &40.82 &61.71 &50.47 &50.34 &45.34 &53.33 &53.33 &0.00 &6.83 &50.92 &50.55 \\
SANA 1.5~\citep{xie2025sana1d5}   &67.15 &65.73 &79.66 &77.08 &79.83 &77.83 &85.57 &83.57 &73.57 &69.82 &61.50 &60.67 &65.32 &56.57 &69.96 &73.09 &62.96 &65.84 &80.00 &80.00 &17.83 &15.83 &71.07 &68.83 \\
Infinity~\citep{han2025infinity} &62.07 &62.32 &73.08 &75.41 &74.33 &76.83 &72.82 &77.57 &72.07 &71.82 &56.64 &54.98 &60.44 &55.57 &74.22 &64.71 &60.22 &59.71 &80.00 &73.33 &10.83 &23.83 &54.28 &56.89 \\
Janus-Pro-7B~\citep{chen2025janus} &66.50 &65.02 &79.33 &78.25 &79.33 &82.33 &78.32 &73.32 &80.32 &79.07 &59.71 &58.82 &66.07 &56.20 &70.46 &70.84 &67.22 &59.97 &60.00 &70.00 &28.83 &33.83 &65.84 &60.25 \\
MidJourney v7~\citep{midjourneyv7} &68.74 &65.69 &77.41 &76.00 &77.58 &81.83 &82.07 &76.82 &72.57 &69.32 &64.66 &60.53 &67.20 &62.70 &\underline{81.22} &71.59 &60.72 &64.59 &83.33 &80.00 &24.83 &20.83 &68.83 &63.61 \\
Seedream 3.0~\citep{gao2025seedream} & 86.02 & 84.31 & \underline{87.07} & 84.93 & \underline{90.50} & \underline{90.00} & \textbf{89.85} &\underline{ 85.94} & \underline{80.86} &78.86 & 79.16 &80.60 & \underline{79.76} & \underline{81.82} & 77.23 & 78.85 & \underline{75.64} &\underline{78.64} & \textbf{100.00} & \underline{93.33} & \textbf{97.17} &\underline{87.78} & 83.21 &83.58 \\
GPT Image 1 [High]~\citep{gptimage} &\textbf{89.15} &\textbf{88.29} &\textbf{90.75} &\textbf{89.66} &\textbf{91.33} &87.08 &84.57&84.57 &\textbf{96.32} &\textbf{97.32} &\textbf{88.55} &\textbf{88.35} &\textbf{87.07} &\textbf{89.44} &\textbf{87.22} &\textbf{ 83.96} &\textbf{85.59} &\textbf{83.21} & \underline{90.00} & \underline{93.33} &89.83 &86.83 &\underline{89.73} &\textbf{93.46} \\
\midrule
\bf Qwen-Image &\underline{86.14} &\underline{86.83} & 86.18 & \underline{87.22} & \underline{90.50} & \textbf{91.50} & \underline{88.22} & \textbf{90.78} & 79.81 & \underline{79.38} & \underline{79.30} & \underline{80.88} & 79.21 &78.94 & 78.85 &\underline{81.69} & 75.57 &78.59 & \textbf{100.00}  &\textbf{100.00} & \underline{92.76} & \textbf{89.14} & \textbf{90.30} &\underline{91.42} \\
\bottomrule
\end{tabular}\label{tab:tiif}
\end{adjustbox}
\end{table}

\paragraph{TIIF} Table~\ref{tab:tiif} shows the performance comparison on \textbf{TIIF Bench mini}~\citep{wei2025tiif}, a benchmark designed to systematically evaluate T2I model's ability to interpret and follow intricate textual instructions. Overall, Qwen-Image ranks second, surpassed only by GPT Image 1~\citep{gptimage}, underscoring its strong instruction-following capabilities.

\begin{table}[!h]
\centering
\caption{Quantitative evaluation results of English text rendering on CVTG-2K~\citep{du2025textcrafter}.}
\begin{adjustbox}{width=\textwidth}
\begin{tabular}{l|ccccc|c|c}
\toprule
\multirow{2}{*}{\textbf{Model}} & \multicolumn{5}{c|}{\bf Word Accuracy$\uparrow$} & \multirow{2}{*}{\bf NED$\uparrow$} & \multirow{2}{*}{\bf CLIPScore$\uparrow$} \\
\cmidrule{2-6}
 & 2 regions & 3 regions & 4 regions & 5 regions & average & & \\ 
\midrule
SD3.5 Large~\citep{esser2024scaling} & 0.7293 & 0.6825 & 0.6574 & 0.5940 & 0.6548 & 0.8470 & 0.7797 \\
FLUX.1 [dev]~\citep{flux2024} & 0.6089 & 0.5531 & 0.4661 & 0.4316 & 0.4965 & 0.6879 & 0.7401 \\
AnyText~\citep{tuo2024anytext} & 0.0513 & 0.1739 & 0.1948 & 0.2249 & 0.1804 & 0.4675 & 0.7432 \\
TextDiffuser-2~\citep{chen2024textdiffuser2} & 0.5322 & 0.3255 & 0.1787 & 0.0809 & 0.2326 & 0.4353 & 0.6765 \\
RAG-Diffusion~\citep{chen2024ragdiffusion} & 0.4388 & 0.3316 & 0.2116 & 0.1910 & 0.2648 & 0.4498 & 0.7797 \\
3DIS~\citep{zhou20243dis} & 0.4495 & 0.3959 & 0.3880 & 0.3303 & 0.3813 & 0.6505 & 0.7767 \\
TextCrafter~\citep{du2025textcrafter} & 0.7628 & 0.7628& 0.7406 &0.6977& 0.7370& 0.8679& 0.7868\\  
Seedream 3.0~\citep{gao2025seedream} & 0.6282&0.5962&0.6043&0.5610&0.5924&0.8537& 0.7821\\
GPT Image 1 [High]~\citep{gptimage} &\textbf{0.8779} &\textbf{0.8659}&\textbf{0.8731}&\textbf{0.8218}&\textbf{0.8569}&\textbf{0.9478}& \underline{0.7982}\\
\midrule
\bf Qwen-Image &\underline{0.8370} &\underline{0.8364}&\underline{0.8313}&\underline{0.8158}&\underline{0.8288}& \underline{0.9116}& \textbf{0.8017}\\
\bottomrule
\end{tabular} \label{tab:text_crafter}
\end{adjustbox}
\end{table}

\paragraph{CVTG-2K} Table~\ref{tab:text_crafter} reports the quantitative results of English rendering on \textbf{CVTG-2K}~\citep{du2025textcrafter}. This benchmark contains 2K prompts, each requiring 2–5 regions of English to be rendered on the generated image. Three metrics: Word Accuracy, NED, CLIPScore are introduced to measure the accuracy of text rendering. As shown in the table, Qwen-Image achieves performance comparable to that of state-of-the-art image generation models, underscoring its powerful English text rendering capability.

\begin{table}[!h]
\centering
\caption{Quantitative comparison results of Chinese text rendering.}
    \begin{tabular}{l|ccc|c}
    \toprule
    \textbf{Model} & \bf Level-1 Acc & \bf Level-2 Acc & \bf Level-3 Acc & \textbf{Overall}$\uparrow$ \\
    \midrule
    Seedream 3.0~\citep{gao2025seedream} & 53.48 & \underline{26.23} &  1.25  & 33.05 \\
    GPT Image 1 [High]~\citep{gptimage} & \underline{68.37} & 15.97 & \underline{3.55} & \underline{36.14} \\
    \midrule
    \bf Qwen-Image & \textbf{97.29} & \textbf{40.53} & \textbf{6.48} & \textbf{58.30} \\
    \bottomrule
    \end{tabular}\label{tab:cw}
\end{table}

\paragraph{ChineseWord} Table~\ref{tab:cw} reports the quantitative results on \textbf{ChineseWord}, our new benchmark for character-level Chinese text rendering. In accordance with the List of Commonly Used Standard Chinese Characters, we group the characters into three difficulty tiers: Level-1 (3500 characters), Level-2 (3000 characters) and Level-3 (1605 characters). We craft several prompt templates that instruct text-to-image models to generate an image containing a single Chinese character. Across all three tiers, Qwen-Image attains the highest rendering accuracy, underscoring its superior ability to render Chinese text.

\begin{table}[t]
\caption{Quantitative evaluation results on LongText-Bench~\citep{geng25xomni}. The best result is in bold and the second best result is underlined.}
  \centering
  \small
  \begin{tabular}{l|cc}
    \toprule
    \textbf{Model} & \bf LongText-Bench-EN & \bf LongText-Bench-ZH \\
    \midrule
    Janus-Pro~\citep{chen2025janus}           & 0.019 & 0.006 \\
    BLIP3-o~\citep{chen2025blip3}               & 0.021 & 0.018 \\
    Kolors 2.0~\citep{Kolors2}            & 0.258 & 0.329 \\
    BAGEL~\citep{deng2025bagel}                  & 0.373 & 0.310 \\
    OmniGen2~\citep{wu2025omnigen2}            & 0.561 & 0.059 \\
    X-Omni~\citep{geng25xomni}                     & 0.900 & 0.814 \\
    HiDream-I1-Full~\citep{cai2025hidream}      & 0.543 & 0.024 \\
    FLUX.1 [Dev]~\citep{flux2024}          & 0.607 & 0.005 \\
    Seedream 3.0~\citep{gao2025seedream}        & 0.896 & \underline{0.878} \\
    GPT Image 1 [High]~\citep{gptimage}                 & \textbf{0.956} & 0.619 \\
    \midrule
    \bf Qwen-Image & \underline{0.943} & \textbf{0.946} \\
    \bottomrule
  \end{tabular}
  \label{tab:longtext_bench}
\end{table}

\paragraph{LongText-Bench} Table~\ref{tab:longtext_bench} reports the quantitative results on \textbf{LongText-Bench}~\citep{geng25xomni}, a benchmark specifically designed to assess a model's ability to precisely render lengthy texts. The dataset contains 160 prompts spanning eight distinct scenarios. As shown in the table, Qwen-Image attains the highest accuracy on long Chinese text and the second-highest accuracy on long English text, illustrating Qwen-Image's superior long text rendering capability. 

\subsubsection{Performance of Image Editing}~\label{sec:ti2i_eval}

We further train a multi-task version of Qwen-Image for image editing (TI2I) tasks, seamlessly integrating both text and image as conditioning inputs. We evaluate our model across two categories of TI2I tasks:
First, for general-purpose image editing, we assess the instruction-based editing capability of our model on the GEdit~\citep{liu2025step1x} and ImgEdit~\citep{ye2025imgedit} benchmarks. These benchmarks test the model’s ability to perform open-ended edits based on textual and visual instructions.
Second, to evaluate the model's spatial understanding and generative capability in 3D vision tasks, we benchmark its performance on novel view synthesis~\citep{downs2022googlescannedobjectshighquality} and depth estimation~\citep{bochkovskii2024depthpro}. These tasks require the model to infer and generate coherent spatial information conditioned on input images and corresponding textual descriptions.
All these tasks can be unified within the TI2I scope, showcasing the general applicability of our approach to diverse multimodal tasks.

\begin{table*}[!h]
    \centering
    \small
    \caption{Comparison of Semantic Consistency (G\_SC), Perceptual Quality (G\_PQ), and Overall Score (G\_O) on GEdit-Bench. All metrics are evaluated by GPT-4.1. G\_O is computed as the geometric mean of G\_SC and G\_PQ, averaged over all samples. Note: FLUX.1 Kontext [Pro] underperforms on GEdit-Bench-CN due to its limited Chinese language capability.}
    \begin{tabular}{l|ccc|ccc}
    \toprule
  \multirow{2}{*}{\bf Model} & \multicolumn{3}{c|}{\bf GEdit-Bench-EN (Full set)$\uparrow$} & \multicolumn{3}{c}{\bf GEdit-Bench-CN (Full set)$\uparrow$} \\
    \cmidrule{2-4} \cmidrule{5-7}
    & G\_SC & G\_PQ & G\_O & G\_SC & G\_PQ & G\_O \\
    \midrule
    Instruct-Pix2Pix \citep{brooks2023instructpix2pix} & 3.58 & 5.49 & 3.68 & - & - & - \\
    AnyEdit~\citep{yu2025anyedit} & 3.18 & 5.82 & 3.21 & - & - & - \\
    MagicBrush~\citep{zhang2023magicbrush} & 4.68 & 5.66 & 4.52 & - & - & - \\
    UniWorld-v1~\citep{lin2025uniworldv1} & 4.93 & 7.43 & 4.85 & - & - & - \\
    OmniGen~\citep{xiao2025omnigen} & 5.96 & 5.89 & 5.06 & - & - & - \\
    OmniGen2~\citep{wu2025omnigen2} & 7.16 & 6.77 & 6.41 & - & - & - \\
    Gemini 2.0~\citep{googleGemini2} & 6.73 & 6.61 & 6.32 & 5.43 & 6.78 & 5.36 \\
    BAGEL~\citep{deng2025bagel} & 7.36 & 6.83 & 6.52 & 7.34 & 6.85 & 6.50 \\
    FLUX.1 Kontext [Pro]~\citep{labs2025kontext} & 7.02 & 7.60 & 6.56 & 1.11 & 7.36 & 1.23 \\
    Step1X-Edit~\citep{liu2025step1x} & 7.66 & 7.35 & 6.97 & 7.20 & 6.87 & 6.86 \\
    GPT Image 1 [High]~\citep{gptimage} & \underline{7.85} & \underline{7.62} & \underline{7.53} & \underline{7.67} & \underline{7.56} & \underline{7.30} \\
    \midrule
    \bf Qwen-Image & \textbf{8.00} & \textbf{7.86} & \textbf{7.56} & \textbf{7.82} & \textbf{7.79} & \textbf{7.52} \\
    \bottomrule
    \end{tabular}
\label{tab:gedit}
\end{table*}

\paragraph{GEdit} Table~\ref{tab:gedit} reports the results on the \textbf{GEdit-Bench}~\citep{liu2025step1x}, which evaluates image editing models on real-world user instructions across 11 diverse categories. We adopt three metrics—Semantic Consistency(SQ), Perceptual Quality~(PQ), and Overall Score~(O)—each ranging from 0 to 10. Qwen-Image ranks at the top of both the English and Chinese leaderboards, demonstrating strong editing capability and generalization to multilingual user instructions.

\begin{table}[!t]
    \centering
    \scriptsize
    \caption{Quantitative comparison results on ImgEdit~\citep{ye2025imgedit}. All metrics are evaluated by GPT-4.1. “Overall” is calculated by averaging all scores across tasks.}
    \resizebox{0.99\linewidth}{!}{
    \begin{tabular}{l|ccccccccc|c}
        \toprule
        \textbf{Model} & \bf Add & \bf Adjust & \bf Extract & \bf Replace & \bf Remove & \bf Background & \bf Style & \bf Hybrid & \bf Action & \bf Overall $\uparrow$ \\
        \midrule
        MagicBrush~\citep{zhang2023magicbrush} & 2.84 & 1.58 & 1.51 & 1.97 & 1.58 & 1.75 & 2.38 & 1.62 & 1.22 & 1.90 \\
        Instruct-Pix2Pix \citep{brooks2023instructpix2pix} & 2.45 & 1.83 & 1.44 & 2.01 & 1.50 & 1.44 & 3.55 & 1.20 & 1.46 & 1.88 \\
        AnyEdit~\citep{yu2025anyedit} & 3.18 & 2.95 & 1.88 & 2.47 & 2.23 & 2.24 & 2.85 & 1.56 & 2.65 & 2.45 \\
        UltraEdit~\citep{zhao2024ultraedit} & 3.44 & 2.81 & 2.13 & 2.96 & 1.45 & 2.83 & 3.76 & 1.91 & 2.98 & 2.70 \\
        OmniGen~\citep{xiao2025omnigen} & 3.47 & 3.04 & 1.71 & 2.94 & 2.43 & 3.21 & 4.19 & 2.24 & 3.38 & 2.96 \\
        ICEdit~\citep{zhang2025context} & 3.58 & 3.39 & 1.73 & 3.15 & 2.93 & 3.08 & 3.84 & 2.04 & 3.68 & 3.05 \\
        Step1X-Edit~\citep{liu2025step1x} & 3.88 & 3.14 & 1.76 & 3.40 & 2.41 & 3.16 & 4.63 & 2.64 & 2.52 & 3.06 \\
        BAGEL~\citep{deng2025bagel} & 3.56 & 3.31 & 1.70 & 3.3 & 2.62 & 3.24 & 4.49 & 2.38 & 4.17 & 3.20 \\
        UniWorld-V1~\citep{lin2025uniworldv1} & 3.82 & 3.64 & 2.27 & 3.47 & 3.24 & 2.99 & 4.21 & 2.96 & 2.74 & 3.26 \\
        OmniGen2~\citep{wu2025omnigen2} & 3.57 & 3.06 & 1.77 & 3.74 & 3.20 & 3.57 & \underline{4.81} & 2.52 & 4.68 & 3.44 \\
        FLUX.1 Kontext [Pro]~\citep{labs2025kontext} & 4.25 & 4.15 & 2.35 & \underline{4.56} & 3.57 & 4.26 & 4.57 & 3.68 & 4.63 & 4.00 \\
        GPT Image 1 [High]~\citep{gptimage} & \textbf{4.61} & \textbf{4.33} & \underline{2.90} & 4.35 & \underline{3.66} & \textbf{4.57} & \textbf{4.93} & \textbf{3.96} & \textbf{4.89} & \underline{4.20} \\
        \midrule
        \bf Qwen-Image & \underline{4.38} & \underline{4.16} & \textbf{3.43} & \textbf{4.66} & \textbf{4.14} & \underline{4.38} & \underline{4.81} & \underline{3.82} & \underline{4.69} & \textbf{4.27} \\
        \bottomrule
    \end{tabular}
    }
    \label{tab:imgedit}
\end{table}

\paragraph{ImgEdit} Table~\ref{tab:imgedit} presents the results on the \textbf{ImgEdit} benchmark~\citep{ye2025imgedit}, which covers nine common editing tasks across diverse semantic categories with a total of 734 real-world test cases. Evaluation metrics include instruction adherence, image-editing quality, and detail preservation, all scored from 1 to 5. Qwen-Image ranks highest overall, closely followed by GPT Image 1 [High] and demonstrating competitive instruction-based editing performance.

\begin{table}[!h]
    \centering
    \caption{Quantitative comparison of novel view synthesis with both specialized models and general image generation models. We report PSNR, SSIM, LPIPS on the GSO~\citep{downs2022googlescannedobjectshighquality} dataset.}
    \resizebox{0.6\linewidth}{!}{
    \begin{tabular}{l | c c c}
        \toprule
        \bf Model & \bf PSNR$\uparrow$ & \bf SSIM$\uparrow$ & \bf LPIPS$\downarrow$ \\
        \midrule
        Zero123~\citep{liu2023zero1to3} & 13.48 & 0.854 & 0.166 \\
        ImageDream~\citep{wang2023imagedream} &15.22&0.883& 0.164\\
        CRM~\citep{wang2024crm} &\textbf{15.93} & \textbf{0.891}& \textbf{0.152}\\
        \midrule
        GPT Image 1 [High]~\citep{gptimage} &12.07&0.804&0.361\\
        BAGEL~\citep{deng2025bagel}&13.78&0.825&0.237\\
        FLUX.1 Kontext [Pro]~\citep{labs2025kontext}&14.50&0.859&0.201\\
        \bf Qwen-Image & \textbf{15.11} & \textbf{0.884} & \textbf{0.153} \\
        \bottomrule
    \end{tabular}
    }
    \label{tab:quantitative_comparison_NVS}
\end{table}

\paragraph{Novel view synthesis} Table~\ref{tab:quantitative_comparison_NVS} shows the results of novel view synthesis on GSO~\citep{downs2022googlescannedobjectshighquality} dataset. We compare the similarity of the generated novel view image of a 3D object given its front view with the ground truth image. We prompt Qwen-Image with prompts like "turn left 90 degrees, a dog" to instruct the model to perform novel view synthesis. Qwen-Image demonstrates highly competitive results among the baselines, achieving state-of-the-art performance of novel view synthesis.

\begin{table*}[!t]
  \centering
  \caption{Quantitative comparison of depth estimation with both specialized models and multi-task models on zero-shot datasets. Qwen-Image can perform \textit{on par} with state-of-the-art models.
  }

\resizebox{.99\linewidth}{!}{%
  \begin{tabular}{@{}l|cc|cc|cc|cc|cc@{}}
    \toprule
	
	\multirow{2}{*}{\bf Model}  & \multicolumn{2}{c|}{\bf KITTI}  & \multicolumn{2}{c|}{\bf NYUv2} & \multicolumn{2}{c|}{\bf ScanNet}
 & \multicolumn{2}{c|}{\bf DIODE} & \multicolumn{2}{c}{\bf ETH3D}\\
	
    \cmidrule{2-11}
	
     &  AbsRel$\downarrow$ & $\delta_1$$\uparrow$ & AbsRel$\downarrow$ & $\delta_1$$\uparrow$ & AbsRel$\downarrow$ & $\delta_1$$\uparrow$ & AbsRel$\downarrow$ & $\delta_1$$\uparrow$ & AbsRel$\downarrow$ & $\delta_1$$\uparrow$ \\

     \midrule
       
    MiDaS~\citep{ranftl20midas}     & 0.236  & 0.630
     		& 0.111	& 0.885
                & 0.121 & 0.846
     		& 0.332	& 0.715
                & 0.184  & 0.752
     		\\

    DPT-large~\citep{ranftl21dptlarge} 	& 0.100  & 0.901
     		& 0.098	& 0.903
                & 0.082 & 0.934
     		& 0.182	& 0.758
                & 0.078 & 0.946
     		\\

    DepthAnything~\citep{yang2024depth} 	& 0.080  & 0.946
     		& 0.043	& 0.980
                & 0.043  & 0.981
     		& 0.261	& 0.759
                & 0.058  & \textbf{0.984}
     		\\

    DepthAnything v2~\citep{yang2024depth2}	& 0.080  & 0.943
     		& 0.043	& 0.979
                & 0.042  & 0.979
     		& 0.321	& 0.758
                & 0.066  & 0.983
     		\\

    Depth Pro~\citep{bochkovskii2024depthpro}  	& 0.055  & 0.974
     		& 0.042	& 0.977
                & 0.041  & 0.978
     		& 0.217	& 0.764
                & 0.043  & 0.974
     		\\

    Metric3D v2~\citep{hu2024metric3d} 	& \textbf{0.052}  & \textbf{0.979}
     		& \textbf{0.039}	& \textbf{0.979}
                & \textbf{0.023}  & \textbf{0.989}
     		& \textbf{0.147}	& \textbf{0.892}
                & \textbf{0.040}  & 0.983
     		\\

    \midrule
    
    GeoWizard~\citep{fu2024geowizard}  & 0.097  & 0.921
     		& \textbf{0.052}	& \underline{0.966}
                & \underline{0.061} & \underline{0.953}
     		& 0.297	& 0.792
                & \textbf{0.064}  & \underline{0.961}
     		\\

    DepthFM~\citep{gui2024depthfm}	& \underline{0.083}  & \underline{0.934}
     		& 0.065	& 0.956
                &  - & -
     		& \underline{0.225} & \underline{0.800}
                & -  & -
     		\\
            
    Marigold~\citep{ke2024marigold} 	& 0.099  & 0.916
     		& 0.055	& 0.964
                & 0.064  & 0.951
     		& 0.308	& 0.773
                & \underline{0.065}  & 0.960
     		\\

    DMP~\citep{lee2024dmp}   & 0.240  & 0.622
     		& 0.109	& 0.891
                & 0.146    & 0.814
     		& 0.361 	& 0.706
                & 0.128    &  0.857
     		\\

    \midrule

    \bf Qwen-Image & \textbf{0.078} & \textbf{0.951}
        & \underline{0.055}	& \textbf{0.967}
            & \textbf{0.047}  & \textbf{0.974}
        & \textbf{0.197}	& \textbf{0.832}
            & 0.066  & \textbf{0.962}
        \\
     
    \bottomrule
  \end{tabular}
  }
  \label{tab:depth}
\end{table*}

\paragraph{Depth Estimation} Table~\ref{tab:depth} summarizes performance on five widely used datasets: NYUv2~\citep{nathan12nyu}, KITTI~\citep{geiger13kitti}, ScanNet~\citep{dai17scannet}, DIODE~\citep{igor19diode}, and ETH3D~\citep{schops17eth3d}. During training, we adopt DepthPro~\citep{bochkovskii2024depthpro} as the teacher model to provide supervisory depth signals, following the protocol used in previous work such as DICEPTION~\citep{zhao2025diception}. Notably, these results are achieved with standalone supervised fine-tuning (SFT), in order to probe the model’s intrinsic task understanding capability.  Qwen-Image demonstrates highly competitive results among diffusion-based models, achieving state-of-the-art performance on several key metrics across these benchmarks.

\subsection{Qualitative Results}
To comprehensively compare the visual generation capabilities of Qwen-Image and state-of-the-art models, we first qualitatively compared the reconstruction performance of VAE on text-rich images in \cref{sec:qualitative_vae}. We further conducted qualitative comparison on two basic visual generation tasks, text-to-image (T2I) generation in \cref{sec:qualitative_t2i} and image editing (TI2I) in \cref{sec:qualitative_ti2i}, to comprehensively evaluate the basic generation capabilities of the models.

\subsubsection{Qualitative Results on VAE Reconstruction}\label{sec:qualitative_vae}

\begin{figure}[t]
\centering
\includegraphics[width=1\linewidth]{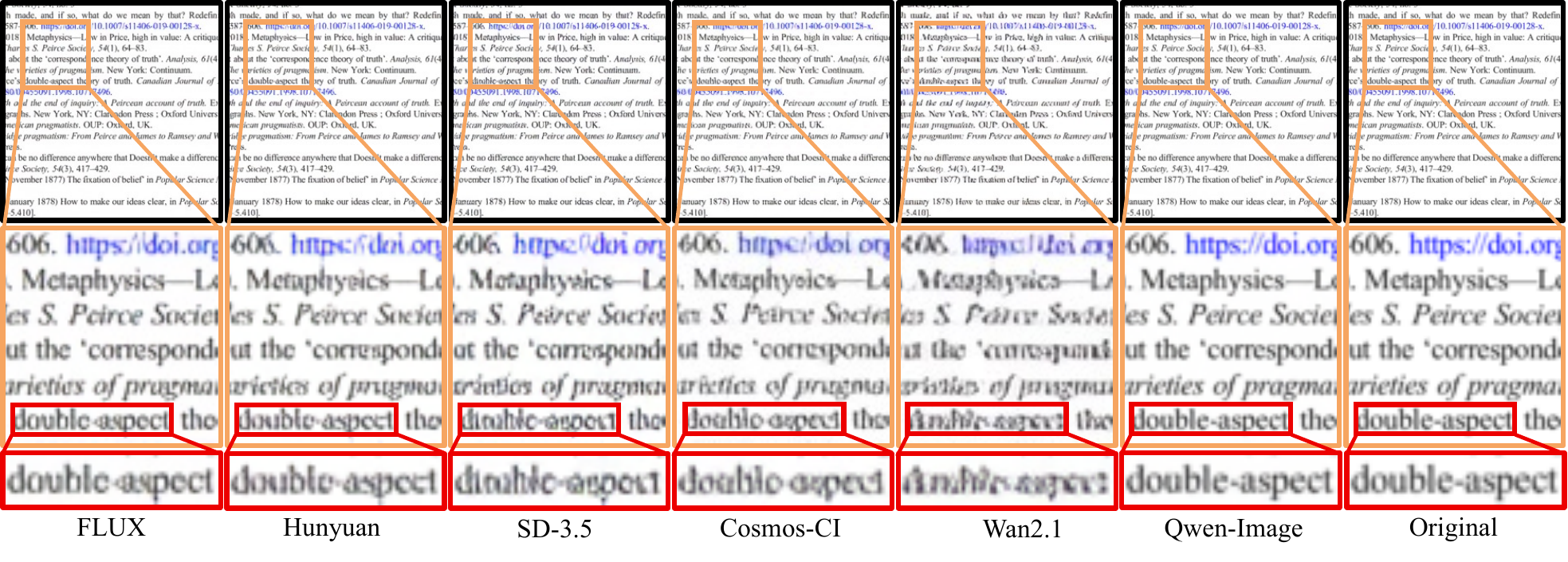}
\caption{Visualization of VAE reconstruction. We progressively zoom into the details across three rows (black, orange, red) to compare how different VAEs reconstruct small text in dense document images.}
\label{fig:vae_cases}
\end{figure}
Figure~\ref{fig:vae_cases} presents qualitative results of reconstructing text-rich images with the state-of-the-art image VAEs. The first row illustrates the reconstruction of a PDF image containing English text. In our result, the phrase “double-aspect” remains clearly legible, whereas it is unrecognizable in the reconstructions produced by other models. Overall, Qwen-Image-VAE delivers more precise reconstructions for images with small texts.

\subsubsection{Qualitative Results on Image Generation}\label{sec:qualitative_t2i}

In order to comprehensively evaluate Qwen-Image's text-to-image generation capability, we conduct qualitative evaluation from four aspects: English Text Rendering, Chinese Text Rendering, Multi-Object Generation, and Spatial Relationship Generation. 
For comparative analysis, we benchmark our model against leading Text-to-Image foundation models, including both close-source models (GPT Image 1 [High]~\citep{gptimage}, Seedream 3.0~\citep{gao2025seedream}, Recraft V3~\citep{recraftv3} and open-source models (Hidream-I1-Full~\citep{cai2025hidream} and Lumina-Image 2.0~\citep{qin2025lumina}).

\begin{figure}[!h]
\centering
\includegraphics[width=0.98\linewidth]{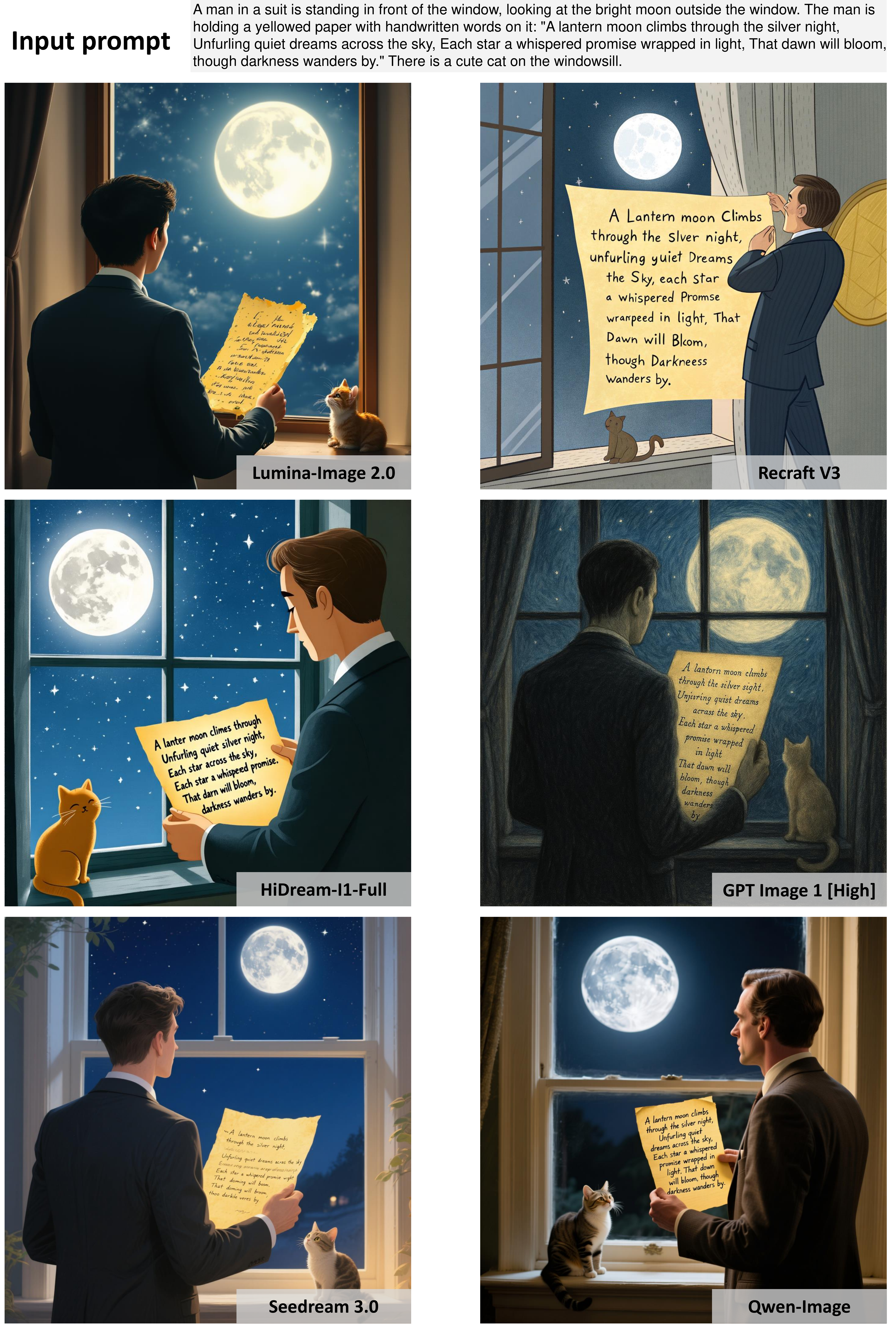}
\caption{Comparison of long English rendering capability in image generation. 
This case requires rendering a long paragraph, and only Qwen-Image and GPT Image 1 [High] manage to render such a long text clearly and almost perfectly. 
The other models either omit words or produce duplicates.}
\label{fig:t2i_moon}
\end{figure}

\begin{figure}[!h]
\centering
\includegraphics[width=0.95\linewidth]{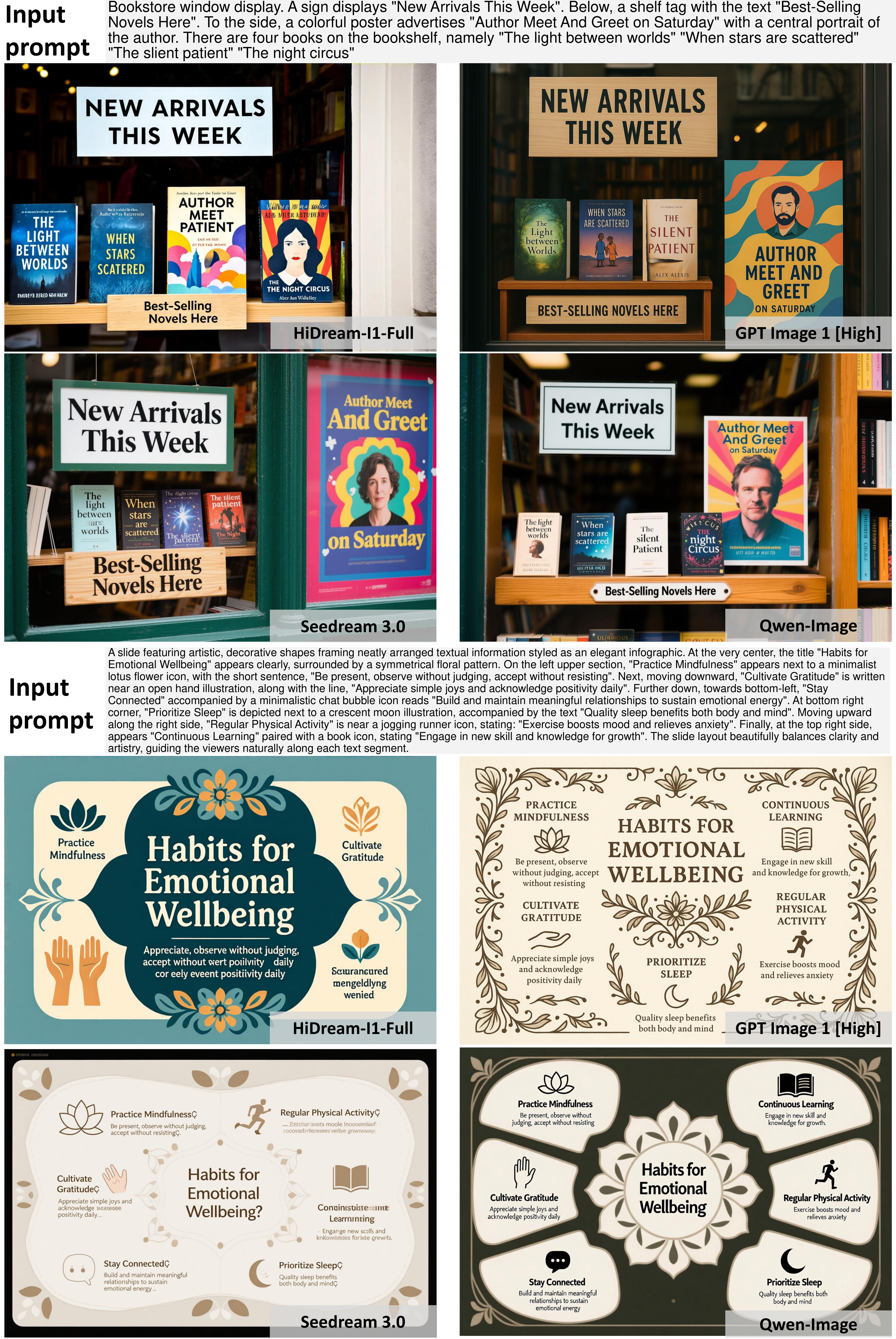}
\caption{Comparison of complex English rendering capability in image generation. 
We present two cases to illustrate the model's ability to generate multiple English texts in different locations of the real scene and the slide. 
Only Qwen-image can follow the complex prompts to successfully render the text in reasonable location.}
\label{fig:t2i_book}
\end{figure}

\begin{figure}[!h]
\centering
\includegraphics[width=0.96\linewidth]{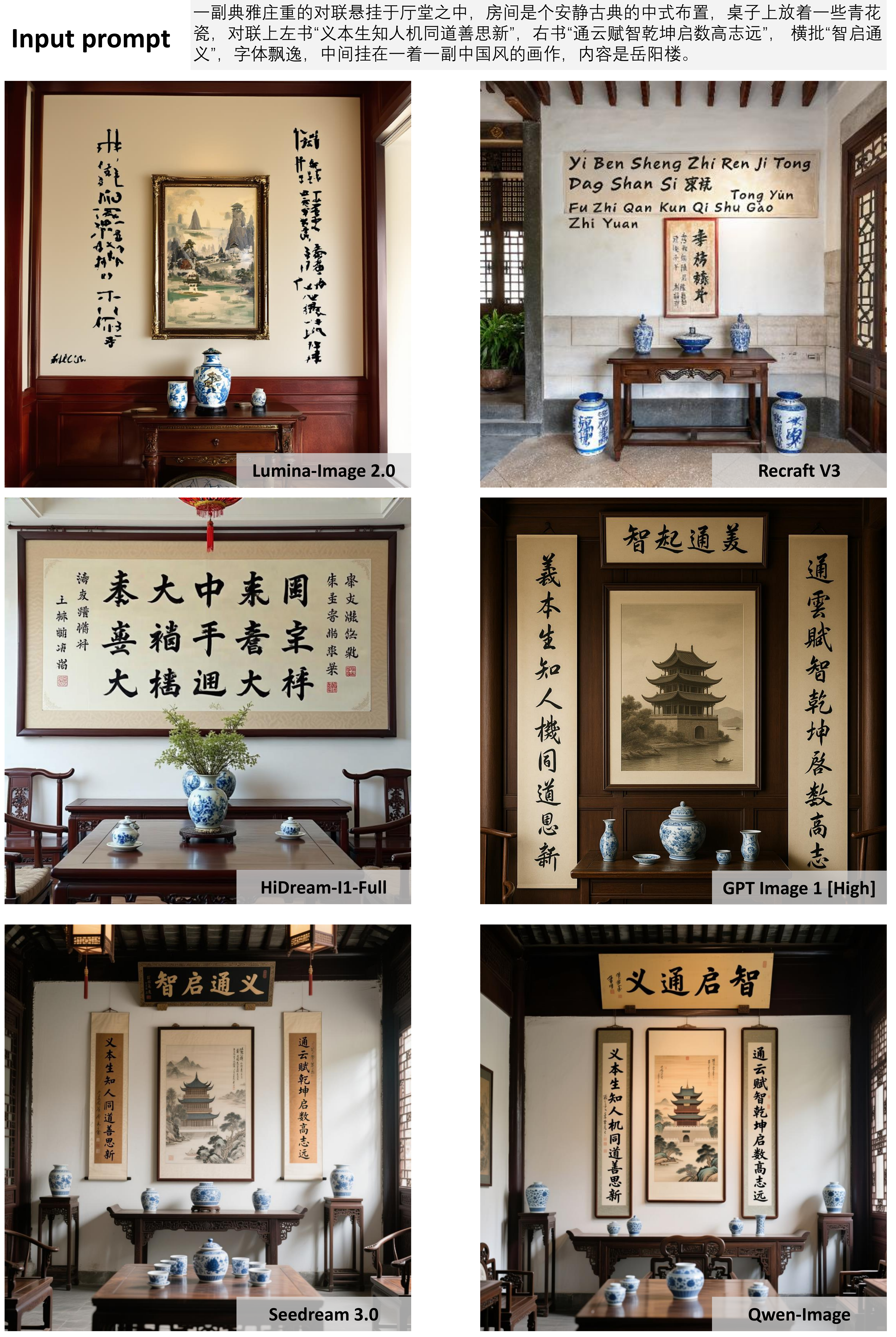}
\caption{Comparison of Chinese text rendering capability in image generation. Qwen-Image accurately generates the expected Chinese couplet. In contrast, GPT Image 1 [high] and Seedream 3.0 miss or generate distorted characters. While other model in comparison cannot generate correct Chinese couplets.}
\label{fig:t2i_tongyi}
\end{figure}

\begin{figure}[!h]
\centering
\includegraphics[width=1\linewidth]{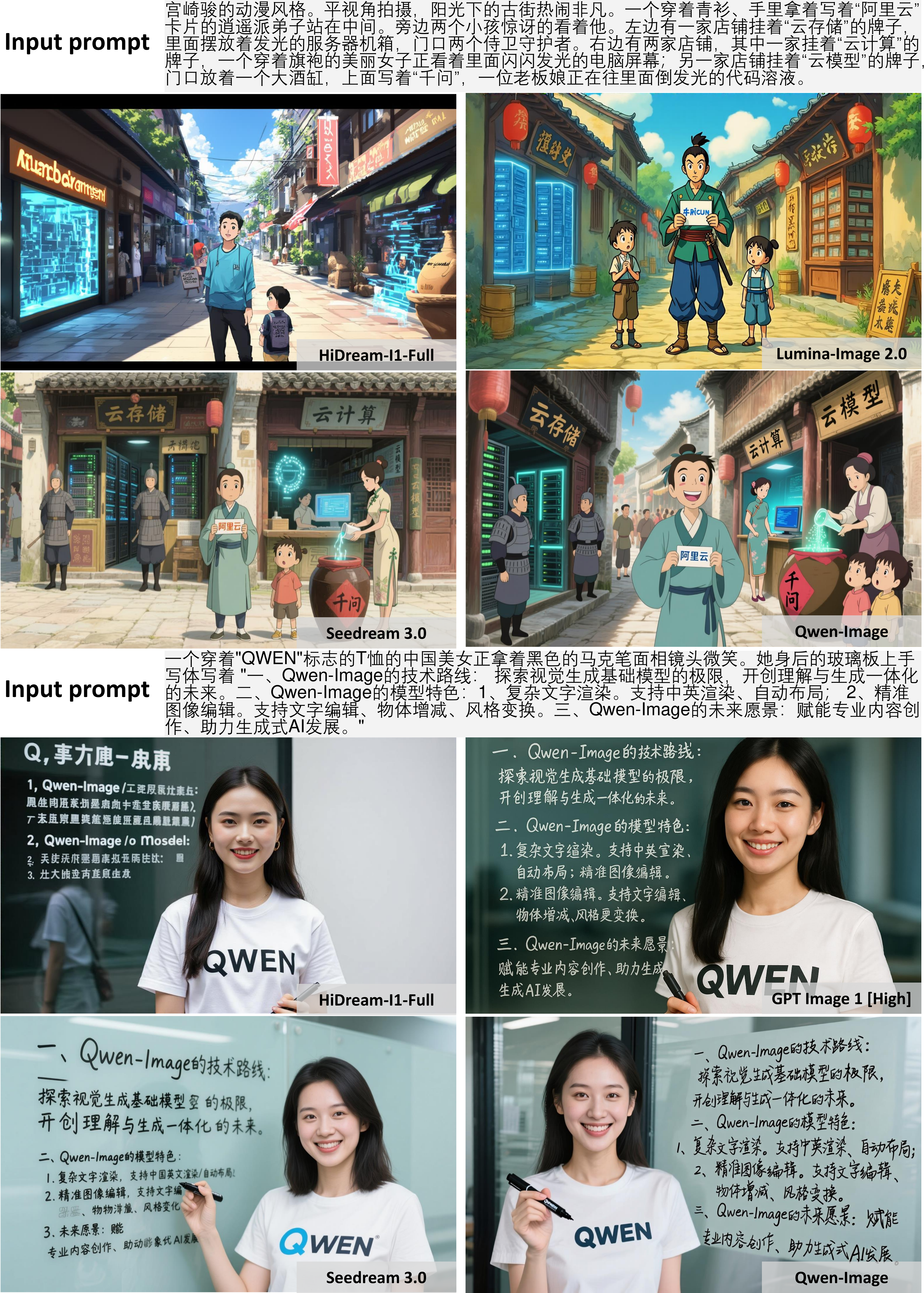}
\caption{Comparison of complex Chinese text rendering in image generation: the first case shows that Qwen-Image can render text on multiple objects while maintaining consistency with the real scene, such as aligning text with the depth and tilt of each plaque; the second case demonstrates its ability to render structured paragraph text in a glass panel. Qwen-Image is the only model capable of accurately rendering long text.\looseness=-1}
\label{fig:t2i_aliyun}
\end{figure}

\begin{figure}[!h]
\centering
\includegraphics[width=0.95\linewidth]{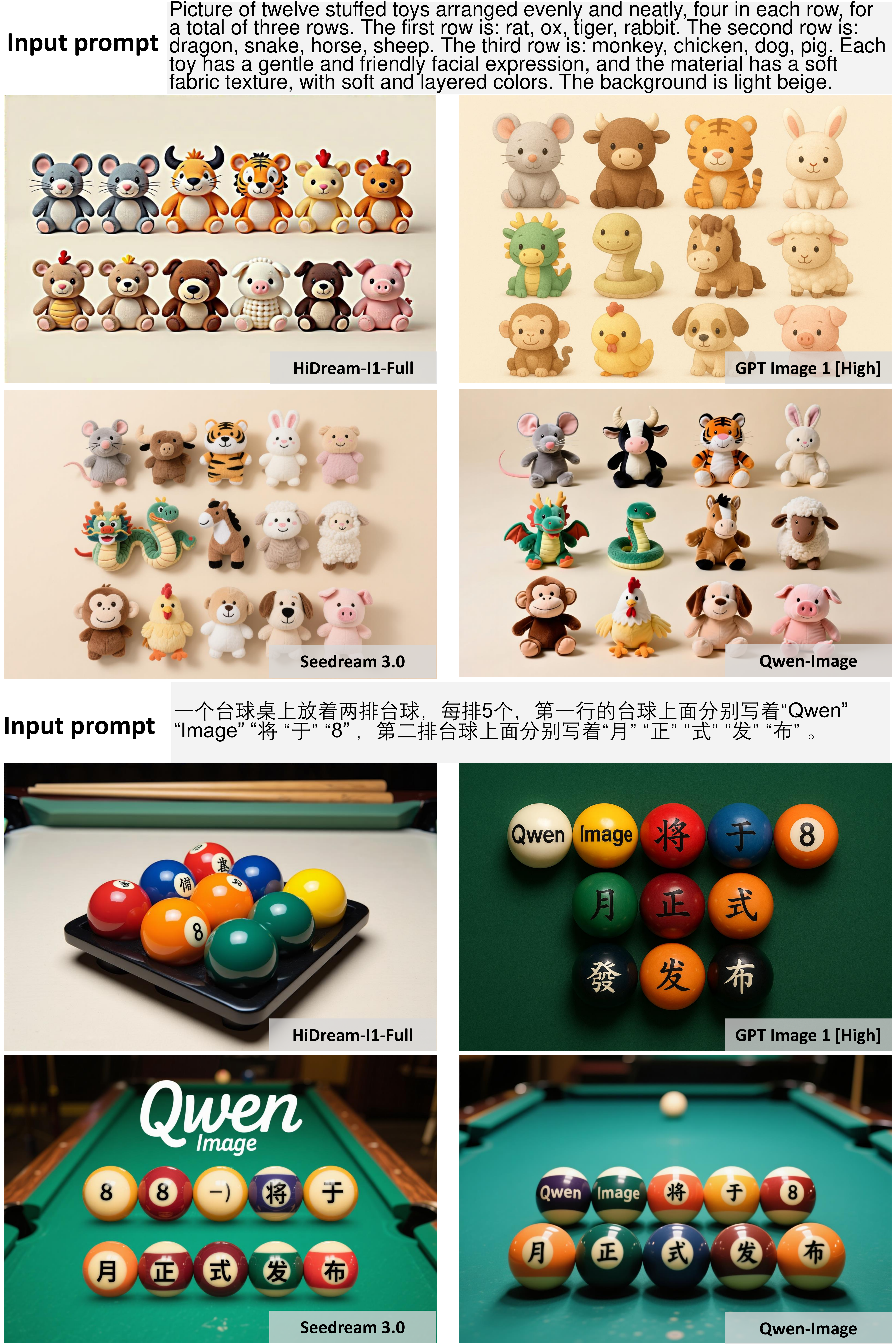}
\caption{Comparison of multi-object modeling in image generation: Qwen-Image accurately renders the 12 Chinese zodiac animals and materials in the first case, and handles complex bilingual text across multiple objects in the second.}
\label{fig:t2i_multiple}
\end{figure}

\begin{figure}[!h]
\centering
\includegraphics[width=1\linewidth]{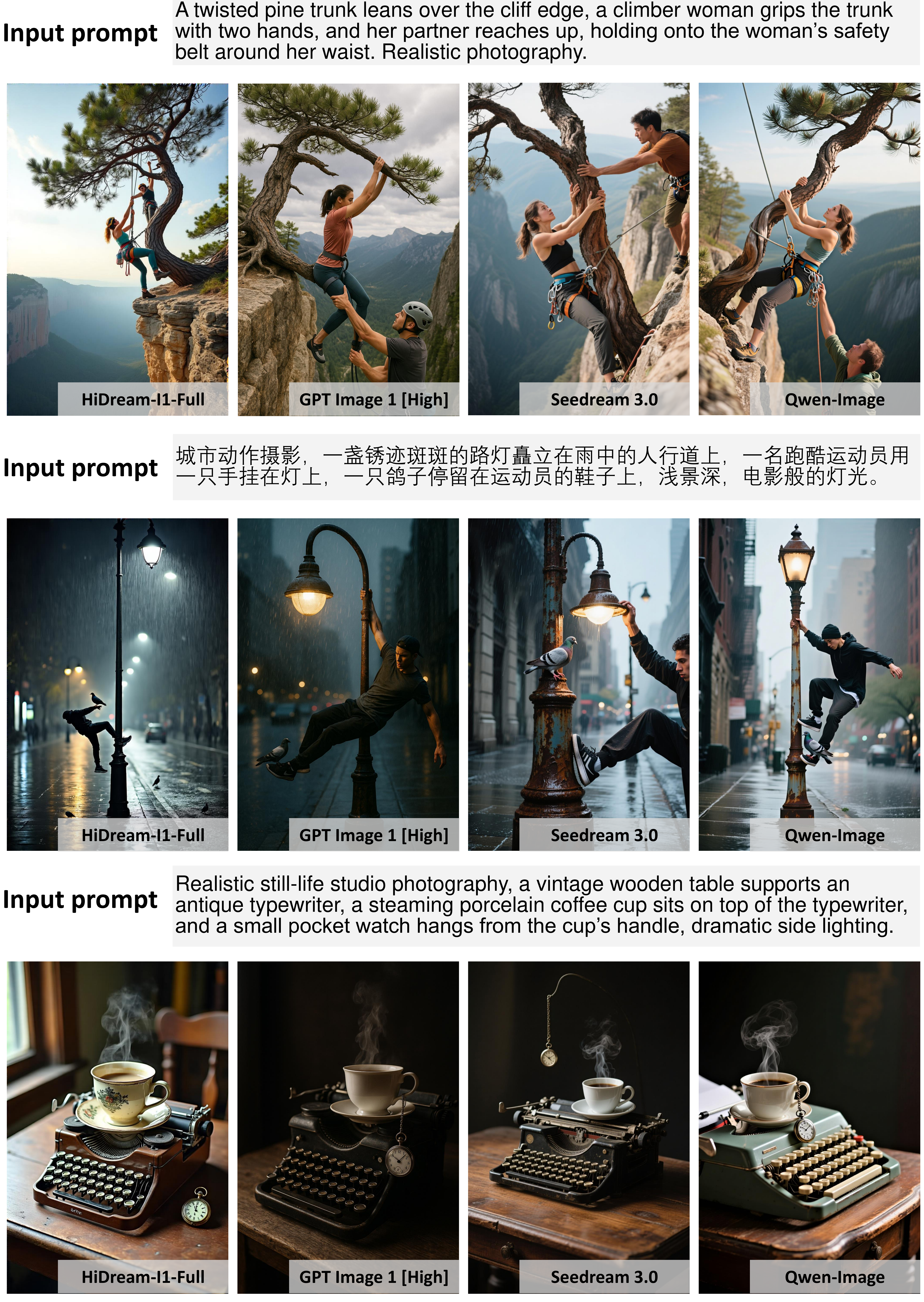}
\caption{Comparison of spatial relationship modeling capability in image generation. We present three cases to demonstrate interactions involving multiple people and multiple objects. We find that both Qwen-Image and GPT Image 1 [High] exhibit strong ability to understand relationships.}
\label{fig:t2i_relation}
\end{figure}

\paragraph{English Text Rendering} 
Figure~\ref{fig:t2i_moon} and Figure~\ref{fig:t2i_book} show the qualitative comparison of English text rendering. 
As shown in Figure~\ref{fig:t2i_moon}, Qwen-Image achieves a more realistic visual style and better rendering quality for a long English paragraph.
With respect to text rendering, our model demonstrates greater fidelity to the given prompts, effectively avoiding issues such as missing, wrong, or duplicate characters (e.g. wrong ''lantern'' and ''Unfurling'' in GPT Image 1 [High], wrong ''silver'' and ''quiet'' in Recraft V3, redundant and distorted text in Seedream 3.0). 
In the upper half of Figure~\ref{fig:t2i_book}, Qwen-Image correctly renders text in seven different locations, demonstrating its complex text rendering capabilities. 
In contrast, GPT Image 1 misses "The night circus", and the text rendered by Seedream 3.0 and Hidream-I1-Full is distorted.
For the bottom half of Figure~\ref{fig:t2i_book}, Qwen-Image not only successfully renders each text segment, but also presents a slide with reasonable layout and visually aesthetic. In comparison, GPT Image 1 misses "Stay Connected", Hidream-I1-Full and seedream 3.0 fail to render correct characters.

\paragraph{Chinese Text Rendering} 
Figure~\ref{fig:t2i_tongyi} and Figure~\ref{fig:t2i_aliyun} show a qualitative comparison of Chinese text rendering. In Figure~\ref{fig:t2i_tongyi}, Qwen-Image accurately generates the expected Chinese couplet, faithfully reproducing the content and style of the text, and accurately depicts the required room layout and placement.
In contrast, GPT Image 1 and Seedream 3.0 miss or generate distorted characters (missed \begin{CJK*}{UTF8}{gbsn}"远"\end{CJK*} and \begin{CJK*}{UTF8}{gbsn}"善"\end{CJK*} in GPT Image 1, missed \begin{CJK*}{UTF8}{gbsn}"智"\end{CJK*} and \begin{CJK*}{UTF8}{gbsn}"机"\end{CJK*} in Seedream 3.0), while other models cannot generate correct Chinese couplets. In Figure~\ref{fig:t2i_aliyun}, the upper part shows an anime style scene, where Qwen-Image can correctly generate multiple characters and store plaques, perfectly following the requirements of input prompt for spatial layout and text rendering. 
Seedream 3.0 encounters difficulties in complex spatial layouts, missing some scenes and characters, while other models cannot correctly understand complex text and spatial instructions. 
The bottom part shows that Qwen-Image can generate realistic and beautifully typeset handwritten text in complex scenes, which perfectly following the input prompt, while other models struggle to generate structured paragraph text.

\paragraph{Multi-Object Generation} 
As shown in the upper half of Figure~\ref{fig:t2i_multiple}, Qwen-Image accurately generates all required animals, faithfully preserves their specified positions, and consistently applies the correct plush style. In contrast, GPT Image 1 fails to generate images in the plush style, while Recraft V3 and Seedream 3.0 produce incorrect animals that do not match the prompt. 
For the bottom part of Figure~\ref{fig:t2i_multiple}, Qwen-Image not only correctly renders mixed-language texts on the billiards, but also strictly follows the instruction to arrange the billiards in two rows.
GPT Image 1 cannot perfectly follow the layout requirements of instruction and also incorrectly generate a Chinese character \begin{CJK*}{UTF8}{bsmi}"發"\end{CJK*}, while other models cannot correctly generate most Chinese characters.

\paragraph{Spatial Relationship Generation}
In the first part of Figure~\ref{fig:t2i_relation}, Qwen-Image generates an image that accurately reflects the prompt, capturing both the correct climbing scene and the specified interaction between the two people. In contrast, GPT Image 1, Seedream 3.0, and Recraft V3 fail to fully follow the prompt: these models produce incorrect interactions between the climbers. This comparison demonstrates the strong ability of Qwen-Image to understand and precisely follow complex prompts.
For the remaining two parts, only Qwen-Image and GPT Image 1 can accurately depict the spatial relationship between the character and the pigeon, as well as the pocket watch and the cup handle.

\begin{figure}[h]
\centering
\includegraphics[width=1\linewidth]{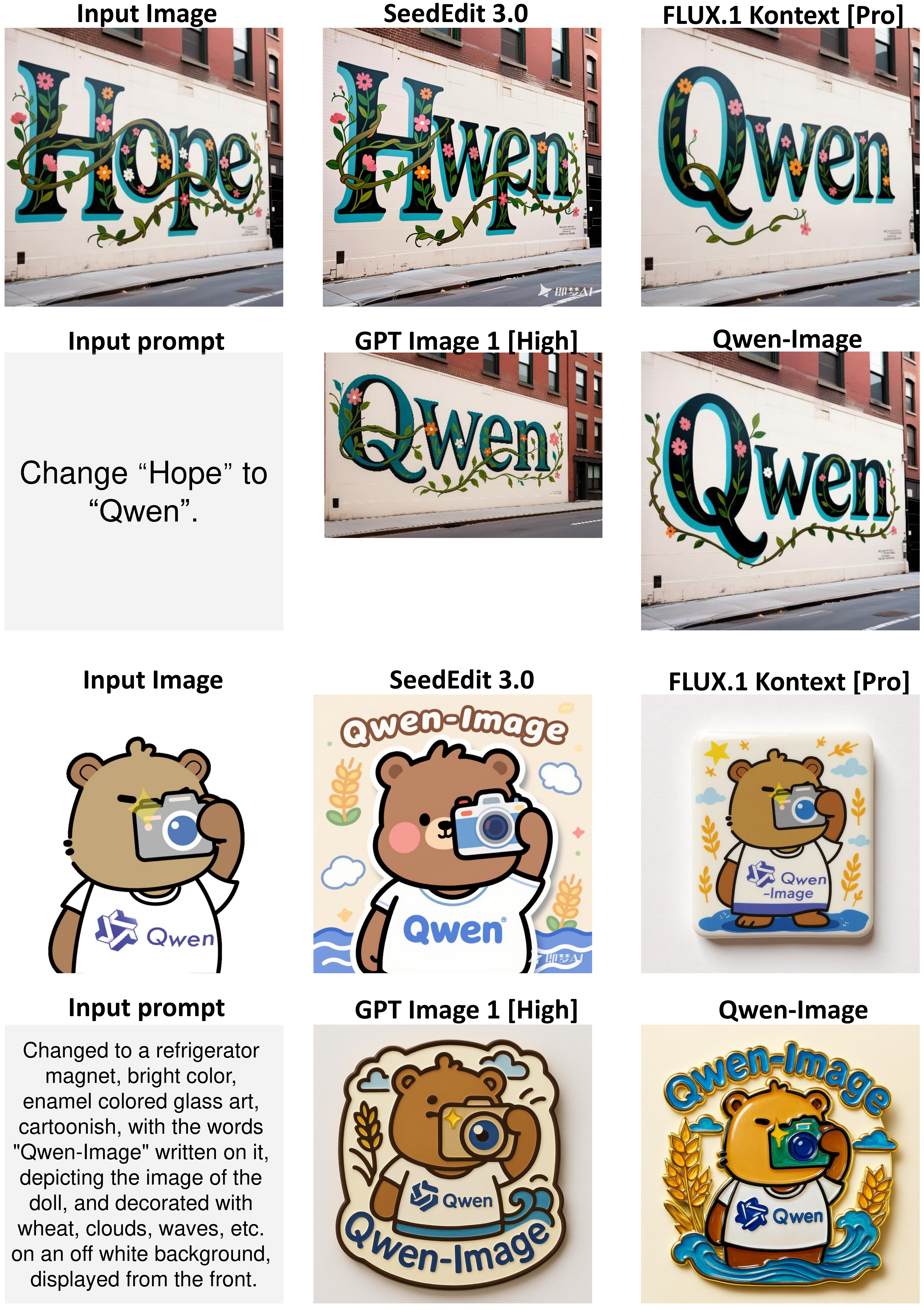}
\caption{Qualitative comparison on text and material modification. Both FLUX.1 Kontext [Pro] and Qwen-Image are able to accurately modify text while preserving the original style. In the example below, Qwen-Image is the only model that successfully presents the enamel material.}
\label{fig:ti2i_hope}
\end{figure}

\begin{figure}[h]
\centering
\includegraphics[width=1\linewidth]{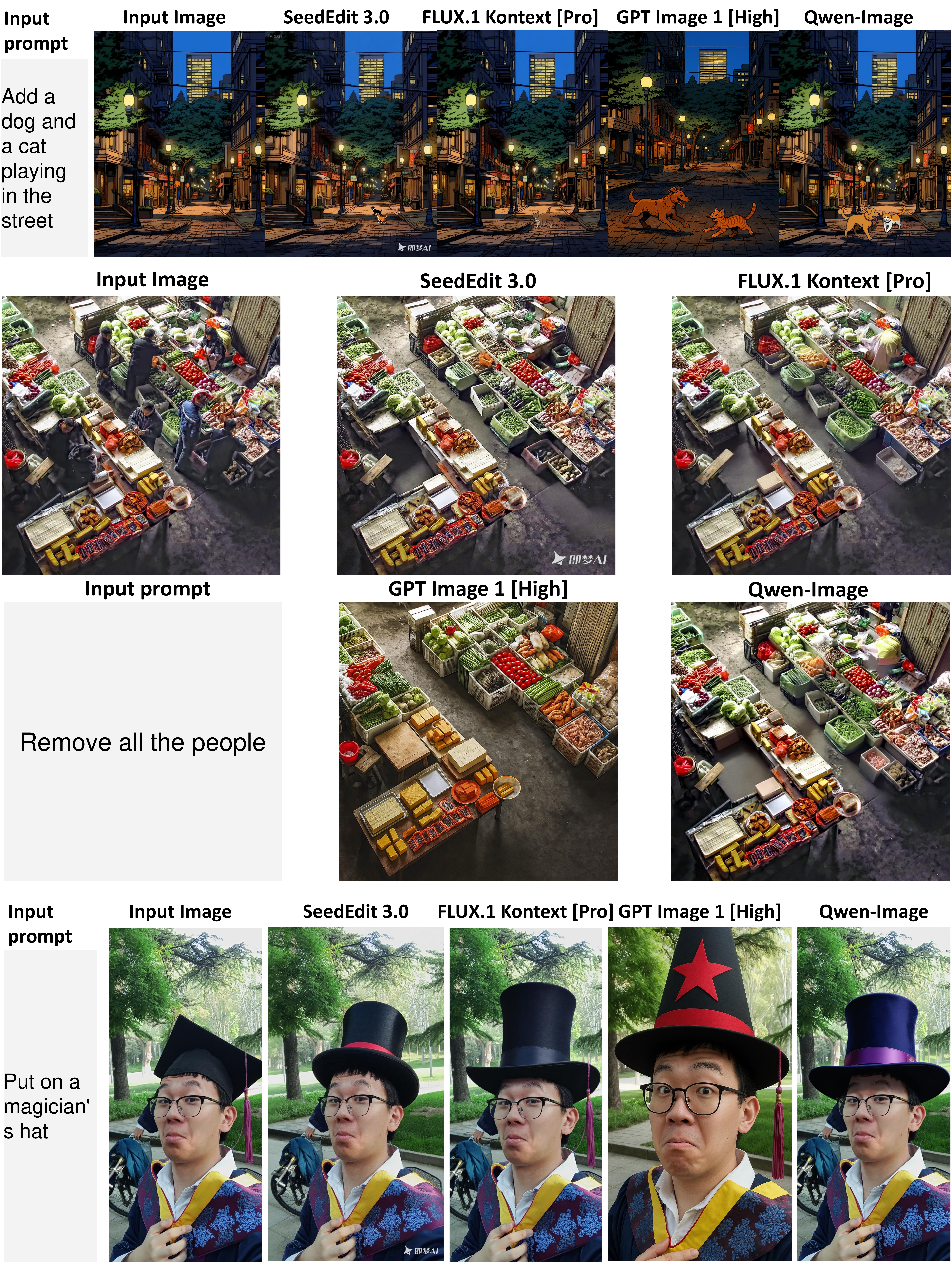}
\caption{Qualitative comparison on object editing (addition, removal, and replacement):
Object editing is a relatively stable capability for all models. Qwen-Image demonstrates superior consistency in unmodified regions and achieves better style alignment for the newly generated objects.}
\label{fig:ti2i_add}
\end{figure}

\begin{figure}[h]
\centering
\includegraphics[width=1\linewidth]{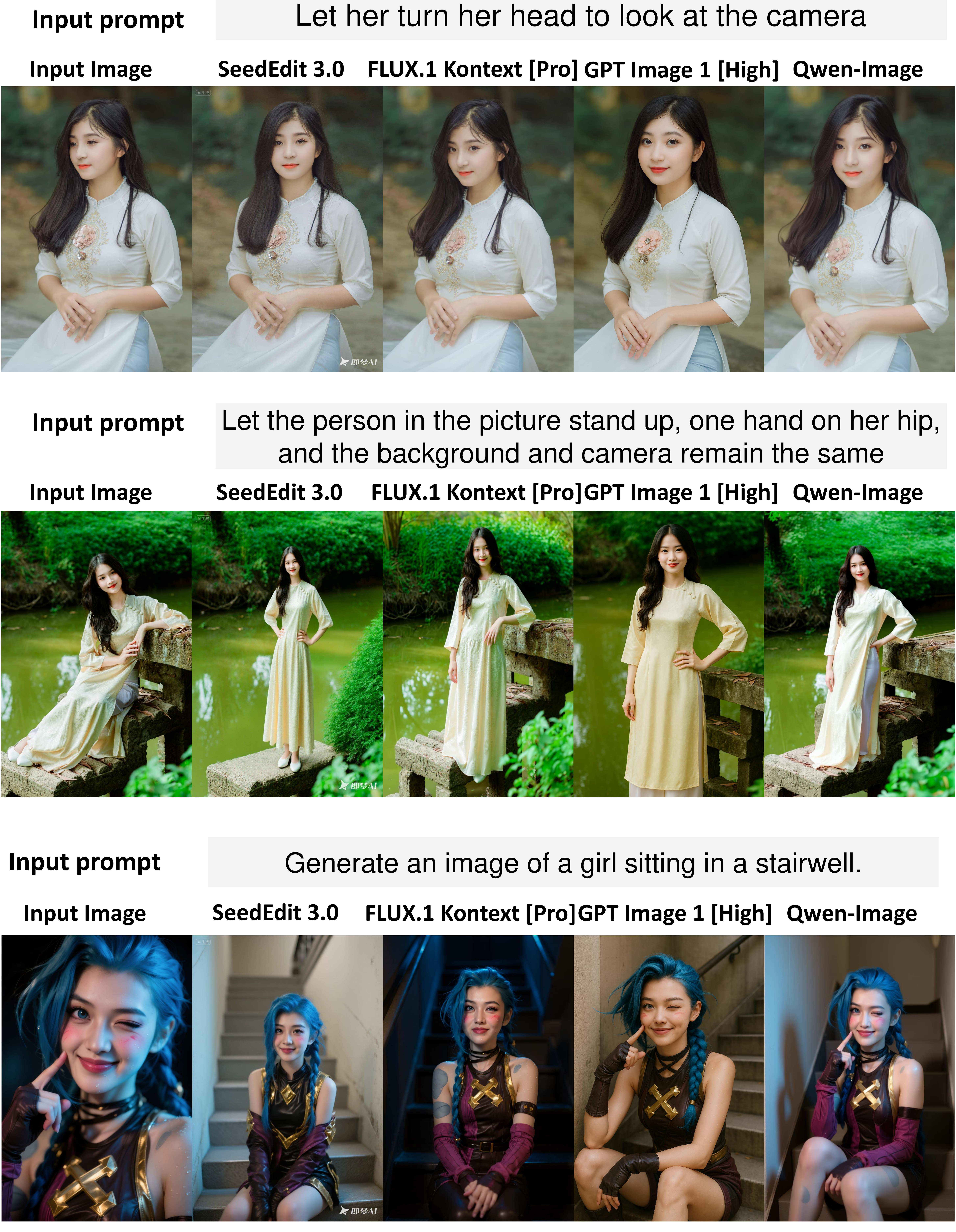}
\caption{Qualitative comparison on pose manipulation. Qwen-Image is able to accurately follow pose manipulation instructions while preserving fine details of the person (such as hair strands) and maintaining consistency in the background (e.g., the stone steps behind the subject).}
\label{fig:ti2i_pose}
\end{figure}

\begin{figure}[h]
\centering
\includegraphics[width=1\linewidth]{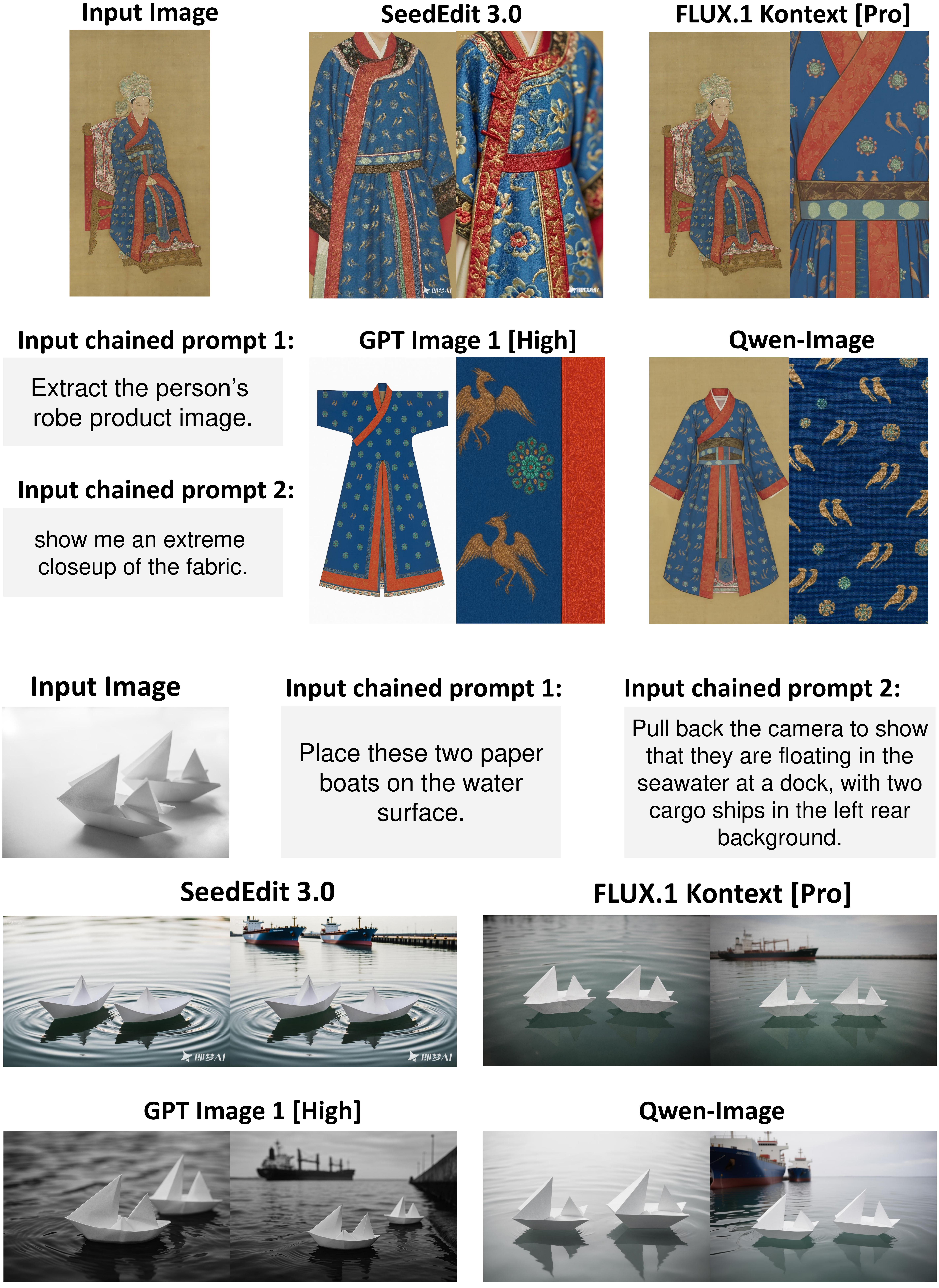}
\caption{Examples of two types of chained editing: extract + zoom-in (top) and placement + zoom-out (bottom). GPT Image 1 [High] and Qwen-Image correctly understand the extract operation, but only Qwen-Image accurately captures and magnifies the garment's texture. In the second case, Qwen-Image preserves the open-ended stern of the paper boat throughout the entire chained editing process.}
\label{fig:ti2i_chain}
\end{figure}

\begin{figure}[h]
\centering
\includegraphics[width=1\linewidth]{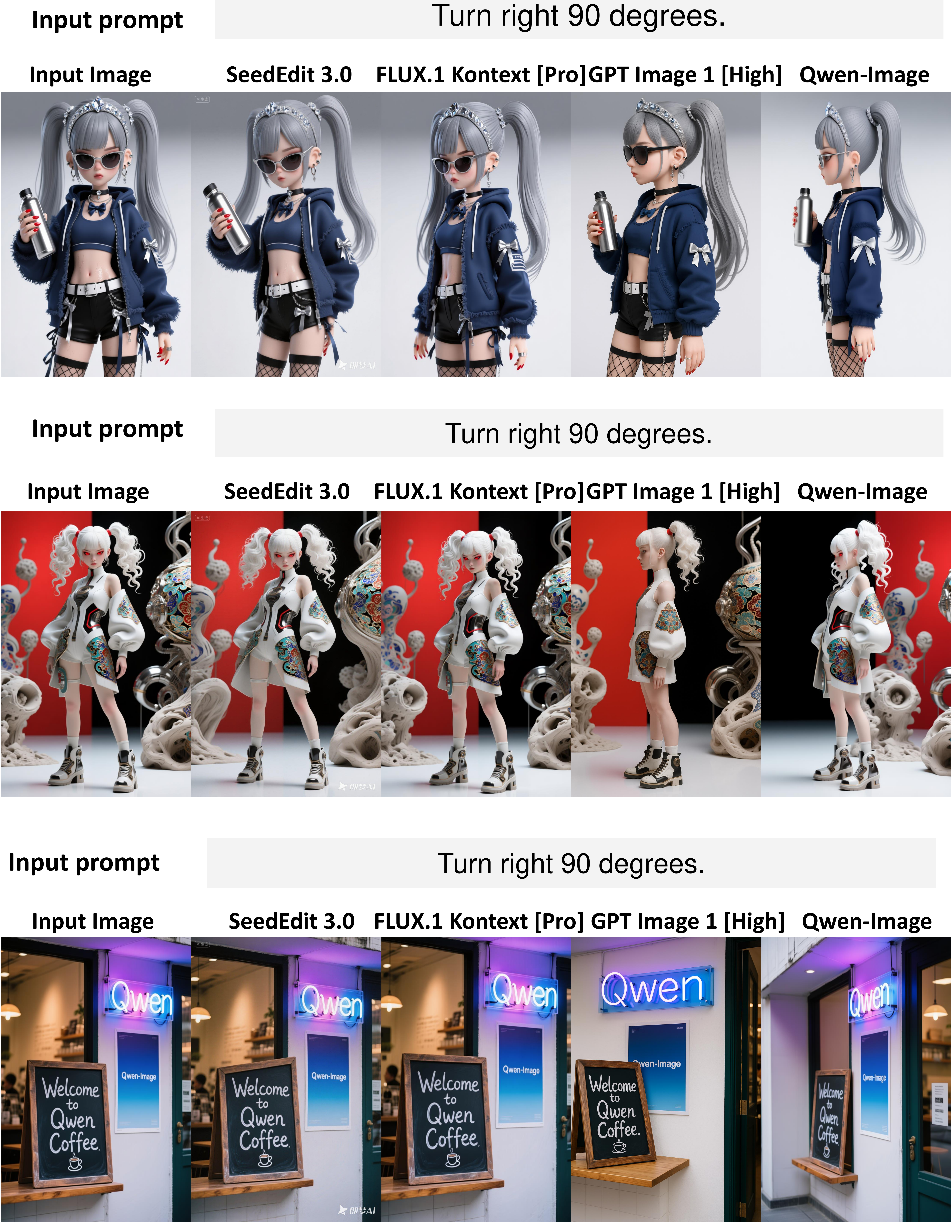}
\caption{Qualitative comparison on novel view synthesis. We evaluate three progressively challenging rotation tasks under the same text prompt "Turn right 90 degrees": (1) single-person rotation, (2) co-rotation of person and background, and (3) real-world scenario. In all cases, Qwen-Image achieves the most accurate and consistent results. While other models (e.g., GPT Image 1 [High]) handle basic subject rotation, they fail to rotate the background or preserve scene details.}
\label{fig:ti2i_90}
\end{figure}

\subsubsection{Qualitative Results on Image Editing}
\label{sec:qualitative_ti2i}

To comprehensively assess the image editing (TI2I) capability of Qwen-Image, we conduct a qualitative evaluation focusing on five key aspects: text and material editing, object addition/removal/replacement, pose manipulation, chained editing, and novel view synthesis. 
For comparative analysis, we benchmark our model against several leading instruction-based image editing models, including SeedEdit 3.0~\citep{wang2025seededit}\footnote{SeedEdit 3.0 does not provide an official API, all watermarked images were obtained via their \href{https://jimeng.jianying.com/}{web interface}.}, FLUX.1 Kontext [Pro]~\citep{labs2025kontext}, and GPT Image 1 [High]~\citep{gptimage}.

\paragraph{Text and Material Editing}
Figure~\ref{fig:ti2i_hope} presents a qualitative comparison of text and material editing capabilities.
In the top example, which involves editing text with a complex style, Seedream 3.0 fails to change the letter "H" to "Q", and GPT Image 1 [High] does not preserve the original style. 
Both Qwen-Image and FLUX.1 Kontext [Pro] successfully modify the text while maintaining style consistency.
In the lower example, all models except FLUX.1 Kontext [Pro] accurately add the required text and related elements. 
Notably, Qwen-Image is the only model that correctly generates the requested enamel colored glass art, demonstrating superior material rendering and instruction-following ability.

\paragraph{Object Addition/Removal/Replacement}
Object addition, removal, and replacement are among the most common tasks in instruction-based image editing. 
In Figure~\ref{fig:ti2i_add}, we compare the performance of various models on challenging real-world scenarios.
With the exception of GPT Image 1 [High], which often fails to maintain overall image consistency, all other models generally perform well in preserving unedited regions.
In the top case, where the task is to add both a cat and a dog in a cartoon style, the model must ensure that the new objects match the overall artistic style. 
We observe that FLUX.1 Kontext [Pro] struggles with consistency when editing non-photorealistic images, while both SeedEdit 3.0 and Qwen-Image produce coherent results that align well with the desired cartoon style.
For the middle example, which involves removing all people from a crowded scene—a relatively complex instruction—all models complete the task accurately, differing only in minor details. 
Additionally, we notice occasional zoom-in and zoom-out effects across different models during editing.

\paragraph{Pose Manipulation}
Figure~\ref{fig:ti2i_pose} presents a qualitative comparison of pose manipulation abilities across different models.
In the first example, only FLUX.1 Kontext [Pro] and Qwen-Image are able to preserve fine details such as the subject’s hair strands during pose editing.
In the second case, which requires maintaining clothing consistency and scene stability during pose changes, Qwen-Image is the only model that keeps both the background and character unchanged. Remarkably, Qwen-Image accurately infers from the input that the person is wearing a side-slit dress over silk trousers, and faithfully reveals the silk trousers in the standing pose.
In the third example, Qwen-Image again outperforms other models by better preserving the original pose and maintaining consistency in clothing decorations.

\paragraph{Chained Editing}
Chained editing refers to scenarios where generated images are iteratively used as context for subsequent editing steps.
In the first case of Figure~\ref{fig:ti2i_chain}, the task requires extracting a clothing item and depicting the close-up of its fabric details. 
We select a Chinese traditional painting as the source image. SeedEdit 3.0 and FLUX.1 Kontext [Pro] fail from the first prompt, while both GPT Image 1 [High] and Qwen-Image accurately extract the paired birds. 
Qwen-Image can better preserve fine texture details against GPT Image 1 [High].
In the second case, the input image features a boat with a double-opening stern. 
Both Qwen-Image and FLUX.1 Kontext [Pro] are able to preserve this structural feature throughout the entire chained editing process. However, 
FLUX.1 Kontext [Pro] fails to add two cargo ships as instructed, whereas Qwen-Image successfully completes the complete editing chain.

\paragraph{Novel View Synthesis}
Figure~\ref{fig:ti2i_90} evaluates the spatial reasoning and novel view synthesis capabilities of different models. 
SeedEdit 3.0 and FLUX.1 Kontext [Pro] cannot perform view rotation well under the same instruction. 
While GPT Image 1 [High] can generate new perspectives when a clear subject is present, it fails to generalize to real-world scenes with complex multiple objects. 
Only Qwen-Image maintains global consistency—including text fidelity and lighting structure—demonstrating superior spatial and semantic coherence in complex editing tasks.

\clearpage

\section{Conclusion}

In this paper, we introduce Qwen-Image, an image generation foundation model within the Qwen series, achieving major advancements in both complex text rendering and precise image editing. By constructing a comprehensive data pipeline and adopting a progressive curriculum learning strategy, Qwen-Image substantially improves its capability to render intricate text within generated images. Moreover, our improved multi-task training paradigm and dual-encoding mechanism significantly enhance the consistency and quality of image editing, effectively improving both semantic coherence and visual fidelity. Extensive experiments on public benchmarks consistently demonstrate the state-of-the-art performance of Qwen-Image across a wide range of image generation and editing tasks. These results underscore not only the technical robustness of Qwen-Image but also its broad applicability in real-world multimodal scenarios, marking a significant milestone in the evolution of large foundation models.

We now turn to a deeper discussion on the broader implications and significance of Qwen-Image:

Qwen-Image as an "image generation" model, in the context of image generation, redefines the priorities in generative modeling. Rather than merely optimizing for photorealism or aesthetic quality ("AI look"), Qwen-Image emphasizes precise alignment between text and image—especially in the challenging task of text rendering. We envision that by strengthening this capability in foundation models, future interfaces can evolve from purely language-based LUIs (Language User Interfaces) to vision-language VLUIs (Vision-Language User Interfaces). When LLMs struggle to convey visual attributes such as color, spatial relationships, or structural layouts, a VLUI empowered by Qwen-Image can generate richly illustrated, text-integrated imagery—enabling structured visual explanations and effective knowledge externalization, where complex ideas are transformed into comprehensible, multimodal representations.

Qwen-Image as an image "generation" model, in the context of image understanding, demonstrates that generative frameworks can effectively perform classical understanding tasks. For instance, in depth estimation, although Qwen-Image does not surpass specialized discriminative models, it achieves performance remarkably close to them. Crucially, while traditional expert models rely on discriminative understanding—directly mapping inputs to outputs without modeling underlying distributions—Qwen-Image leverages generative understanding: it first constructs a holistic distribution over visual content, from which depth is naturally inferred. This shift from direct inference to distributional reasoning opens new avenues for unified multimodal understanding.

Qwen-Image as an "image" generation model, in the context of 3D and video generation, shows strong generalization beyond 2D image synthesis. Through the lens of image editing, we apply Qwen-Image to novel view synthesis and find that, as a general-purpose image foundation model, it outperforms dedicated 3D models in several challenging rendering scenarios, exhibiting exceptional consistency across views. In pose editing tasks, Qwen-Image maintains remarkable coherence in both subject identity and background structure despite significant motion changes—an essential requirement for video generation. Moreover, unlike most image generation approaches that rely on image VAEs, we adopt a video VAE for visual representation. While this introduces greater modeling complexity, it aligns with our core objective: to build a foundation model that generalizes across diverse visual modalities, not just static images.

Qwen-Image as a "visual generation" model, in the context of integrated understanding and generation, advances the vision of seamless integration between perception and creation. We argue that achieving true understanding-generation unity rests on three foundational pillars: (1) mastering understanding, (2) mastering generation, and (3) mastering their synergistic integration. As the first work in the Qwen series dedicated to visual generation, Qwen-Image fills a critical gap in the second pillar—generation capability—complementing Qwen2.5-VL, which excels in visual understanding (the first pillar). Together, they form a balanced foundation for the next generation of multimodal AI, paving the way toward Visual-Language Omni systems that are not only capable of perceiving and reasoning but also of generating text-rich, visually coherent imagery—where language and vision are seamlessly fused into illustrative, readable, and semantically faithful visual outputs.

In summary, Qwen-Image is more than a state-of-the-art image generation model—it represents a paradigm shift in how we conceptualize and build multimodal foundation models. Its contributions extend beyond technical benchmarks, challenging the community to rethink the roles of generative models in perception, interface design, and cognitive modeling. By emphasizing complex text rendering in image generation and addressing classical understanding tasks such as depth estimation through the lens of image editing, Qwen-Image points toward a future in which: (1) generative models do not merely produce images, but genuinely understand them; and (2) understanding models go beyond passive discrimination, achieving comprehension through intrinsic generative processes. As we continue to scale and refine such systems, the boundary between visual understanding and generation will blur further, paving the way for truly interactive, intuitive, and intelligent multimodal agents.

\section{Authors}

\textbf{Core Contributors\footnote{Alphabetical order.}:} Chenfei Wu, Jiahao Li, Jingren Zhou, Junyang Lin, Kaiyuan Gao, Kun Yan, Shengming Yin, Shuai Bai, Xiao Xu, Yilei Chen, Yuxiang Chen, Zecheng Tang, Zekai Zhang, Zhengyi Wang

\textbf{Contributors\footnote{Alphabetical order.}:} An Yang, Bowen Yu, Chen Cheng, Dayiheng Liu, Deqing Li, Hang Zhang, Hao Meng, Hu Wei, Jingyuan Ni, Kai Chen, Kuan Cao, Liang Peng, Lin Qu, Minggang Wu, Peng Wang, Shuting Yu, Tingkun Wen, Wensen Feng, Xiaoxiao Xu, Yi Wang, Yichang Zhang, Yongqiang Zhu, Yujia Wu, Yuxuan Cai, Zenan Liu
\clearpage
\bibliography{colm2024_conference}
\bibliographystyle{colm2024_conference}

\end{document}